\documentclass{article}

\usepackage{arxiv}

\usepackage[utf8]{inputenc} 
\usepackage[T1]{fontenc}    
\usepackage{hyperref}       
\usepackage{url}            
\usepackage{booktabs}       
\usepackage{amsfonts}       
\usepackage{nicefrac}       
\usepackage{microtype}      
\usepackage{lipsum}
\usepackage{graphicx}

\usepackage{amsmath}
\usepackage{algorithm}
\usepackage[noend]{algpseudocode}
\usepackage[table,xcdraw]{xcolor}
\usepackage{graphicx}

\usepackage{subcaption}
\usepackage{mathtools, nccmath}
\usepackage{url}
\usepackage{xcolor}
\usepackage{amssymb}
\usepackage{latexsym}

\graphicspath{ {./images/} }

\title{Real-time Automatic M-mode Echocardiography Measurement with Panel Attention from Local-to-Global Pixels}

\author{
 Ching-Hsun Tseng \\
  Department of Computer Science\\
  The University of Manchester\\
  Manchester M13 9PR, UK\\
  \tiny Co-first author for the contribution of the methodology/algorithm development and empirical analysis
  \And
  Shao-Ju Chien \\
  Division of Pediatric Cardiology\\
  Department of Pediatrics\\
  Kaohsiung Chang Gung Memorial Hospital\\
  Kaohsiung, Taiwan\\
  \\
  School of Traditional Chinese Medicine\\
  Chang Gung University College of Medicine\\
  Tao-Yuan, Taiwan\\
  \\
  Department of Early Childhood Care and Education\\
  Cheng Shiu University\\
  Kaohsiung, Taiwan\\
  \tiny Co-first author for the contribution of providing the data and medical domain knowledge
  \\
  \And
  Po-Shen Wang \\
  Institute of Management of Technology\\
  National Yang Ming Chiao Tung University\\
  Hinschu 30010, Taiwan\\
  \And
  Shin-Jye Lee \\
  Institute of Management of Technology\\
  National Yang Ming Chiao Tung University\\
  Hinschu 30010, Taiwan\\
  \And
  Wei-Huan Hu \\
  College of Computer Science\\
  National Yang Ming Chiao Tung University\\
  Hinschu 30010, Taiwan\\
  \And
  Bin Pu \\
  Department of Electronic and Computer Engineering\\
  The Hong Kong University of Science and Technology\\
  Hongkon, China\\
  \And
  Xiao-jun Zeng \\
  Department of Computer Science\\
  The University of Manchester\\
  Manchester M13 9PR, UK\\
  \tiny Corresponding Author
  \texttt{x.zeng@manchester.ac.uk}
}

\begin{document}
\maketitle
\begin{abstract}
Motion mode (M-mode) recording is an essential part of echocardiography to measure cardiac dimension and function. However, the current diagnosis cannot build an automatic scheme, as there are three fundamental obstructs: Firstly, there is no open dataset available to build the automation for ensuring constant results and bridging M-mode echocardiography with real-time instance segmentation (RIS); Secondly, the examination is involving the time-consuming manual labelling upon M-mode echocardiograms; Thirdly, as objects in echocardiograms occupy a significant portion of pixels, the limited receptive field in existing backbones (e.g., ResNet) composed from multiple convolution layers are inefficient to cover the period of a valve movement. Existing non-local attentions (NL) compromise being unable real-time with a high computation overhead or losing information from a simplified version of the non-local block. Therefore, we proposed RAMEM, a real-time automatic M-mode echocardiography measurement scheme, contributes three aspects to answer the problems: 1) provide MEIS, a dataset of M-mode echocardiograms for instance segmentation, to enable consistent results and support the development of an automatic scheme; 2) propose panel attention, local-to-global efficient attention by pixel-unshuffling, embedding with updated UPANets V2 in a RIS scheme toward big object detection with global receptive field; 3) develop and implement AMEM, an efficient algorithm of automatic M-mode echocardiography measurement enabling fast and accurate automatic labelling among diagnosis. The experimental results show that RAMEM surpasses existing RIS backbones (with non-local attention) in PASCAL 2012 SBD and human performances in real-time MEIS tested. The code of MEIS and dataset are available at \url{https://github.com/hanktseng131415go/RAMEM}.
\end{abstract}


\section{Introduction}
\label{sec1}
\begin{figure}[!t]
    \centering
    \includegraphics[width=1.0\columnwidth]{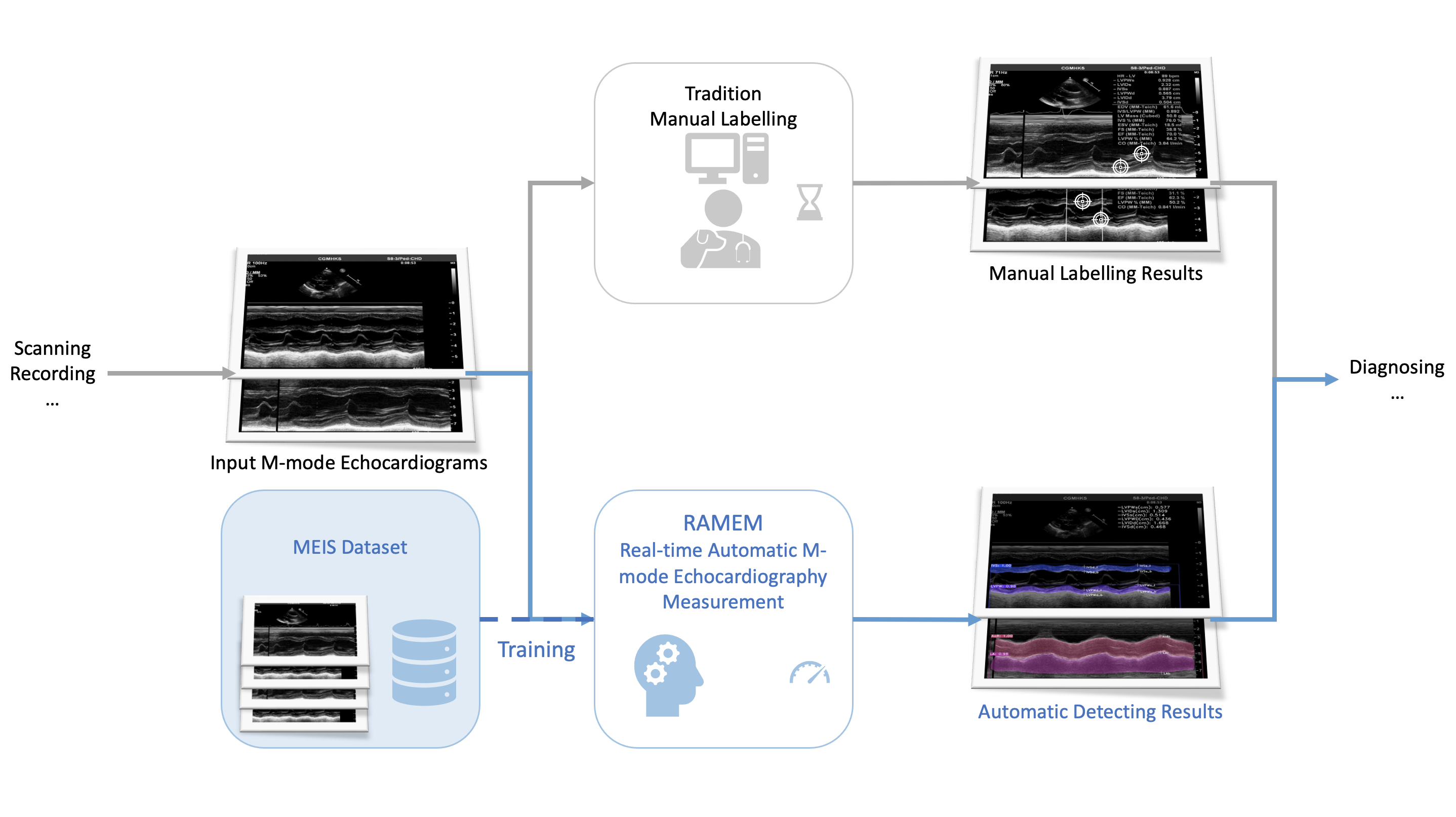}
    \caption{The proposed framework of RAMEM. The current diagnosis process is in grey, and the proposed one is in blue. The proposed process involves being trained by the proposed MEIS dataset and then detecting upon the learned labelling agreement from experts to fulfil the automatic echocardiogram measurement.}
    \label{fig1}
\end{figure}

More than 90\% of heart problems can be detected through the cardiac ultrasound screening examination \cite{otto2013textbook}. However, pediatric echocardiography is difficult because of the diverse capturing angles, patient resistance, variance of manual labelling, and time-consuming labelling. These factors influence the diagnostic results and could lead to an inaccurate diagnosis. Regarding the variance of manual labelling, requiring the same results from different examiners in a series of examination steps is impossible. Considering the current order of analysis process is: 1) scanning the left ventricle (LV) from different angles to capture the best LV image; 2) recording a sequence period image from systolic to diastolic phase; 3) manual anchor labelling on the sequence image; 4) diagnosing based on the measuring indices from the labelling and image. The issue of being unable to make a seamless diagnosis is caused by time-consuming manual anchor labelling 3). Apart from the issue of variance and time-costing in labelling, how to ensure an accurate examination is ushered. Objects in motion mode images from a sequence recording LV images (M-mode) typically possess a significant portion of pixels in the echocardiogram. To make an accurate and efficient diagnosis by object detection, using pure convolutional neuron networks (CNN) as existing backbones (e.g., ResNet) in the most real-time instance segmentation (RIS) works with limited receptive field seem unreasonable because CNN often loses connection from pixel-to-pixel \cite{guo2022beyond}. The high computational overhead in non-local (NL) attention \cite{wang2018non} makes CNN over-relying on NL that sacrifices efficiency \cite{cao2019gcnet}. The stacking of multiple layers of CNN might make the model cumbersome, which transfers the risk of losing information from pixel to depth. Therefore, global attention without compromising information and efficiency is needed. This work manages to solve the problems above with corresponding solutions. We propose RAMEM, a real-time automatic M-mode echocardiography measurement scheme, seeing the overall framework compared with the traditional pipeline in Fig. \ref{fig1}. 

Regarding the labelling variance, this work aims to replace human bias with an automatic computer vision method based on an agreement with multiple experts’ labels. The current status is that there is a series of automatic deep learning works \cite{leclerc2019ru}\cite{smistad2020real}\cite{pu2021fetal}\cite{saeed2022contrastive}\cite{painchaud2022echocardiography}\cite{girum2021learning} training on CAMUS \cite{leclerc2019deep}, which is a dataset toward 2D echocardiography (some works refer it B-mode), but there is no dataset for M-mode echocardiography to the best of our knowledge. On the one hand, to justify the need for an M-mode dataset, while most works estimate LV indicators using B-mode spectral flow Doppler based on the predicted segmentation, a direct M-mode examination remains a precise linear measurement of cardiac dimensions. In other words, in a practical clinical routine, M-mode examination is still the most intuitive non-invasive way to assess cardiac function and wall thickness \cite{otto2013textbook} by observing a sequence of LV contraction in one recording image, which is not so easy to observe from a B-mode, especially for a precise period observation among systolic and diastolic phase. No work, on the other hand, is trying to bridge the field of echocardiography with RIS, viewing the features in RIS have: 1) already existed a variety of fine works for boosting performance; 2) been real-time for seamless diagnosis; 3) been able to predict mask for segmentation and measurement; 4) bounding-boxes for dynamic excluding unwanted noise; 5) followed mainstream format and standard (e.g., COCO). To bridge the RIS and M-mode echocardiography, this work offers MEIS, a dataset following the RIS framework and standard toward M-mode echocardiography for instance segmentation. This makes it possible to ensure consistent results from data based on strict expert data agreement.

\begin{figure}[!t]
    \centering
    \includegraphics[width=1.0\columnwidth]{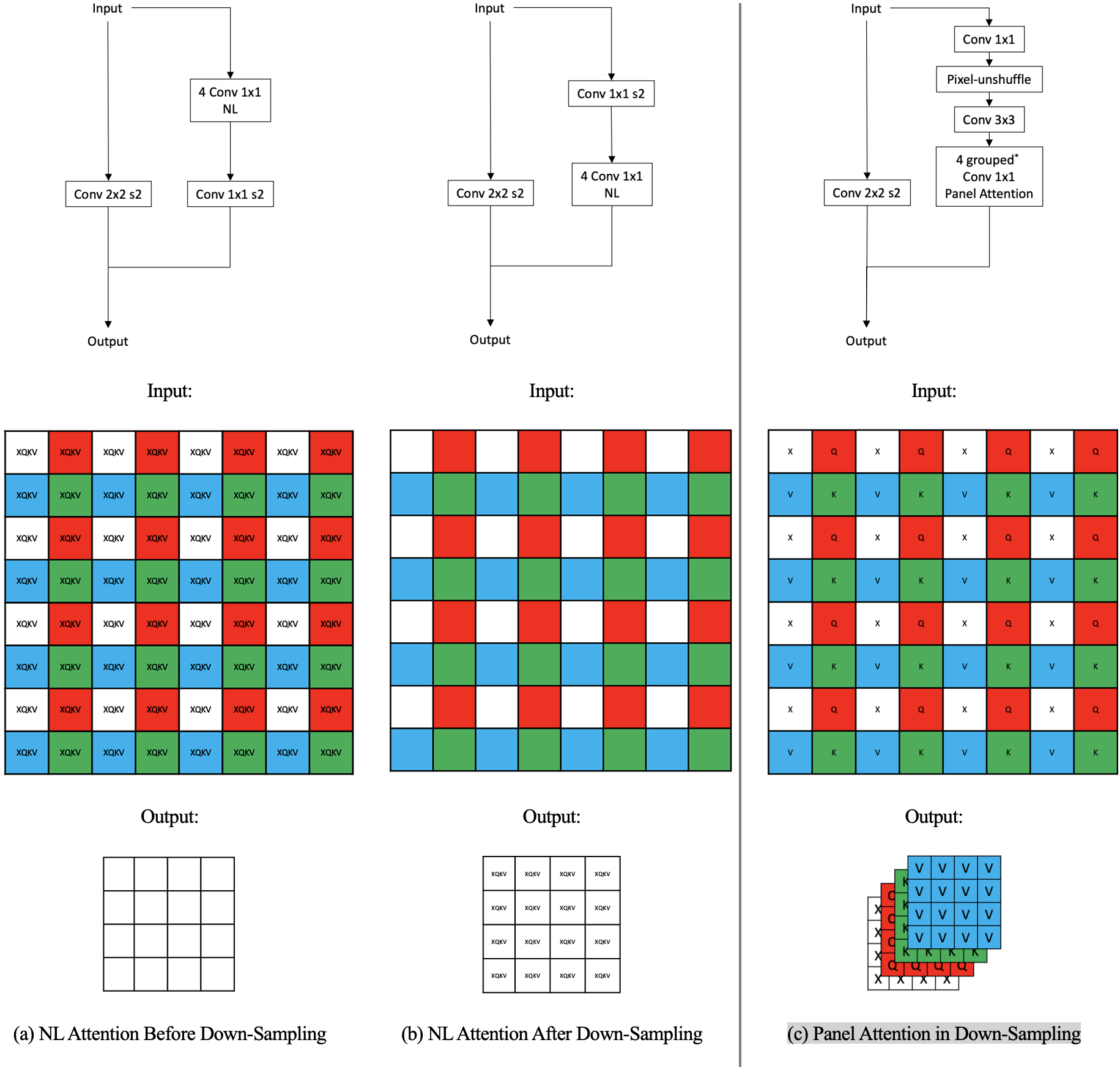}
    \caption{Panel attention comparison. Different NL attention scenarios in (a) and (b), where (a) operates NL attention on the linear operation (weight assignment) of XQKV firstly and down-sample in 1×1 CNN stride 2 followingly, this makes only white pixels with global information pass through, please see an explanation of XQKV of NL in (\ref{equ1}); (b) processes in the opposite way, this makes only white pixels involve global attention; (c) assigns weight by the same linear operation but in a square-clockwise order like panel, which creates a group-like weight connection toward different value (*). Namely, it is less weight and less information loss than NL. It, then, operates attention along with respective weight in down-sampled size. This operation covers all pixels with attention in an efficient way. QKV is the typical linear operation, X is the linear operation toward original information or skip-connection value toward output, and NL inter-channel is neglected for the simple demonstrations.}
    \label{fig2}
\end{figure}

\begin{table}[]
\centering
\caption{\label{tab1}Properties Comparison among CNN, NL, and Panel Attention}
\resizebox{\columnwidth}{!}{%
\begin{tabular}{llll
>{\columncolor[HTML]{F2F2F2}}l }
\hline
                & CNN                                                                          & \begin{tabular}[c]{@{}l@{}}NL\\ (Attention Before Down-Sampling)\end{tabular} & \begin{tabular}[c]{@{}l@{}}NL\\ (Attention After Down-Sampling)\end{tabular} & \begin{tabular}[c]{@{}l@{}}Panel\\ (Attention In Down-Sampling)\end{tabular}                                   \\ \hline
Complexity      & $O\left(ck^2\right)$                                                         & $O\left({(cwh)}^2\right)$                                                     & $O\left({(c\frac{wh}{2})}^2\right)$                                          & \cellcolor[HTML]{EFEFEF}$O\left({(\frac{c}{16}\frac{wh}{2})}^2\right)$                                         \\
Receptive Field & Local                                                                        & Global                                                                        & Global                                                                       & Local-Global                                                                                                   \\ \hline
Method          & \begin{tabular}[c]{@{}l@{}}Down-sampling\\ Dilated\\ Depth-wise\end{tabular} & Dot-product                                                                   & Dot-product                                                                  & \begin{tabular}[c]{@{}l@{}}Down-sampling\\ (pixel-unshuffle)\\ Dilated\\ Depth-wise\\ Dot-product
\end{tabular}\\ \hline
\multicolumn{5}{l}{$c$ indicates a channel number, $wh$ indicates the width and height of images}
\end{tabular}%
}
\end{table}

Offering accurate measurement on M-mode echocardiograms must rely on accurate detection through a big receptive field in the backbone, as the capturing region shall be big enough to cover the period of a valve movement. A mature scheme is an excellent stepping-stone to building the measurement. In the field of RIS, YOLACT \cite{bolya2019yolact}\cite{bolya1912yolact++} is the first and most robust one. Despite the merits of YOLACT's backbone of ResNets \cite{he2016deep}, the CNN-based backbone suffers from the limited receptive field \cite{wang2018non} and then causes the sub-optimal problem hovering. The sub-optimal issue could further drag down the detected efficiency \cite{lee2020centermask}, which is responsible for frame per second (FPS) toward detecting an image. For fixing the problem, updating the backbone to an advanced one, such as DenseNets \cite{huang2017densely} and UPANets \cite{tseng2022upanets}, is helpful. The same direction can also be seen from other works in RIS as well, such as CenterMaks by updating with a more efficient VoVNets V2 \cite{lee2020centermask}; RTMDet with CSP-DarkNets \cite{lyu2022rtmdet}. However, the limited horizon by CNN still remains. The most common techniques to conquer the limitation include 1) down-sampling convolution by making a big object small enough to fit in a receptive field; 2) dilated convolution by expanding distance between kernel elements; 3) depth-wise convolution \cite{chollet2017xception} by creating deep fashion among two micro convolution layers. Nonetheless, the fundamental problem still exists in these CNN tips: the receptive fields are essentially limited because the kernel size shapes limited receptive size. In fact, despite expanding the kernel in dilated convolution to cover the whole image in 2) and 3), the identical downside of losing information in 1) merely transfers from a down-sampling operation into occurring in the sparse blanks. While ViT \cite{dosovitskiy2020image} and non-local \cite{wang2018non} attention have demonstrated that a series of dot-products across pixels can make global attention, the sacrifice is high computation overhead (attention before down-sampling). We furtherly add another argument that it will also lose too much information for giving efficient attention when applying to a down-sampled size (attention after down-sampling) for bypassing the overhead issue, the same scene occurring in simplified non-locals. In this context, a question arises: \textit{Is there an efficient method operating with the merits from local to global?} Our answer to this question is: Panel attention, depth-wise local-to-global attention by pixel-unshuffling. This creates a third category of attention in down-sampling. The process covers local attention with a generally grouped 3×3 kernel to mimic depth-wise dilated convolution cause of pixel-unshuffle. Viewing the advantage of lossless in pixel-unshuffle as down-sampling, it makes the product of weight assigning in “panel groups”, like the order in a LED panel, form an efficient NL attention. By embedding the Panel attention into our developed UPANets \cite{tseng2022upanets}, the whole operation composes an all-around avoiding sub-optimal and local-to-global attention with the cared merits in both CNN and NL. Please see the demonstration and comparison of this discussion in Fig. \ref{fig2} and Table \ref{tab1}.

An automatic algorithm to be proposed and developed for different views of M-mode echocardiograms can make a seamless diagnosis. Despite the impossibility of entirely removing human operations, automatic detection and measurement can maintain consistency and help to release the burden from medical professionals, which could keep a potential misdiagnose at bay. The algorithm to be proposed and developed is called AMEM, automatic M-mode echocardiography measurement, and could create an ideal environment by directly starting a diagnosis from an instant-showing wanted indices based on the current ultrasound image.

This work aims to fulfil the gaps and overcomes the existing issues in M-mode echocardiography examination along with real-time instance segmentation, including 1) the variance of manual labelling, 2) the compromising information and efficiency, and 3) unable seamlessly diagnose in M-mode echocardiography. To achieve these aims, the proposed framework is given in Fig. \ref{fig1}. The contributions of this work toward the problems to be overcome are listed as follows: 
\begin{itemize}
\item The variance of manual labelling – open-access MEIS, a dataset of M-mode echocardiograms for instance segmentation. The proposed RIS scheme building can diminish the variance upon the dataset with experts’ agreement.
\item Compromising information and efficiency – provide Panel attention, a depth-wise local-to-global efficient attention by pixel-unshuffling, embedding with updated UPANets V2 in RIS with global receptive field.
\item Unable seamless diagnosis in M-mode echocardiography – propose an efficient algorithm of AMEM targeting fast and accurate automatic labelling among diagnoses.
\end{itemize}

The background, motivation, and contributions have been discussed in this section, \ref{sec1}. Section \ref{sec2} covers related works from echocardiography, RIS, and attention in computer vision. Section \ref{sec3} reveals the details of MEIS. Followingly, Section \ref{sec4} presents panel attention in UPANets V2 and AMEM in RAMEM. A series of evaluations and discussions with competitors toward PASCAL 2012 SBD and MEIS are revealed in Section \ref{sec5}. The conclusion is placed in Section \ref{sec6}.

\section{Related Work}
\label{sec2}
This section contains the related works from echocardiogram datasets, automatic echocardiography, RIS, and the backbone's local \& global attention in computer vision. The details can be seen in the following subsections.

\subsection{M-mode Echocardiography Datasets and Automatic Echocardiogram}
\label{sec2.1}
Echocardiogram Datasets – The direction \cite{zamzmi2020harnessing} of automatic echocardiogram detection has recently shrunk into focusing on one direction, B-mode. A survey work \cite{zamzmi2020harnessing} on the development of automatic echocardiograms since 2004 has supported this view by showing the distribution: 63 works in B-mode, 28 in Doppler echo, and only 3 in M-mode. Even worse, viewing three datasets for B-mode in 63 works, there is no dataset for M-mode. 

Automatic Echocardiography – Applying an automatic scheme to diminish the bias from a human operator is one of the intuitive approaches. Still, the state quo in M-mode diagnoses remains manual operation. Although EchoNet \cite{ghorbani2020deep}\cite{ouyang2020video} follows a similar notion to eliminate the bias or some works \cite{painchaud2022echocardiography}\cite{zamzmi2022real} try to estimate the LV mass from B-mode, they are still un-paired with the intuitiveness and accuracy by directly viewing M-mode. This situation could explain why the daily clinic examination of M-mode still stays in manual labelling. To the best of our knowledge, there are only three works which try to break the ice from mode classification, machine learning, and deep learning for animals in order: 1) vanilla fully-connected CNN \cite{madani2018fast} is applied to classify 15 modes of images (including M-mode) for replacing manual categorizing and speeding up the future down-stream work; 2) The work of using pair-wise distance offsets in \cite{fancourt2008segmentation} might be the first approach try to segment the anterior and posterior wall from a vessel in a motion mode image but the downside of noise disturbing could cloud the performance; 3) MENN \cite{duan2022fully} by Pfizer faces the automatic issue by deep learning framework toward animal B-mode and M-mode echocardiography. As a result, MENN might be the closest one to our work. However, it still involves some manual hyperparameter setting (e.g., setting sampling period) in detecting tissue boundaries and cannot tell the process in real-time or not. 

\subsection{Real-time Instance Segmentation}
\label{sec2.2}
Instance segmentation is the task of classifying the pixel category on an image. Real-time generally indicates a whole process from input to output after non-maximum suppression (NMS) is under roughly 0.33 sec ($>$30 FPS) \cite{bolya2019yolact} or 0.42 ($>$24 FPS) \cite{oksuz2021mask}. YOLACT \cite{bolya2019yolact}, a single-stage anchor-based instance segmentation model, first arrived at the real-time standard based on the structure of YOLOs \cite{redmon2018yolov3}\cite{ge2021yolox} and the scheme in RetinaNet \cite{lin2017focal}. YOLACT combines speed considerations and comprises backbone, neck, heads, label assignment, and NMS as a whole framework. The output includes classes, bounding boxes, and masks. The module for each part is backbone: ResNets50; neck: FPN; heads: shallow RetinaNet head; label assignment: anchor-based IoU assignment; NMS: FastNMS. Until a throughout investigation between anchor-based and anchor-free in ATSS \cite{zhang2020bridging}, the essential difference between the two was foggy. By ATSS, if aligning every part of the framework with a dynamic label assignment, the performance has no significant difference from the anchor-based and anchor-free framework. Another YOLACT-based framework with ATSS and the mask-aware intersection of union (maIoU), maYOLACT \cite{oksuz2021mask}, had become the best performance work in RIS in 2022 and has concreted the standpoints in ATSS. Recently, RTMDet \cite{lyu2022rtmdet} has tried to optimize every submodule based on YOLOX \cite{ge2021yolox}, which uses CSPDarkNet, to be the latest model in RIS. Since the introduction of YOLACT, every work ends with finetuned ResNets from ImageNet pre-trained weight to gain a superior performance. That intention causes keeping the same backbone (ResNets) in most works, such as {ma}YOLACT{++}, SipMask \cite{cao2020sipmask}, BlendMask \cite{chen2020blendmask}, and SOLO V2 \cite{wang2020solov2}. However, as the pre-trained weight from ImageNet significantly differs from the nature of echocardiography, applying such backbones will not benefit our task. Moreover, the same intention could restrain the development of the global attention module in this field. 

\subsection{Attention in Computer Vision}
\label{sec2.3}
Local Attention – Based on the aggregation of the kernel, CNN can be viewed as local attention because the attention area is constrained into k×k in comparing the result of NL. That can be said, from a broad perspective, most CNNs belong to this category. ConvNext \cite{liu2022convnet} has argued that modifying ResNets with modern techniques can make CNN outshine Transformer-based networks, such as SwinTransformer \cite{liu2021swin}. However, from a narrow viewpoint, LR-Net \cite{hu2019local} argues that local relation is also vital. It generates appearance composability from the local connection by applying Softmax toward a specific dimension. Similarly, SASA \cite{ramachandran2019stand} replaced CNNs with the proposed local attention to prove the concept with superiorities. 

Global (Non-local) Attention – The introduction of ViT and NL has caused great attention on global attention toward an image. To begin with, capturing such information from a series of dot-products indeed contains spatial and channel information into one. Taking a 2D image feature $X \in \mathbb{R}^{c \times s}, s=w \times h$ as an example, the NL can be expressed as follows:
\useshortskip
\begin{flalign}
\label{equ1}
Y=\left[Softmax\left(Q^T\otimes K,\ s\right)\right]\otimes V^T,&&
\end{flalign}
where $Softmax(input,dim)$, and  $Y,Q,K,V\in\mathbb{R}^{c\times s}$, especially $Q,K,V$ belonging the products of weight assign by linear operations from $X$. By equation (\ref{equ1}), the spatial and channel information is aggregated to the product. Nonetheless, because of the notorious high computation overhead, this hardware-unfriendly method is unsuitable for an efficiency-demanded environment. Thus, there can be three categories to deal with global information efficiently: 1) CNN-mimics, 2) Algorithm-simulators, and 3) NL-variants. 1) is referred to works claiming that using a big enough kernel can catch the same effect as NL, seeing dilated convolution in VAN \cite{guo2022visual} and VapSR \cite{zhou2022efficient}, but we do not view CNN that merely using a big kernel in this category, e.g., ConvNext; The standpoint of 2) is simulated non-local attention can be generated from a delicate algorithm. In EMANet \cite{li2019expectation}, the authors described multiple iterations of expectation-maximization with hyperparameters can find the robust attention status. At the same path as HAMs \cite{geng2021attention}, it believes that the matrix decomposition can do better than non-local attention with less memory and computation; 3) is a modification upon NL. Among the modifications, A2 \cite{chen20182} changes the operation of NL in doing $\left[Softmax\left(Q,s\right)\otimes K^T\right]\otimes Softmax(V,c)$ makes the complexity lower. On the one hand, GCNet \cite{cao2019gcnet} argues that attention maps across c arguably perform an identical effect, and thus the need to operate in c is unnecessary. Therefore, SENet \cite{hu2018squeeze} following simplified NL in $X^\prime\in\mathbb{R}^{1\times s}$ should be enough. EANet \cite{guo2022beyond} pushed this direction even further with two layers of perceptron accompanying with $Softmax(X^\prime,s), X^\prime\in\mathbb{R}^{\frac{c}{64}\times s}$ as inter feature maps. In the remaining variants, ANNN \cite{zhu2019asymmetric} uses adaptive average pooling to down-size feature maps, and AA \cite{bello2019attention} divides feature maps in multi-head form toward spatial to alleviate the overhead in spatial. We put our Panel Attention in category 3) with caring not only for efficiency but also for local and global information intact.

\section{MEIS Dataset}
\label{sec3}

This work presents a dataset for bridging M-mode echocardiography and RIS. To our best knowledge, it is the first M-mode echocardiogram dataset and bigger than the existing large datasets, CAMUS. Therefore, we name this dataset MEIS, M-mode echocardiograms for instance segmentation. Table \ref{tab2} and the following sub-sections reveal the dataset distribution and details. 

\begin{table}[]
\centering
\caption{\label{tab2}\\Data distribution of MEIS.}
\resizebox{\columnwidth}{!}{%
\begin{tabular}{llll}
\hline
\begin{tabular}[c]{@{}l@{}}View\\ \end{tabular}            & \begin{tabular}[c]{@{}l@{}}Object\\ (mask)\end{tabular}                                                         & \begin{tabular}[c]{@{}l@{}}Indicators\\ \end{tabular}                         & \begin{tabular}[c]{@{}l@{}}Training/Testing\\ (number)\end{tabular} \\ \hline
\begin{tabular}[c]{@{}l@{}}Aortic Valve\\ (AV)\end{tabular}   & \begin{tabular}[c]{@{}l@{}}Aortic Root (AoR),\\ Left Atrium (LA)\end{tabular}                                   & \begin{tabular}[c]{@{}l@{}}AoR Diameter,\\ LA Dimension\end{tabular}             & 747/559                                                             \\
\begin{tabular}[c]{@{}l@{}}Left Ventricle\\ (LV)\end{tabular} & \begin{tabular}[c]{@{}l@{}}Interventricular Septum (IVS),\\ Left Ventricular Posterior Wall (LVPW)\end{tabular} & \begin{tabular}[c]{@{}l@{}}LVIDs, LVPWs, IVSs,\\ LVIDd, LVPWd, IVSd\end{tabular} & 774/559                                                             \\ \hline
\end{tabular}%
}
\end{table}

\subsection{Description}
\label{sec3.1}
MEIS comprises a total of 2,639 images in the size of $1024\times768$ toward two recording views (Aortic Valve (AV) and Left Ventricle (LV)) with 1,521 (747 in AV + 774 in LV) images for training and 1,118 (559 in AV + 559 in LV) for testing, respectively. Each view must be detected with two objects to calculate the measurement indicators.  That is in total with four object classes (two objects in each view):  aortic root (AoR) and left atrium (LA) in AV; interventricular septum (IVS) and left ventricular posterior wall (LVPW) in LV. The medical meaning and purpose of each indicator are listed in the following:
\begin{itemize}
\item AV: LA-Dimension and AoR-Dimension can be measured for calculating different indicators, such as AoR/LA ratio, to examine the state of the aortic valve.
\item LV: 6 measurements include IVSs, IVSd, LVIDs, LVIDd, LVPWs, and LVPWd. These concerned thicknesses and dimensions in LV recording are used to estimate other cardiac functions through specific medical formulas, including LV mass, LV ejection fraction, end-diastolic volume, end-systolic volume, and more \cite{wandt1999echocardiographic}\cite{mizukoshi2016normal}.
\end{itemize}

\subsection{Preparation}
\label{sec3.2}
The source of images is collected and approved within the regulation set by the ethical committee of the Chang Gung Medical Foundation Institutional Review Board. The dataset is recorded from Philips EPIQ7 and iE33 ultrasound machines toward infants. After de-identification under physicians’ supervision, the preparation results require object locations, object classes, and object masks. The data preparation, thus, is done with physicians through a series of preparations, from manually drawing masks to file conversion. The detailed descriptions in each section are below.

Manual Mask Drawing – Masks of objects are critical for image segmentation tasks. For M-mode echocardiograms, the traditional way is manually setting multiple anchors in interested locations among a valve wall. This method, however, could lead to a wrong location. To avoid this, a careful mask drawing on a still recorded image could be a solution because the mask can represent the accurate belonging boundary from object to object. Thus, in raw AV-recording images, the contours of AoR and LA are drawn manually along with the experts. In raw LV-recording images, IVS and LVPW are processed similarly. 

\begin{figure*}[!t]
    \centering
    \includegraphics[width=1.0\textwidth]{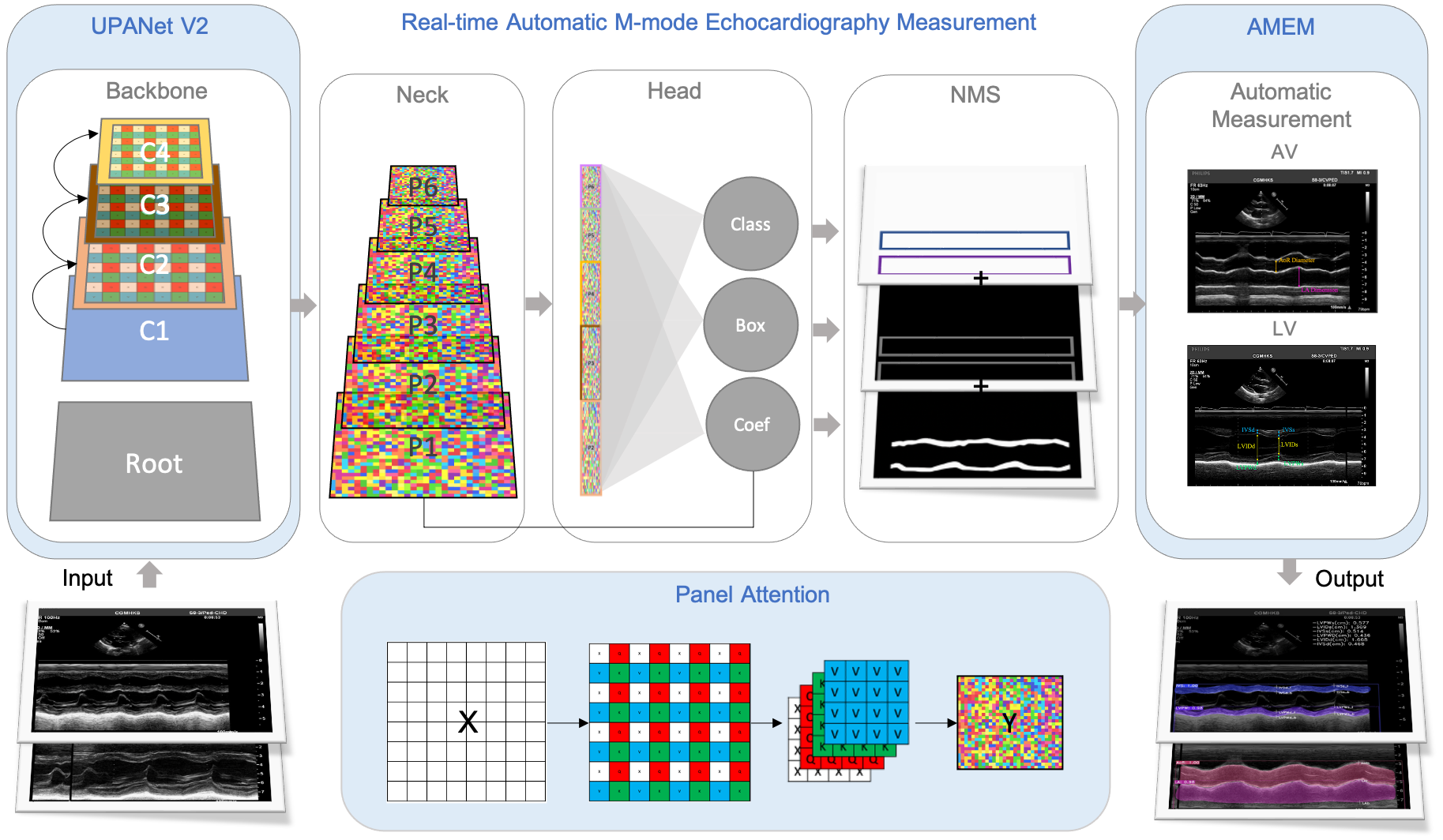}
    \caption{The Detailed of RAMEM. The proposed methods are in blue. UPANets V2 is updated with panel attention equipped with global attention ability. After going through the process from backbone to NMS, the proposed automatic algorithm, AMEM, is for measuring wanted indicators toward M-mode echocardiograms.}
    \label{fig3}
\end{figure*}

File Conversion (Bounding Box to COCO Format) – Bridging between the M-mode echocardiography and RIS is one of the purposes of this work. Thus, a dataset should follow the mainstream RIS, COCO format in JSON. This action also brings another advantage to M-mode echocardiography: “bounding box”. Considering different viewing angles could cause different mask scales, fixed and manual-based postprocessing could cloud the meaning of automation. In contrast, the bounding box of an object could be the perfect solution, as the bounding box already has the merit of focusing on the mask in the bounding box. In this case, the only work left for automation is perfecting the generation of masks. Moreover, the current detection strategy of RIS is also based on a bounding box, making the bounding box even more vital. After the mask drawing, bounding boxes of objects are decided upon the location of the masks. The classes of masks, along with the bounding box location, are also organized in JSON format. Finally, because the images in COCO format are typical in JPG format \cite{lin2014microsoft}, we transfer the original file of Digital Imaging and Communications in Medicine (DICOM) into JPG. In sum, a COCO format M-mode echocardiography is followed with JPG images, masks, classes, bound boxes, actual measurement index values, and image pixel-to-cm ratio.

\section{RAMEM}
\label{sec4}

This work is developed based on a mature RIS scheme, (ma)YOLACT. The proposed methods are operating upon the scheme. The proposed methods include new local-to-global attention with an updated backbone and an automatic measuring algorithm for M-mode echocardiography. As an overall pipeline compared with the traditional one has been shown in Fig. \ref{fig1}, a more detailed framework demonstration that covers the backbone, attention, and algorithm can be seen in Fig. \ref{fig3} and the following subsections.

\subsection{Panel attention}
\label{sec4.1}
Typically, objects in echocardiograms occupy a great portion of the pixels. However, this scene is not a privilege that big objects also appear in the real-world scene. In fact, having a big receptive field is one of the ultimate goals of computer vision. The typical stacking deep CNN makes expanding the receptive field unintuitive and somewhat cumbersome. While learning information in deep, the risk of losing vital features during passing ramps up, making the sub-optimal problem worse. On the other hand, to capture the whole picture in one receptive field like NL, the cumbersome problem only transfers from depth to spatial complexity; see the issue among different NL cases in Fig. \ref{fig2}. To showcase our answer, panel attention formulation should be a better candidate in these aspects: 1) lossless operation in pixel-unshuffle, 2) caring local information in CNN, and 3) capturing global information more efficiently. Please see the detailed demonstration in Fig. \ref{fig4} and along with the formulation, which is later embedded in updated UPANets to replace the channel and spatial attention as UPANets V2 acting as a backbone of RIS. 

\subsubsection{Formulation}
\label{sec4.1.1}
\begin{figure}[!t]
    \centering
    \includegraphics[width=1.0\columnwidth]{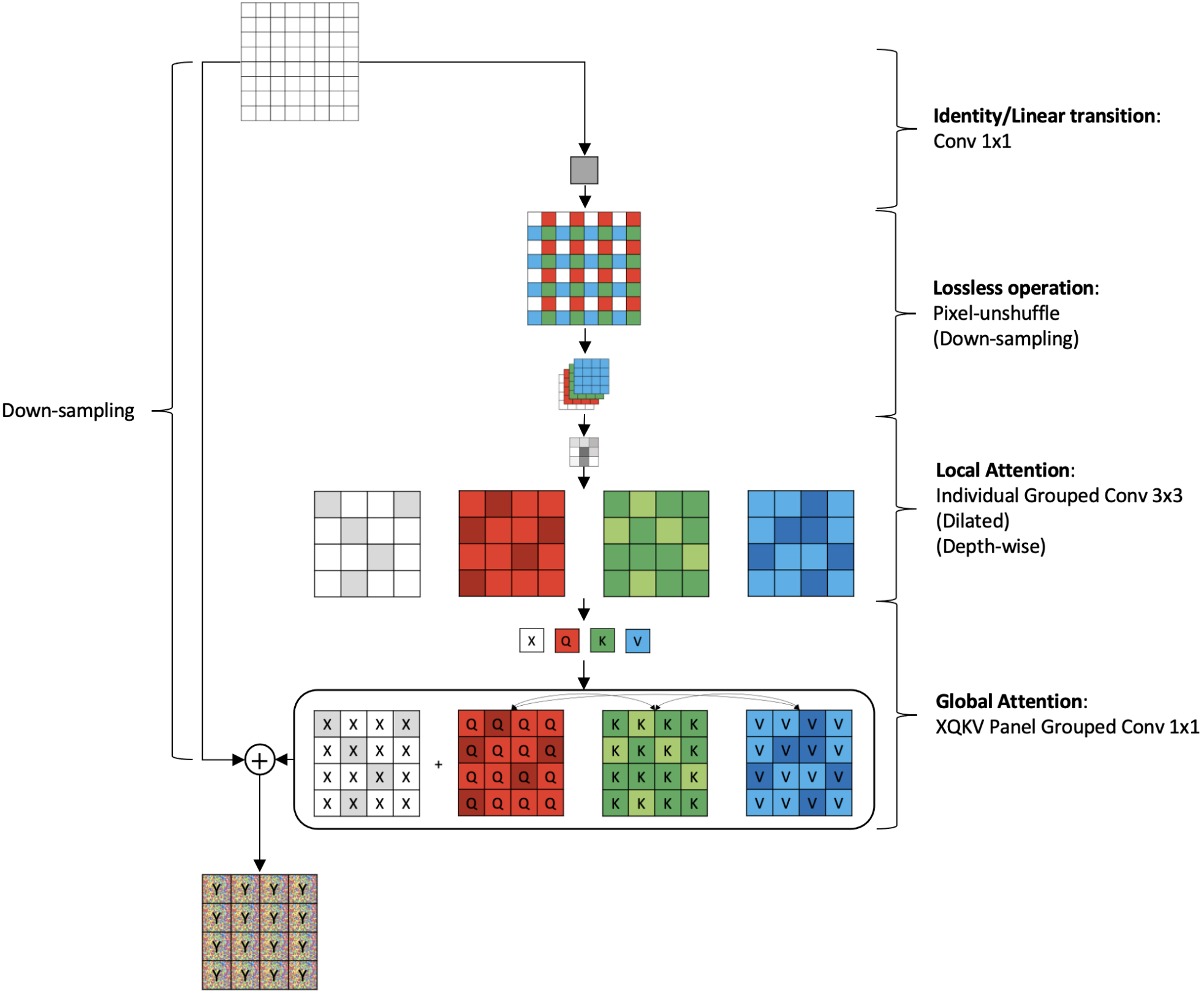}
    \caption{The components cooperate as a whole in panel attention. Taking the input feature map in $1\times8\times8$, the output is $1\times4\times4$. The Conv stride in 1 is ignored for easy presentation.}
    \label{fig4}
\end{figure}

Panel attention follows and contains four components from Identity/linear transition (Conv $1\times1$), pixel-unshuffle, local attention (grouped Conv $3\times3$), and global attention (panel grouped Conv $1\times1$). Apart from the first transition, the rest of the components shape the characters of the panel attention. Please refer to Fig. \ref{fig4} with the following description in each sub-section.

Identity/Linear transition – This transition follows other NLs to align channel numbers and can be replaced in an identity operation. If a learnable linear transition is applied, a pixel weight assignment is fulfilled by a fully-connected layer. Otherwise, the original information is sent intact to the next operation. Please see the equation below:
\useshortskip
\begin{flalign}
    \label{equ2}
    X_{_{l}}=\left\{\begin{array}{c}
    W\left ( X \right )\\ 
    X
    \end{array}\right.,&&
\end{flalign}
where $W$ refers to linear transition, and ${X,X}_l\in\mathbb{R}^{c\times s}$. 

Lossless operation – Pixel-(un)shuffle is known as preserving information among moving spatial and depth bilaterally. This transition by pixel-unshuffle is suitable for the most severe computation overhead issue from the spatial dimension in NL because the spatial information is transferred into the depth. This operation down-samples feature maps in a lossless way as follows:
\useshortskip
\begin{flalign}
    \label{equ3}
    X_s=P^-\left(X_l\right),&&
\end{flalign}
where  $X_s\in\mathbb{R}^{4c\times\frac{s}{4}}$, and $P^-$ represents a pixel-unshuffle operation. Pixel-unshuffle can be seen in Fig. \ref{fig4} where different colour pixels in the dilated arrangement are then organized into a small matrix in the belonging colour. However, this transition brings another issue: increasing the channel number (depth) by four times. The remedy for this side effect is presented in global attention by panel grouped Conv $1\times1$.

Local attention – Having $3\times3$ Conv upon the unshuffled pixel creates a dilated effect, expanding the receptive field to  $5\times5$; please see the effect in Fig. \ref{fig4} where pixels separate according to colour. Also, using individual grouped CNN further forms a depth-wise convolution for the following layers. Besides, because of the in-existing of normalization and activation between depth-wise layers for the next layers, it preserves the local aggregation information to build appearance composability:
\useshortskip
\begin{flalign}
    \label{equ4}
    X_{la}=\theta\left(X_s,\ K\right),&&
\end{flalign}
where $\theta$ refers to the general CNN aggregation with kernel $K\in\mathbb{R}^{c\times k\times k}$, la belonging to the abbreviation of local attention. Default $k=3$. 

Global attention – Continuing the former compounds, two missions must be accomplished as fine attention: remedy the increased channel dimension and make non-local attention. A special grouped weight assign policy, panel group, is proposed to remedy the expanded channel. Unlike generally grouped CNN, the panel group selects each channel in each p step as a group. In other words, the sum of weights equals general grouped CNN in 4. Please see the equations:
\useshortskip
\begin{flalign}
    \label{equ5}
    \left\{{X_{la}}^\prime,Q_{la},\ K_{la},V_{la}\ \right\}=W\times select\left(X_{la,i},\ p\right),&&
\end{flalign}
\useshortskip
\begin{flalign}
    \label{equ6}
    X_{ga}=f\left(Q_{la},\ K_{la},V_{la}\right),&&
\end{flalign}
\useshortskip
\begin{flalign}
    \label{equ7}
    Y=\sigma\left(N\left({X_{la}}^\prime+X_{ga}\right)\right)+X,&&
\end{flalign}
here $p=4$ in default, $\left\{{X_{la}}^\prime,Q_{la},\ K_{la},V_{la}\ \right\}\in\mathbb{R}^{4c\times\frac{s}{4}}$ with the same meaning as equation (\ref{equ1}) but in a small size, ${X_{la}}^\prime=WX_{la,\ i=1,5,\cdots n-4}$ as a linear transition. $\left\{Q_{la},\ K_{la},\ V_{la}\right\}$ follow the same weight assignment (linear transition) policy as ${X_{la}}^\prime$ but with $i=2,3,4$ as the start, respectively. For example, ${X_{la}}^\prime$ takes the $1st ,5th, …, n-4th$ channels, $Q_{la}$ takes $2nd, 6th, …, n-3th$ channels, etc.  $f$ represents the same dot-product at equation (\ref{equ1}) in NL. N and $\sigma$ refer to normalization and activation separately in skip-connection. ga belongs to the abbreviation of global attention. Finally, the output will be added upon $X$ as the final output.

\subsection{UPANets V2}
\label{sec4.2}
Considering the sub-optimal risk and the potential inefficiency in ResNet, we opt for an efficiency backbone, UPANets, as the backbone and update upon it in the YOLACT scheme. In ConvNext, there are debates about whether to use normalization/activation, their number, and how to properly down-sample. The updated propositions, except updating channel \& spatial pixel attention by Panel attention, are as follows:
\begin{itemize}
    \item Activation: apply PReLU viewing the parametric can be identity output as $p=1$ and $0\le p<1$ as typical activation. 
    \item Normalization: except $Softmax$ as global attention normalization, layer normalization is used as spatial normalization and batch normalization is used afterwards to form a double normalization inspired by EANet double normalization policy.
    \item Down-sampling: a patch (separate) down-sampling of stride 2 Conv $2\times2$ is applied as skip-connection in origin. 
\end{itemize}

\subsection{Automatic M-mode Echocardiography Measurement}
\label{sec4.3}
\begin{figure}[!t]
    \centering
    \includegraphics[width=1.0\columnwidth]{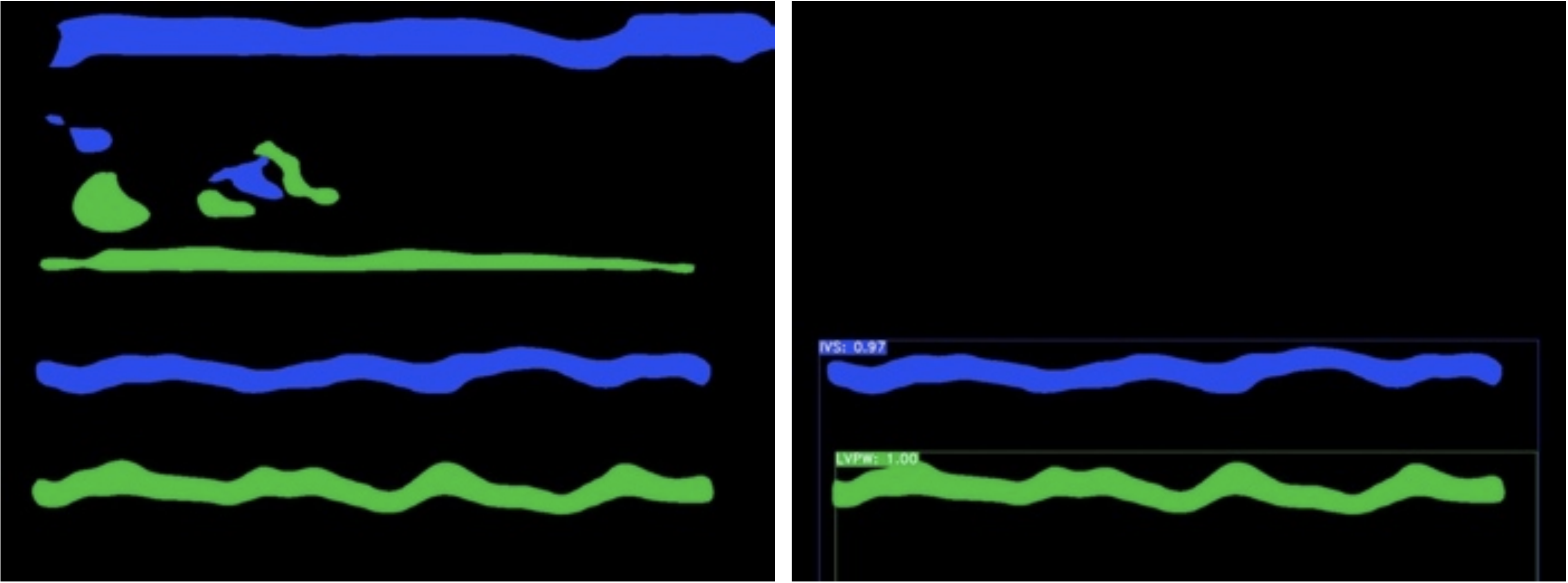}
    \caption{Postprocessing with a bounding box, the pre-image on the left and the post-image with filtering out un-wanted masks on the right.}
    \label{fig5}
\end{figure}

After getting classes, bounding boxes, and masks from the scheme in YOLACT, a post-process is implemented after NMS to eliminate noise outside the bounding boxes. The bounding boxes aim to make the following measurement focus on the wanted area, seeing Fig. \ref{fig5}. According to the desired indicators in Table \ref{tab2}, the measuring method is divided into two views:
\begin{itemize}
    \item AV: for AoR Diam from AoR; and LA Dim from LA;
    \item LV: for LVPWs and LVPWd from LVPW (systole/diastole); for IVSs and IVSd from IVS (systole/diastole); for LVIDs and LVIDd among LVPW to IVS (systole/diastole).
\end{itemize}
The indicators are measured with corresponding algorithms based on medical experiences.

\begin{figure}[!t]
    \centering
    \includegraphics[width=1.0\columnwidth]{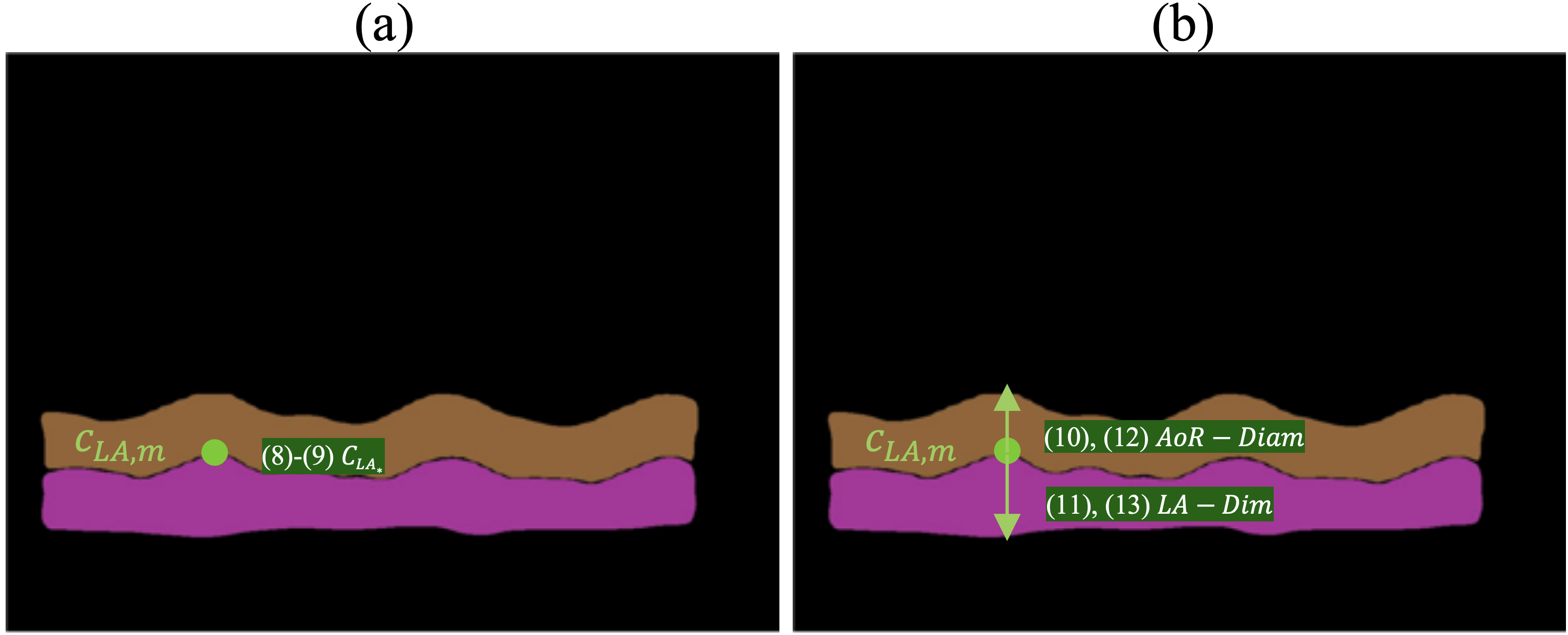}
    \caption{AV indicators measurement. (a) is the coordinate of the topmost. (b) is the result of finding by scanning up and down through AoR and LA.}
    \label{fig6}
\end{figure}

Aortic Valve – The AoR Diam and LA Dim indicators serve to observe the state of AV. AoR Diam is relatively simpler to be examined, as the sequence image can intuitively show a wall movement. In particular, the character of parallel walls in the aorta moving anteriorly in systole and posteriorly in diastole would not change in dimension in either state. The indicators, therefore, can be measured at any place of the detected aortic root mask. Conversely, LA-Dim is generally measured at end-systole, where the LA mask volume is maximum at each period. To do that, this work makes this locating even easier by directly capturing the topmost point among the mask. This method involves finding the contour coordinates by topological algorithm \cite{suzuki1985topological}. A collective coordinate from FindCountours can be represented as: 
\useshortskip
\begin{flalign}
    \label{equ8}
    c_{LA, m}=\left ( x_{LA, m}, y_{LA, m} \right )=FinedCounters\left ( Mask_{LA} \right ),&&
\end{flalign}
\useshortskip
\begin{flalign}
    \label{equ9}
    C_{LA_{*}}=\max_{-x,y}c_{LA, m},&&
\end{flalign}
which $\left ( x_{LA, m}, y_{LA, m} \right )$ stands a mask contour coordinate C in the x-axis and y-axis, and $\max_{-x,y}c_{LA, m}$ is the operation for finding the topmost point by taking the maximum of converted coordinates in $(-x,y)$. Thus the topmost one should be the first point in the clockwise order. Then, by finding $C_{LA_{*}}$ in ${Mask}_{LA}$, going upwards and downwards pixel-wise obtains the total number of vertical pixels of AoR-Diam and LA-Dim as (\ref{equ10}) and (\ref{equ11}). Finally, multiply these numbers with the echocardiogram image $Scale$, in (\ref{equ12}) and (\ref{equ13}), to get the actual length. The $Scale$ represents the ratio between actual $cm$ compared with the number of pixels in an image's high, pixel-to-cm ratio in \ref{sec3.2}. Please see the demonstration from (a) to (b) in Fig. \ref{fig6} and the equations below. 
\useshortskip
\begin{flalign}
    \label{equ10}
    C_{AoR}=Sign\left({Mask}_{AoR}\left(C_{{LA}_\ast}\right)\uparrow\right),&&
\end{flalign}
\useshortskip
\begin{flalign}
    \label{equ11}
    C_{LA}=Sign\left({Mask}_{AoR}\left(C_{{LA}_\ast}\right)\downarrow\right),&&
\end{flalign}
\useshortskip
\begin{flalign}
    \label{equ12}
    AoR-Diam=Scale\times\left(C_{AoR}\left(y\right)-C_{{LA}_\ast}\left(y\right)\right),&&
\end{flalign}
\useshortskip
\begin{flalign}
    \label{equ13}
    LA-Dim=Scale\times\left(C_{{LA}_\ast}\left(y\right)-C_{LA}\left(y\right)\right).&&
\end{flalign}

Left Ventricle – Examining indicators of LV involves carefully locating where systole and diastole happen, although MENN chooses to use an average value. Systole typically indicates the smallest volume, and diastole is the opposite. The same AV strategy can easily gain systole moment to get the most point in LVPW (\ref{equ14}) to (\ref{equ15}). By scanning up and down, we can sequentially extract LVIDs, IVSs, and LVPWs pixels multiplying with $Scale$, as the following equations (\ref{equ16})-(\ref{equ21}):
\useshortskip
\begin{flalign}
\begin{split}
    \label{equ14}
    c_{LVPW, m}=\left ( x_{LVPW, m}, y_{LVPW, m} \right )\\=FinedCounters\left ( Mask_{LVPW} \right ),
\end{split}&&
\end{flalign}
\useshortskip
\begin{flalign}
    \label{equ15}
    C_{LVPW_{\sigma }}=\max_{-x,y}c_{LVPW, m},&&
\end{flalign}
\useshortskip
\begin{flalign}
    \label{equ16}
    C_{LVIDs}=Sign\left({Mask}_{background}\left(C_{{LVPW}_\sigma}\right)\uparrow\right),&&
\end{flalign}
\useshortskip
\begin{flalign}
    \label{equ17}
    C_{IVSs}=Sign\left({Mask}_{IVS}\left(C_{LVIDs}\right)\uparrow\right),&&
\end{flalign}
\useshortskip
\begin{flalign}
    \label{equ18}
    C_{LVPWs}=Sign\left({Mask}_{LVPW}\left(C_{{LVPW}_\sigma}\right)\downarrow\right),&&
\end{flalign}
\useshortskip
\begin{flalign}
    \label{equ19}
    LVIDs=Scale\times\left(C_{LVIDs}\left(y\right)-C_{{LVPW}_\sigma}\left(y\right)\right),&&
\end{flalign}
\useshortskip
\begin{flalign}
    \label{equ20}
    IVSs=Scale\times\left(C_{IVSs}\left(y\right)-C_{LVIDs}\left(y\right)\right),&&
\end{flalign}
\useshortskip
\begin{flalign}
    \label{equ21}
    LVPWs=Scale\times\left(C_{{LVPW}_\sigma}\left(y\right)-C_{LVPWs}\left(y\right)\right).&&
\end{flalign}
Diastole moment, however, needs some tricky methods to finish. Firstly, find the diastole point, which creates the biggest volume lying in one of the defect points among a convex hull, so structuring the convex hull connection among the mask by Sklansky's algorithm \cite{sklansky1982finding} is applied.
\useshortskip
\begin{flalign}
\begin{split}
    \label{equ22}
    c_{LVPW,m}^H=\left(x_{LVPW,m}^H,y_{LVPW,m}^H\ \right)\\=ConvexHull\left({Mask}_{LVPW}\right).
\end{split}&&
\end{flalign}
With the contour points and hulls determined upon $c_{LVPW,m}$, Secondly, defect points among hulls can be extracted by calculating the maximum distance in each hull, (\ref{equ23}). And then, not to take the defect points in the bottom of the mask as the candidates, these points are excluded by only picking the upper defect points with the biggest distance among all the defect points as the diastole point, (\ref{equ24}). The operation in (\ref{equ24}) can also contribute to saving time by ignoring the unwanted area in the bottom part of the mask. 
\useshortskip
\begin{flalign}
    \begin{split}
    \label{equ23}
        Defects=\left({starts}_{x,y},{ends}_{x,y},{defects}_{x,y},distances\right)\\=FindDefects\left(c_{LVPW,m}\ ,\ c_{LVPW,m}^H\right).
    \end{split}&&
\end{flalign}
\useshortskip
\begin{flalign}
    \label{equ24}
    C_{{LVPW}_\delta}=\max_{sign\left({Defects(ends}_x-{starts}_x)\right)\in+}{Defects\left(distances\right)}.&&
\end{flalign}
Finally, following the same strategy in scanning up and down pixels as IVSs, LVIDs, and LVPWs in (\ref{equ16}) - (\ref{equ18}), the equations for getting the diastole points of IVSd, LVIDd, and LVPWd along with multiplying $Scale$ are as below:
\begin{figure}[!t]
    \centering
    \includegraphics[width=1.0\columnwidth]{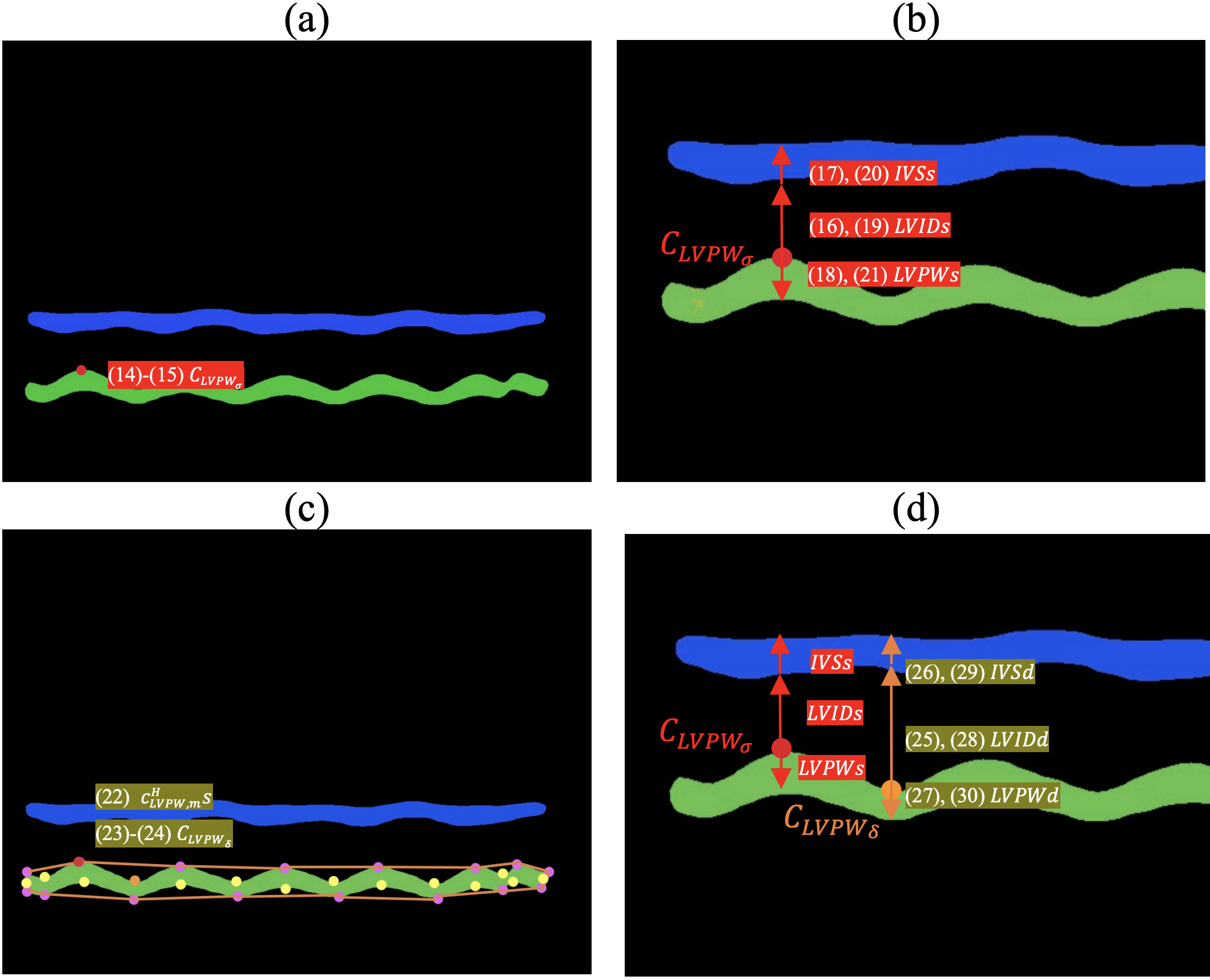}
    \caption{LV indicators measurement. (a) finds the topmost coordinate in LVPW; (b) searches up and down to get LVIDs, IVSs, and LVPWs; (c) constructs the convex hull with peak points in purple, defects in yellow, and connection line in light orange. And then, pick the one with the largest distance from the defect to the connection line in orange. (d) follows the same strategy to search up and down, based on the finding at (c), and gets LVIDd, IVSd, and LVPWd.}
    \label{fig7}
\end{figure}

\useshortskip
\begin{flalign}
    \label{equ25}
    C_{LVIDd}=Sign\left({Mask}_{background}\left(C_{{LVPW}_\delta}\right)\uparrow\right),&&
\end{flalign}
\useshortskip
\begin{flalign}
    \label{equ26}
    C_{IVSd}=Sign\left({Mask}_{IVS}\left(C_{LVIDd}\right)\uparrow\right),&&
\end{flalign}
\useshortskip
\begin{flalign}
    \label{equ27}
    C_{LVPWd}=Sign\left({Mask}_{LVPW}\left(C_{{LVPW}_\delta}\right)\downarrow\right),&&
\end{flalign}
\useshortskip
\begin{flalign}
    \label{equ28}
    LVIDd=Scale\times\left(C_{LVIDd}\left(y\right)-C_{{LVPW}_\delta}\left(y\right)\right),&&
\end{flalign}
\useshortskip
\begin{flalign}
    \label{equ29}
    IVSd=Scale\times\left(C_{IVSd}\left(y\right)-C_{LVIDd}\left(y\right)\right),&&
\end{flalign}
\useshortskip
\begin{flalign}
    \label{equ30}
    LVPWd=Scale\times\left(C_{{LVPW}_\delta}\left(y\right)-C_{LVPWd}\left(y\right)\right).&&
\end{flalign}

With the well-explained procedure, the demonstration of LV can be seen in Fig. \ref{fig7}. The whole process is organized as the algorithm of AMEM.

\begin{center}
    \includegraphics[width=0.5\columnwidth]{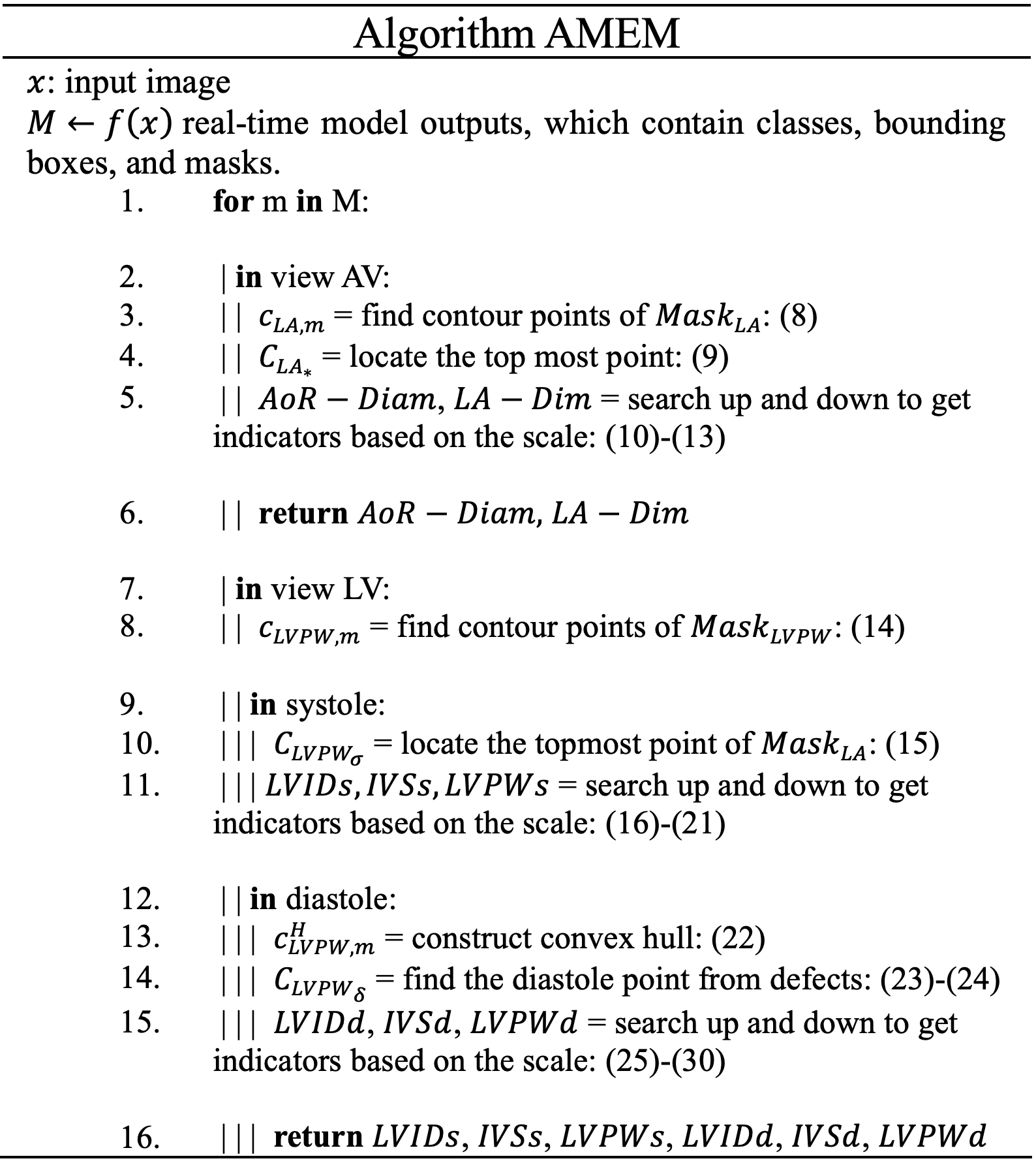}
\end{center}

\section{Experiment}
\label{sec5}
A series of evaluations are conducted to prove the capabilities of the proposed methods, from ablation studies toward panel attention to the real-world indicator bias evaluation based on the provided dataset. Especially, the COCO matric in mean average precision (mAP) consisting of different predict confidence in average precisions from 50 to 95 in every five steps is applied. The metric is a more complex and throughout benchmark, as each step threshold represents a hesitate (or confidence) index toward an object, compared with naïve indexes, such as the mean of intersection over union (IoU) and DICE (an IoU variant). Lastly, we opt for $>$24 FPS ($<$0.042 sec) as the real-time standard. The details of the specific experiment settings and the discussion of our proposed methods are shown in each sub-section below.

\begin{table*}[!t]
\centering
\caption{\label{tab3}\\PASCAL 2012 SBD results.}
\resizebox{\textwidth}{!}{%
\begin{tabular}{lllllllll}
\hline
\# & Work                                                         & Backbone                                                                & Avg-mAP$\uparrow$ & Mask-mAP$\uparrow$ & Box-mAP$\uparrow$ & FPS (RTX 4090)$\uparrow$ & FLOPs (G)$\downarrow$ & Size (M)$\downarrow$ \\ \hline
1  & YOLACT                                                       & ResNet50                                                                & 35.73             & 35.12              & 36.65             & 81.11                    & 48.26                 & 30.41                \\
2  & maYOLACT                                                     & ResNet50                                                                & 37.39             & 37.27              & 37.50             & \textbf{81.27}           & 48.26                 & 30.41                \\
3  & maYOLACT                                                     & CSP-DarkNet                                                             & 40.29             & 39.75              & 40.83             & 80.36                    & 61.73                 & 45.44                \\
4  & maYOLACT                                                     & \begin{tabular}[c]{@{}l@{}}SwinTransformer\\ (Tiny)\end{tabular}        & 21.30             & 21.16              & 21.44             & 71.88                    & \textbf{24.96}        & 33.85                \\
5  & \begin{tabular}[c]{@{}l@{}}RTMDet\\ (ins-X)$^1$\end{tabular} & CSP-DarkNet                                                             & 32.39             & 30.56              & 34.22             & 68.42                    & 93.87                 & 83.33                \\
6  & \begin{tabular}[c]{@{}l@{}}RAMEM\end{tabular}    & \begin{tabular}[c]{@{}l@{}}UPANet64 V2\\ (Vanilla)\end{tabular}         & 41.50             & 41.25              & 41.75             & 68.00                    & 73.12                 & 28.73                \\
7  & \begin{tabular}[c]{@{}l@{}}RAMEM\end{tabular}    & \begin{tabular}[c]{@{}l@{}}UPANet80 V2\\ (Vanilla)\end{tabular}         & 41.72             & 41.51              & 41.93             & 64.50                    & 100.55                & 40.32                \\ \hline
8  & \begin{tabular}[c]{@{}l@{}}RAMEM\end{tabular}    & \begin{tabular}[c]{@{}l@{}}UPANet80 V2\\ (GC, Att 1st$^2$)\end{tabular}     & 42.24             & 42.20              & 42.47             & 63.91                    & 98.32                 & 39.30                \\
9  & \begin{tabular}[c]{@{}l@{}}RAMEM\end{tabular}    & \begin{tabular}[c]{@{}l@{}}UPANet80 V2\\ (EA, Att 1st)\end{tabular}     & 41.86             & 41.65              & 42.07             & 64.50                    & 98.31                 & 39.30                \\
10 & \begin{tabular}[c]{@{}l@{}}RAMEM\end{tabular}    & \begin{tabular}[c]{@{}l@{}}UPANet80 V2\\ (A2, Att 1st)\end{tabular}     & 41.26             & 41.7               & 42.62             & 53.25                    & 101.87                & 40.50                \\
11 & \begin{tabular}[c]{@{}l@{}}RAMEM\end{tabular}    & \begin{tabular}[c]{@{}l@{}}UPANet80 V2\\ (NL, Att 1st)\end{tabular}     & -                 & -                  & -                 & -                        & 102.58                & 40.77                \\ \hline
12 & \begin{tabular}[c]{@{}l@{}}RAMEM\end{tabular}    & \begin{tabular}[c]{@{}l@{}}UPANet80 V2\\ (GC, Ds 1st$^3$)\end{tabular}      & 42.35             & 42.20              & 42.51             & 62.86                    & 98.32                 & 39.30                \\
13 & \begin{tabular}[c]{@{}l@{}}RAMEM\end{tabular}    & \begin{tabular}[c]{@{}l@{}}UPANet80 V2\\ (EA, Ds 1st)\end{tabular}      & 41.33             & 41.04              & 41.62             & 64.28                    & 98.31                 & 39.30                \\
14 & \begin{tabular}[c]{@{}l@{}}RAMEM\end{tabular}    & \begin{tabular}[c]{@{}l@{}}UPANet80 V2\\ (A2, Ds 1st)\end{tabular}      & 41.13             & 40.92              & 41.33             & 60.34                    & 99.20                 & 40.50                \\
15 & \begin{tabular}[c]{@{}l@{}}RAMEM\end{tabular}    & \begin{tabular}[c]{@{}l@{}}UPANet80 V2\\ (NL, Ds 1st)\end{tabular}      & 42.24             & 42.17              & 42.30             & 61.64                    & 99.02                 & 40.23                \\ \hline
\rowcolor[HTML]{EFEFEF} 
16 & maYOLACT                                                     & \begin{tabular}[c]{@{}l@{}}ResNet50\\ (Panel attention)\end{tabular}    & 38.92             & 38.71              & 39.15             & 64.56                    & 91.88                 & 74.59$^4$            \\
\rowcolor[HTML]{EFEFEF} 
17 & \begin{tabular}[c]{@{}l@{}}RAMEM\end{tabular}    & \begin{tabular}[c]{@{}l@{}}UPANet64 V2\\ (Panel attention)\end{tabular} & 41.74             & 41.44              & 42.04             & 74.22                    & 73.12                 & \textbf{28.73}       \\
\rowcolor[HTML]{EFEFEF} 
18 & \begin{tabular}[c]{@{}l@{}}RAMEM\end{tabular}    & \begin{tabular}[c]{@{}l@{}}UPANet80 V2\\ (Panel attention)\end{tabular} & \textbf{42.69}    & \textbf{42.42}     & \textbf{42.96}    & 60.93                    & 100.85                & 40.32                \\ \hline
\multicolumn{9}{@{}l}{$^1$RTMDet-ins X: Considering the main contributions of RTMDet laying in backbone, neck, and label assignment, we only apply CSPDarkNet X +}\\
\multicolumn{9}{@{}l}{PAFPN + soft SimOTA. The rest of the modules remain as YOLACTs for a fair comparison.}\\
\multicolumn{9}{@{}l}{$^2$Att 1st: Attention before down-sampling}\\
\multicolumn{9}{@{}l}{$^3$Ds 1st: Attention after down-sampling}\\
\multicolumn{9}{@{}l}{$^4$ResNet50-Panel: the significant parameter increasing result from the needed significant channel number that makes Panel attention grows. Based}\\
\multicolumn{9}{@{}l}{on the observation in \#11 \& 15, NL will be the worst scene.}
\end{tabular}%
}
\end{table*}

\subsection{PASCAL 2012 SBD}
To evaluate the effect of the proposed Panel attention and modules among it, the existing backbones testing experiment is conducted firstly on PASCAL 2012 SBD, which is a standard open dataset for evaluating an object detection model. It contains 20 categories with roughly 8k images in training data and 2k in testing data. As this work focuses on real-time instance segmentation, the simulation in a mature scheme, YOLACT, is followed. However, as in the M-mode echocardiogram, there is no trained weight to finetune, the end-to-end training evaluation is conducted to mimic the current situation in echocardiography. Also, to make the comparison process faster, the training is set to 64k iteration, which is evaluated at the first end of the epoch 60th of the learning rate schedule in YOLACT. The rest of the setting remains unchanged. In Table \ref{tab3}, The complete maYOLACT and RTMDet are also tested in this simulation, with input shape in $3\times544\times544$. Note that the size in Table \ref{tab3} indicates the whole model’s parameter, where different backbones will affect the FPN input channel number accordingly. The results of mask and box only reported mAP considering the aesthetic, added another index of average mAP from mask and box in Avg-mAP, and the general priority of index in order is: Avg-mAP, Mask-mAP, Box-mAP, FPS, FLOPs, and then size. 

\begin{table}[]
\centering
\caption{\label{tab4}\\Attention ablation study.}
\resizebox{\columnwidth}{!}{%
\begin{tabular}{llllll}
\hline
\#  & \begin{tabular}[c]{@{}l@{}}Backbone\\ UPANet80 V2\\ (Without attention)\end{tabular} & Local Attention & Global Attention & Panel Attention & Avg-mAP$\uparrow$ \\ \hline
1 & -                                                                                  & -               & -                & -               & 37.39             \\
2 & $\checkmark$                                                                                & -               & -                & -               & 41.72             \\
3 & $\checkmark$                                                                                & $\checkmark$             & -                & -               & 41.85             \\
4 & $\checkmark$                                                                                & -               & $\checkmark$              & -               & 42.51             \\
\rowcolor[HTML]{F2F2F2} 
5 & $\checkmark$                                                                                & $\checkmark$           & $\checkmark$            & =$\checkmark^1$           & \textbf{42.69}    \\ \hline
\multicolumn{6}{@{}l}{$^1$Panel attention: local attention + global attention, both local and global attention having lossless}\\
\multicolumn{6}{@{}l}{operation involving as attention in down-sampling, please refer Section \ref{sec4.1.1}}
\end{tabular}%
}
\end{table}

\begin{table*}[!t]
\centering
\caption{\label{tab5}\\MEIS results.}
\resizebox{\textwidth}{!}{%
\begin{tabular}{lllllllll}
\hline
\# & Work                                                      & Backbone                                                         & Avg-mAP$\uparrow$ & Mask-mAP$\uparrow$ & Box-mAP$\uparrow$ & FPS (RTX 4090)$\uparrow$ & FLOPs (G)$\downarrow$ & Size (M)$\downarrow$ \\ \hline
1  & YOLACT                                                    & ResNet50                                                         & -                 & -                  & -                 & -                       & 48.26                & \textbf{30.38}      \\
2  & maYOLACT                                                  & ResNet50                                                         & 46.29             & 42.99              & 49.59             & 36.13                   & 48.26                & \textbf{30.38}      \\
3  & maYOLACT                                                  & CSP-DarkNet                                                      & 46.18             & 42.11              & 50.25             & 34.65                   & 61.73                & 45.40               \\
4  & maYOLACT                                                  & \begin{tabular}[c]{@{}l@{}}SwinTransformer\\ (Tiny)\end{tabular} & 39.55             & 38.10              & 40.99             & 32.99                   & \textbf{24.96}       & 33.80               \\
5  & \begin{tabular}[c]{@{}l@{}}RTMDet\\ (ins-X)\end{tabular}  & CSP-DarkNet                                                      & 44.75             & 40.18              & 49.33             & \textbf{56.22}          & 93.87                & 80.37               \\
\rowcolor[HTML]{EFEFEF} 
6  & \begin{tabular}[c]{@{}l@{}}RAMEM\end{tabular} & UPANet80 V2                                                      & \textbf{47.15}    & \textbf{43.09}     & \textbf{51.20}    & 52.22                   & 100.85               & 40.28               \\ \hline
\end{tabular}%
}
\end{table*}

\textbf{UPANets V2 outperforms other backbones in mAPs} (Table \ref{tab3} \#1 - \#7) – Apart from applying SwinTransformer-Tiny in maYOLACT, this group contains the existing backbones in RIS: ResNets (in [ma]YOLACT[++], SipMask, BlendMask, and SOLOV2) and CSPDarkNets (in RTMDet). We firstly examine the capability of vanilla UPANets V2 (without Panel attention) with mentioned backbones in the YOLACT scheme. ResNets \#1 - \#2 have superior performance in complexity, but they are compromised in mAPs. In fact, the FPS performance is also affected by the mAPs because a hesitated object detection will increase the suppressing candidates in NMS, which causes the FPS to degrade despite the low FLOPs. An opposite scene can be seen in \#2, with firm prediction boosting FPS. Another example is SwinTransformer \#4. Despite holding the least FLOPs and more parameters, SwinTransformer using the patch in global attention does not progress much regarding mAP. Conversely, the CSP-DarkNet \#3 \& \#5 possesses excellent balance in these spheres. The same trend can be witnessed in UPANets V2 \#6 - \#7 with better Avg-mAP results. Therefore, the following comparison will focus on the NL variants in different scenes upon UPANets V2 viewing a better capability of UPANets V2 than other backbones.

\textbf{Att 1st makes NL inefficiency with mAPs improvement} (Table \ref{tab3} \#8 - \#11) – This second group follows the workflow as (a) in Fig. \ref{fig2}, using different NL variants in stage {2, 3, 4} as general. The candidates across GCNet (GC), EANet (EA), A2, and NL use the default setting. \#8 in GC has outshined others with the least size, speedy FPS, and mAPs, which aligns with the GCNet results \cite{cao2019gcnet}. However, because of the unbearable complexity caused by spatial, \#11 in NL was inoperable in a 24G GPU. That issue is bypassed by A2 \#10 in operating toward a different dimension order: the channel and then spatial, but the performance is compromised. 

\textbf{Ds 1st makes NL efficiency with information losing and suboptimal mAPs} (Table \ref{tab3} \#12 - \#15) – Following the workflow as (b) in Fig. \ref{fig2}. This group has experienced the same scene as the last group (Att 1st). Still, NL \#15 outperforms EA \#13 and A2 \#14, which could indicate that using simplified NL might lose some vital features to supplement complexity. Moreover, this tradeoff even places simplified NL in a vulnerable position compared with pure CNN, comparing with \#6 and \#7.

\textbf{Panel attention is the balance between efficiency and mAPs} (Table \ref{tab3} \#16 - \#18 \& Table \ref{tab4} \#1 - \#5 ) – The ablation study in Table \ref{tab4} can be divided into three aspects: \#3 local attention, \#4 global attention,  and \#5 panel attention as a whole (local + global attention). Table \ref{tab4} \#1 aligns to  Table \ref{tab3} \#2 and Table \ref{tab4} \#2 to  Table \ref{tab3} \#7. Local attention can contribute to mAPs because the dilated depth-wise CNN expands the receptive field in dilated effect. On the other hand, global attention shows already surpassing results, compared with the candidates in Table \ref{tab3} \#8 and \#15. This surpassing also solidifies that having a global receptive field is crucial. By combining local-to-global attention into panel attention, UPANets64 V2 has been the lightest backbone among NLs, which indicates a small parameter tradeoff. Most importantly, the best in Table \ref{tab3} \#18 shows that including desiring merits from CNNs and NL is helpful, the same can be seen in Table \ref{tab3} \#16 - \# 17. That is, a better balance is reached between mAPs and other efficiency indexes. 

\begin{figure*}[h]
    \centering
    \begin{tabular}{cccc>{\columncolor[HTML]{F2F2F2}}c}
        \tiny Input & \tiny ResNet50 stage2 & \tiny CSP-DarkNet stage2 & \tiny SwimTransformer stage2 & \tiny UPANet80 V2 stage2\\
        \includegraphics[width=0.175\textwidth]{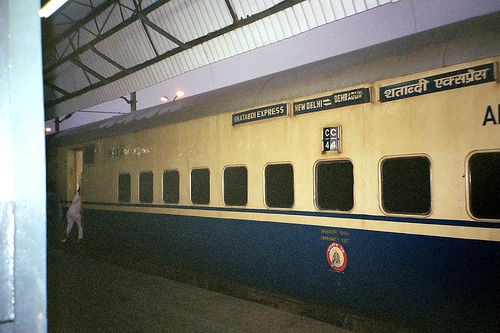} &
        \includegraphics[width=0.175\textwidth]{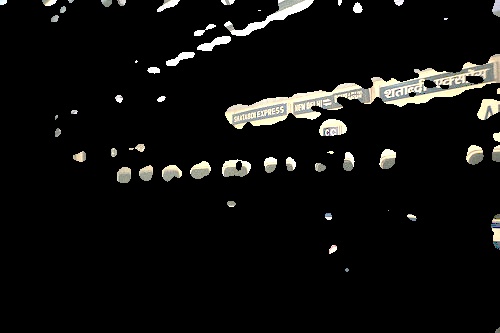} &
        \includegraphics[width=0.175\textwidth]{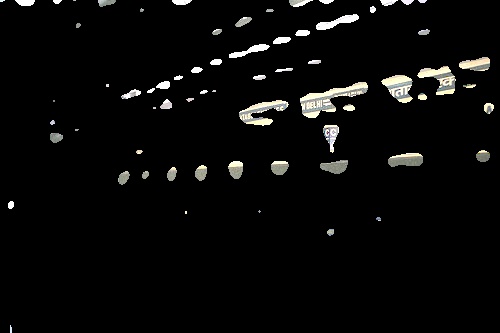} &
        \includegraphics[width=0.175\textwidth]{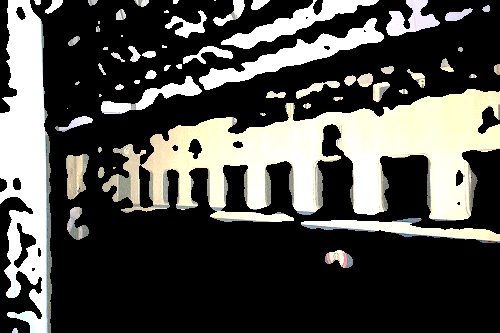} &
        \includegraphics[width=0.175\textwidth]{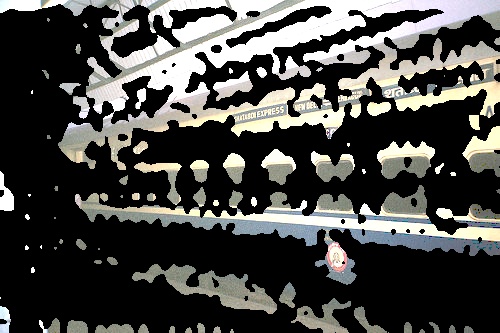}
        \\
        \tiny Ground truth & \tiny ResNet50 stage3 & \tiny CSP-DarkNet stage3 & \tiny SwimTransformer stage3 & \tiny UPANet80 V2 stage3\\
        \includegraphics[width=0.175\textwidth]{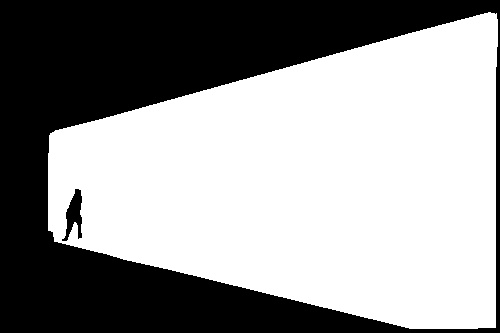} &
        \includegraphics[width=0.175\textwidth]{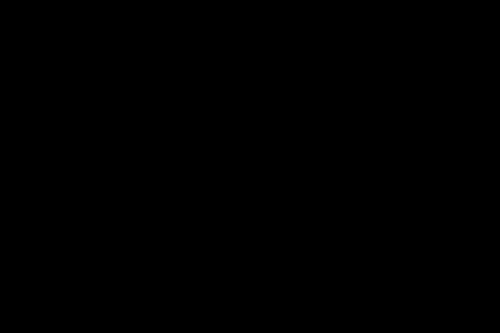} &
        \includegraphics[width=0.175\textwidth]{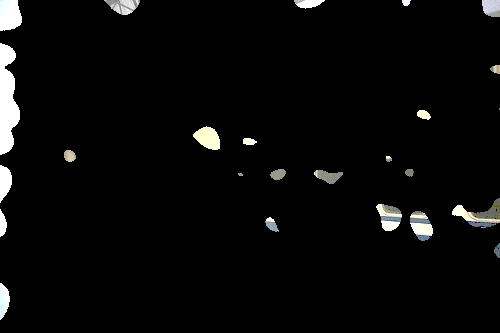} &
        \includegraphics[width=0.175\textwidth]{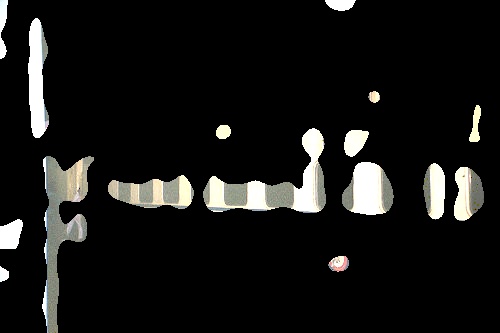} &
        \includegraphics[width=0.175\textwidth]{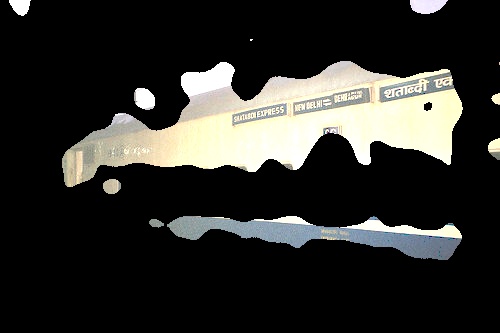}
        \\
    \end{tabular}
\caption{PASCAL sampled feature maps with different backbones. Sampled images only show the responding area where the value is over 0.5 in a [0, 1] range. The responding area generally degrades to under 0.5 in non-global capability backbones, such as ResNet50 and CSP-DarkNet. Global attention backbones of SwimTransformer and UPANet80 V2 can maintain a wider responding area active toward the ground truth.}
\label{fig8}
\end{figure*}

\begin{figure*}[!t]
    \centering
    \begin{tabular}{cccc>{\columncolor[HTML]{F2F2F2}}c}
        \tiny Input & \tiny ResNet50 stage2 & \tiny CSP-DarkNet stage2 & \tiny SwimTransformer stage2 & \tiny UPANet80 V2 stage2\\
        \includegraphics[width=0.175\textwidth]{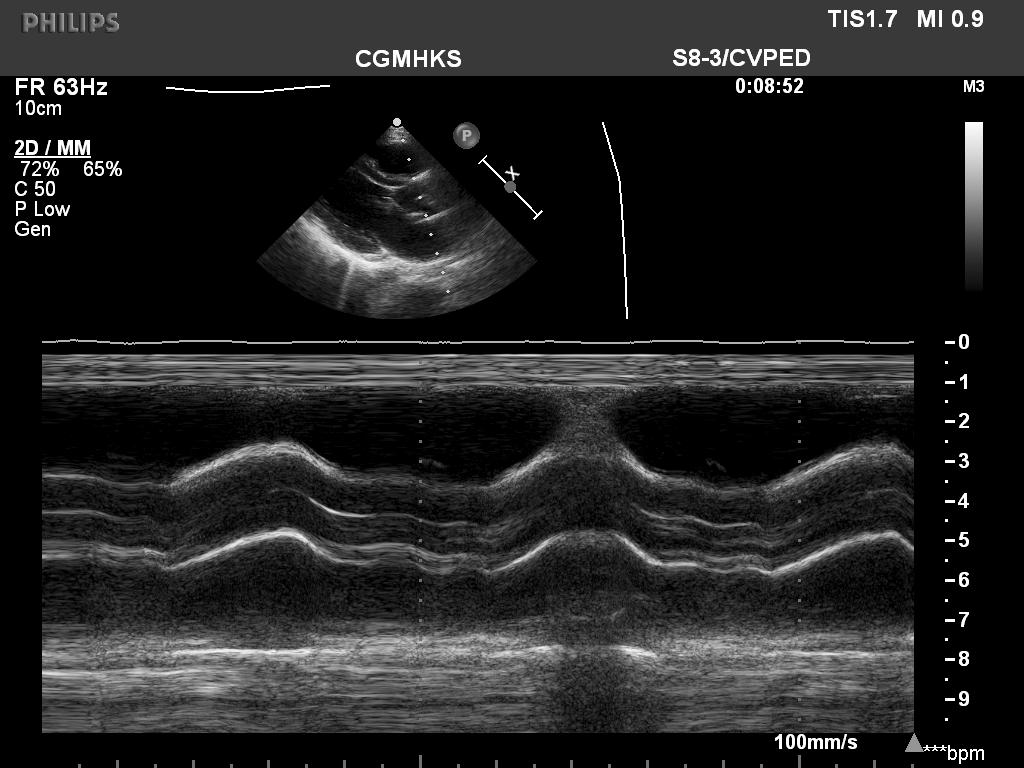} & 
        \includegraphics[width=0.175\textwidth]{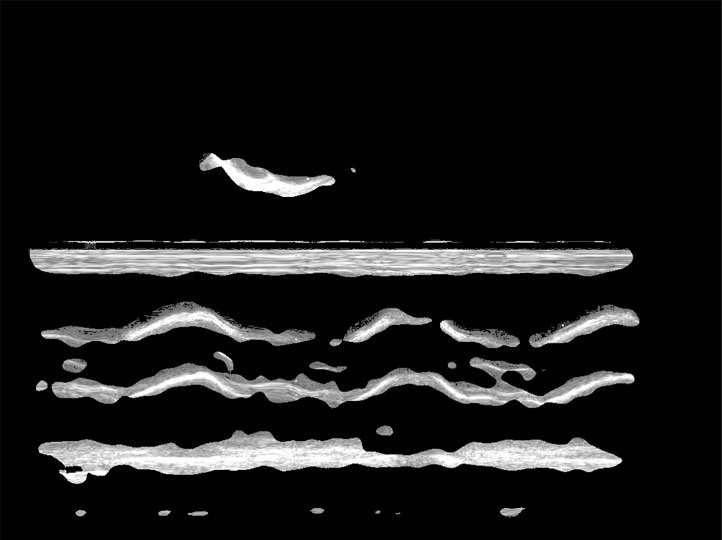} &
        \includegraphics[width=0.175\textwidth]{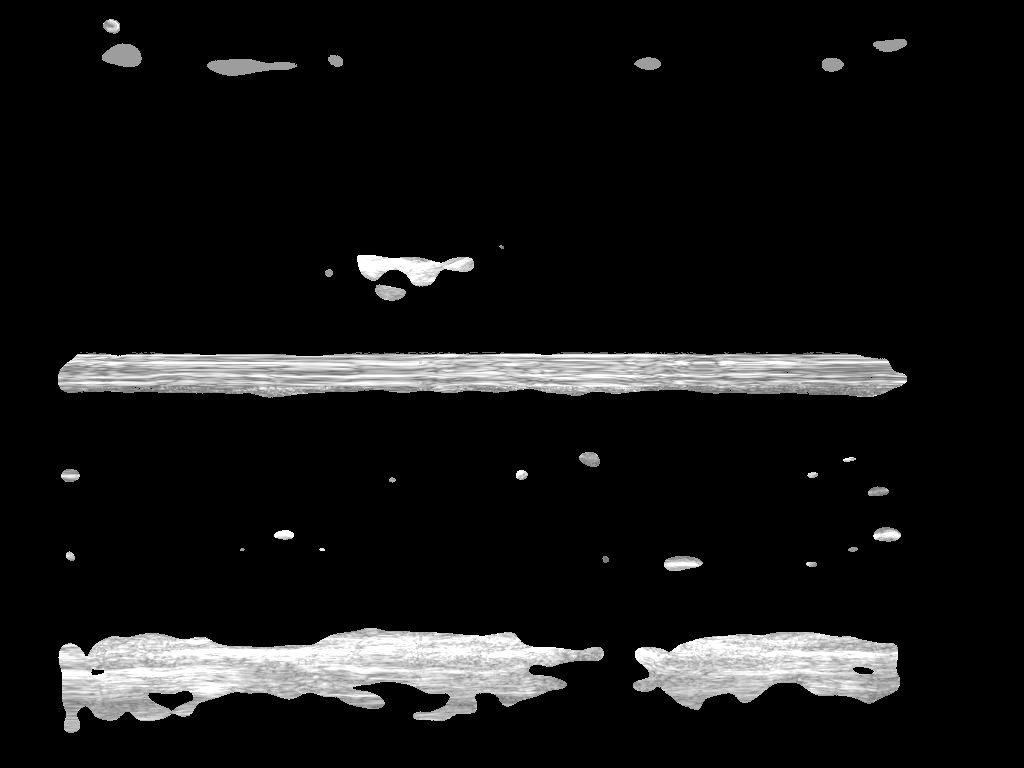} &
        \includegraphics[width=0.175\textwidth]{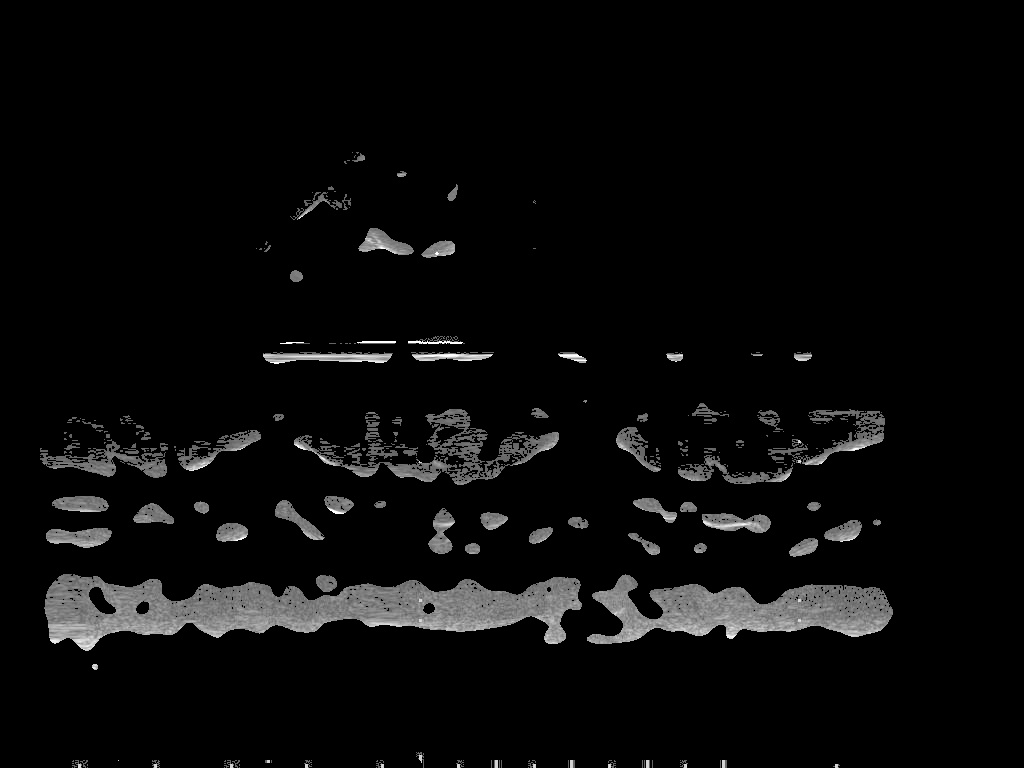} &
        \includegraphics[width=0.175\textwidth]{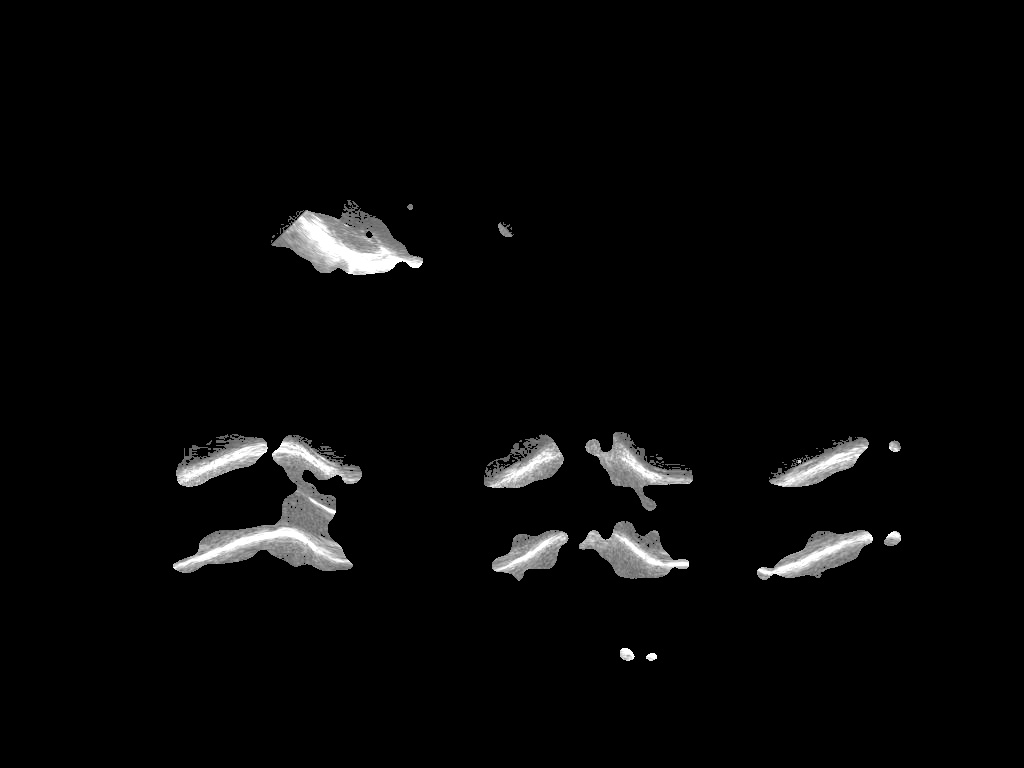}
        \\
        \tiny Ground truth & \tiny ResNet50 stage3 & \tiny CSP-DarkNet stage3 & \tiny SwimTransformer stage3 & \tiny UPANet80 V2 stage3\\
        \includegraphics[width=0.175\textwidth]{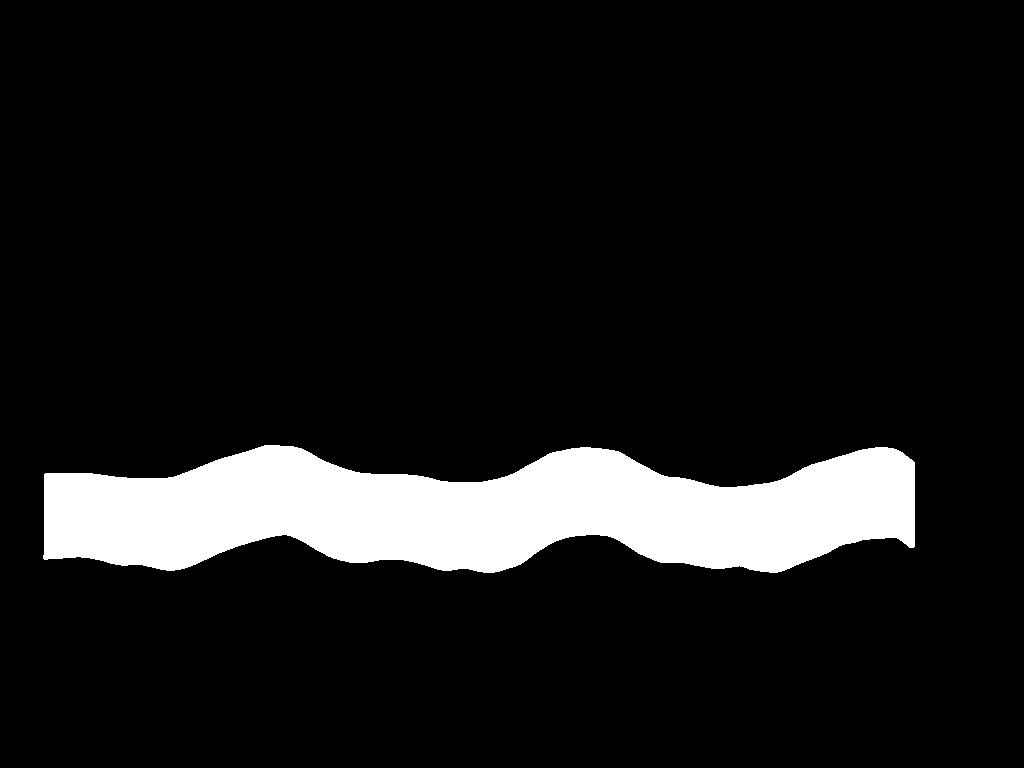} & 
        \includegraphics[width=0.175\textwidth]{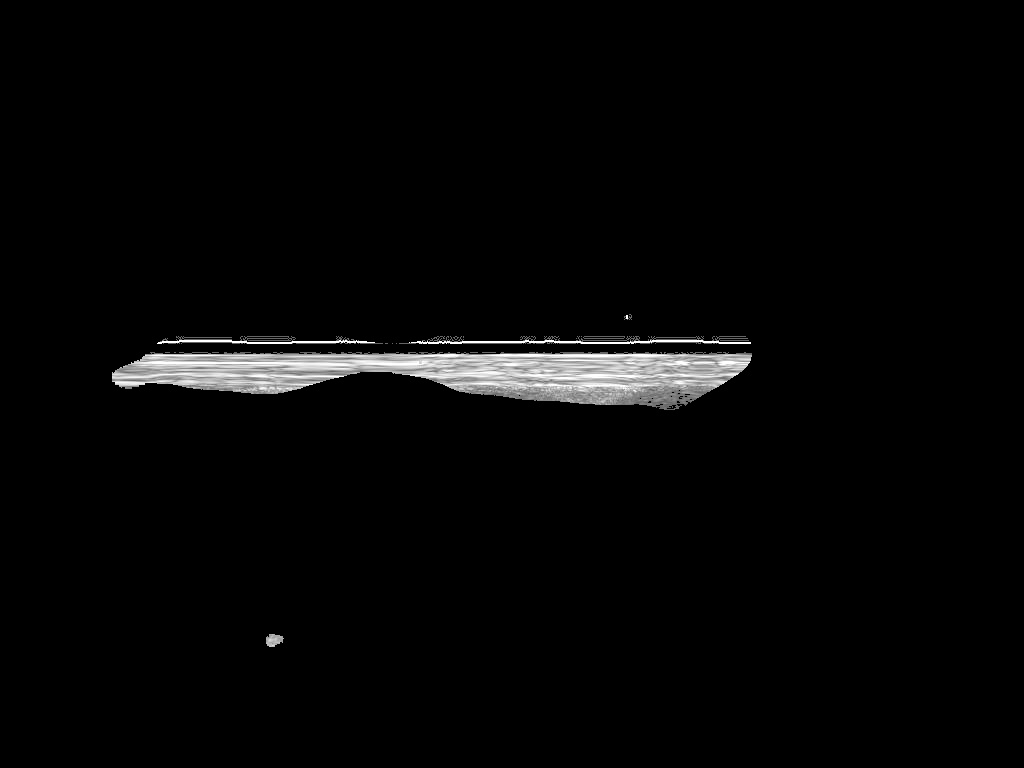} &
        \includegraphics[width=0.175\textwidth]{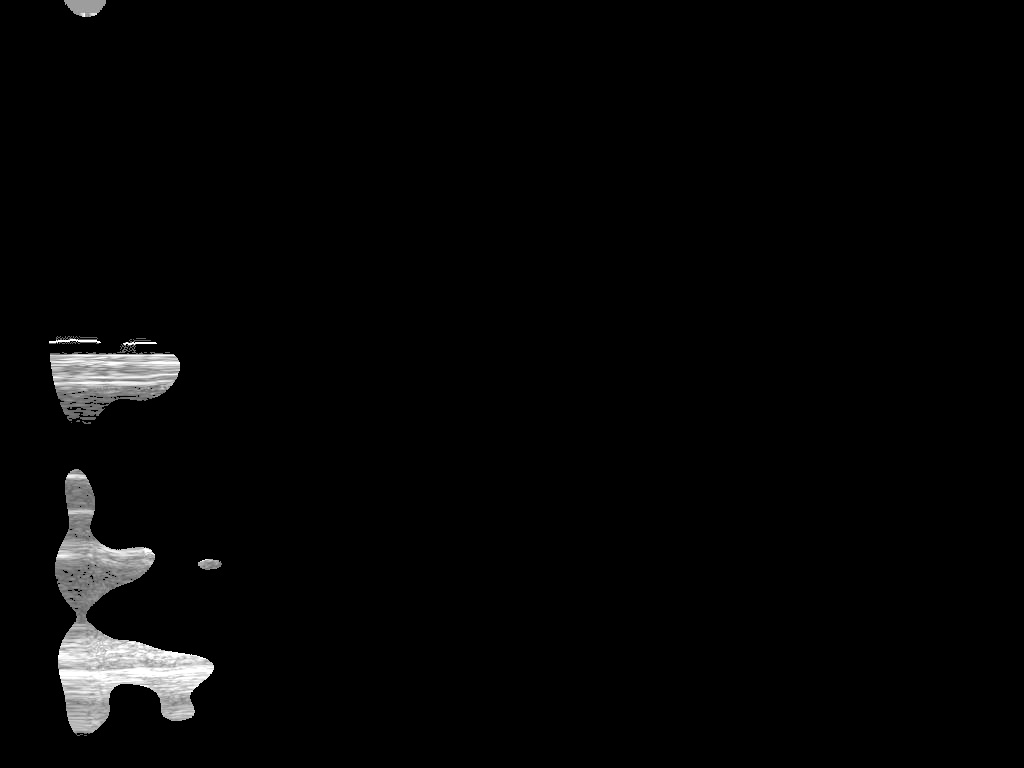} &
        \includegraphics[width=0.175\textwidth]{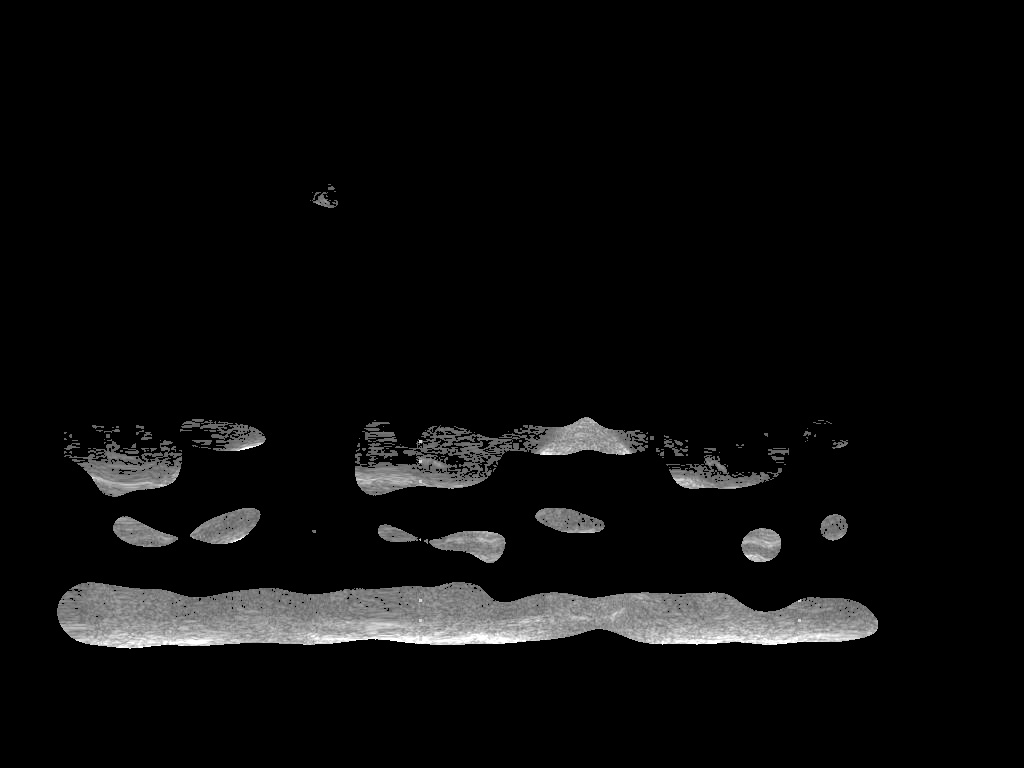} &
        \includegraphics[width=0.175\textwidth]{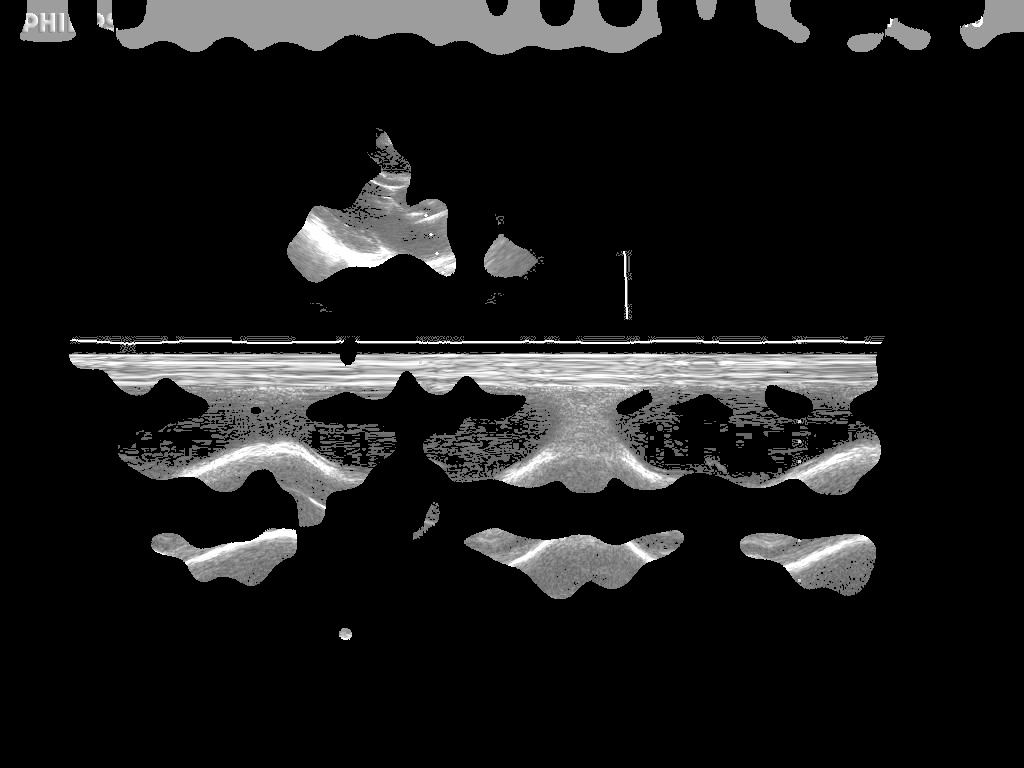}
        \\
    \end{tabular}
\caption{MEIS sampled feature maps with different backbones. By following the same sampling policy, a similar pattern can be witnessed in this figure. A wired responding area occurs in ResNet50 and CSP-DarkNet, which could be caused by a limited receptive field.}
\label{fig9}
\end{figure*}

\subsection{MEIS}
This sub-section inherits the results of \#18 in Table \ref{tab3} PASCAL 2012 SBD as our final proposed model to test on another contribution, the MEIS dataset. By following the same experimental setting as PASCAL 2012 SBD but with 120 epochs, this simulation not only represents the real-world clinic scene but also tests the backbone capability in detecting big objects, which means owning receptive field ability shall perform well. Recorded results do not include measurement times in FPS and FLOPs in this discussion. Please see Table \ref{tab5}.

The results reflect the same pattern as Table \ref{tab3}. From our proposed dataset, the influence of the receptive field among a backbone can be amplified, as most of the objects occupy a great portion of pixels which is what pure CNN-based lacks. Although ResNet50s have gained plausible performance in PASCAL SBD, the accuracy and efficiency are compromised compared with the bigger receptive field backbones. This statement can be observed from RTMDet that applying a big kernel successfully presents exceptional speed in FPS \#5. However, the detection accuracy is still unpair with the backbone equipping non-local attention, the Panel attention in UPANet V2 \#6. In fact, because of the firm prediction in the proposed backbone, the FPS is contributed despite a higher complexity. Therefore, having PASCAL results upon MEIS examination, UPANets V2 has better-detected accuracy, speed, and parameter trade-off. 

\subsection{Attention Map Explanation}
\label{sec5.3}
To get a picture of what attention is learned and the difference from CNN, a feature map explainable method Score-CAM \cite{wang2020score} has been opted. The sampled maps are extracted from four models in the same maYOLACT scheme: ResNet, CSP-DarkNet, SwimTransformer and Panel attention in UPANet80 V2. Also, stages from stage 2 and stage 3 for viewing the effect of the deep are sampled as well.

\textbf{Great receptive field helps to connect to right area} (Fig. \ref{fig8} and Fig. \ref{fig9}) – Observing the sampled responding feature maps from PASCAL in Fig. \ref{fig8} and MEIS in Fig. \ref{fig9}, it surprisingly shows that the “non-local” information is transforming from a receptive field to the relationship toward ground truth under the method of Score-CAM. Another scene in stage 3, deeper responding layers, indicates the focus will narrow to the related area toward the ground truth. However, what CNNs (ResNet and CSP-DarkNet) fail is the responding area degrading or, even worst, vanishing into a blank and irrelevant area. Evidence examples in SwimTransformer and UPANet show a wider response area, indicating the contribution of global attention. The attention maps reflect the performance behaviour in Table \ref{tab4} and Table \ref{tab5}, where the backbones with a big receptive field perform better. 

\begin{table*}[!t]
\centering
\caption{\label{tab6}\\Measurement results on MEIS.}
\resizebox{\textwidth}{!}{%
\begin{tabular}{lllllll
>{\columncolor[HTML]{EFEFEF}}l 
>{\columncolor[HTML]{EFEFEF}}l 
>{\columncolor[HTML]{EFEFEF}}l }
\hline
                                                         & \multicolumn{3}{l}{Humans}                                                                           & \multicolumn{3}{l}{MENN}                                                                                                 & \multicolumn{3}{l}{\cellcolor[HTML]{EFEFEF}AMEM (RAMEM)}                                                                                                                                                                                                       \\
                                & MAE   & MSE   & \begin{tabular}[c]{@{}l@{}}FPS\\ (Time-RTX 4090)\end{tabular}                        & MAE            & MSE            & \begin{tabular}[c]{@{}l@{}}FPS\\ (Time-RTX 4090)\end{tabular}                          & MAE                                                              & MSE                                                              & \begin{tabular}[c]{@{}l@{}}FPS\\ (Time-RTX 4090)\end{tabular}                                                           \\ \hline
AoR-Diam                                                 & 0.568 & 0.441 &                                                                                      & -$^1$              & -              &                                                                                        & \textbf{0.133}                                                   & \textbf{0.027}                                                   & \cellcolor[HTML]{EFEFEF}                                                                                                \\
LA-Dim                                                   & 0.52  & 0.46  &                                                                                      & -              & -              &                                                                                        & \textbf{0.138}                                                   & \textbf{0.035}                                                   & \cellcolor[HTML]{EFEFEF}                                                                                                \\
LVIDd                                                    & 0.783 & 1.177 &                                                                                      & 0.373          & 0.340          &                                                                                        & \textbf{0.165}                                                   & \textbf{0.062}                                                   & \cellcolor[HTML]{EFEFEF}                                                                                                \\
LVPWd                                                    & 0.124 & 0.022 &                                                                                      & 0.130          & 0.028          &                                                                                        & \textbf{0.143}                                                   & \textbf{0.028}                                                   & \cellcolor[HTML]{EFEFEF}                                                                                                \\
IVSd                                                     & 0.27  & 0.099 &                                                                                      & \textbf{0.113} & \textbf{0.018} &                                                                                        & 0.128                                                            & 0.022                                                            & \cellcolor[HTML]{EFEFEF}                                                                                                \\
LVIDs                                                    & 0.651 & 0.653 &                                                                                      & 0.394          & 0.295          &                                                                                        & \textbf{0.161}                                                   & \textbf{0.044}                                                   & \cellcolor[HTML]{EFEFEF}                                                                                                \\
LVPWs                                                    & 0.245 & 0.08  &                                                                                      & \textbf{0.124} & \textbf{0.021} &                                                                                        & 0.190                                                            & 0.049                                                            & \cellcolor[HTML]{EFEFEF}                                                                                                \\
IVSs                                                     & 0.236 & 0.094 & \begin{tabular}[c]{@{}l@{}}-\\ (\textgreater{}60 sec)\end{tabular} & 0.100          & 0.017          & \begin{tabular}[c]{@{}l@{}}13.76\\ ($\approx$0.073 sec)\end{tabular} & \textbf{0.097}                                                   & \textbf{0.017}                                                   & \cellcolor[HTML]{EFEFEF}\textbf{\begin{tabular}[c]{@{}l@{}}24.45\\ ($\approx$0.041 sec)\end{tabular}} \\ \hline
\begin{tabular}[c]{@{}l@{}}Mean\\ (LV only)\end{tabular} & 0.425 & 0.378 &                                                                                      & 0.206          & 0.12           &                                                                                        & \textbf{\begin{tabular}[c]{@{}l@{}}0.144\\ (0.147)\end{tabular}} & \textbf{\begin{tabular}[c]{@{}l@{}}0.036\\ (0.037)\end{tabular}} &                                                                                                                         \\
\begin{tabular}[c]{@{}l@{}}Std\\ (LV only)\end{tabular}  & 0.221 & 0.371 &                                                                                      & 0.126          & 0.14           &                                                                                        & \textbf{\begin{tabular}[c]{@{}l@{}}0.026\\ (0.030)\end{tabular}} & \textbf{\begin{tabular}[c]{@{}l@{}}0.014\\ (0.016)\end{tabular}} &                                                                                                                         \\ \hline
\multicolumn{10}{@{}l}{$^1$MENN does not target toward AV view, so the overall mean and std error only indicate LV view.}\\
\end{tabular}%
}
\end{table*}

\begin{figure*}[!t]
    \centering
    \resizebox{\textwidth}{!}{%
    \begin{tabular}{ccccc>{\columncolor[HTML]{F2F2F2}}c}
        Ground truth & Human candidate1 & MENN UPANet80 V2 & AMEM maYOLACT & AMEM RTMDet & AMEM UPANet80 V2 (RAMEM)\\
        \includegraphics[width=0.3\textwidth]{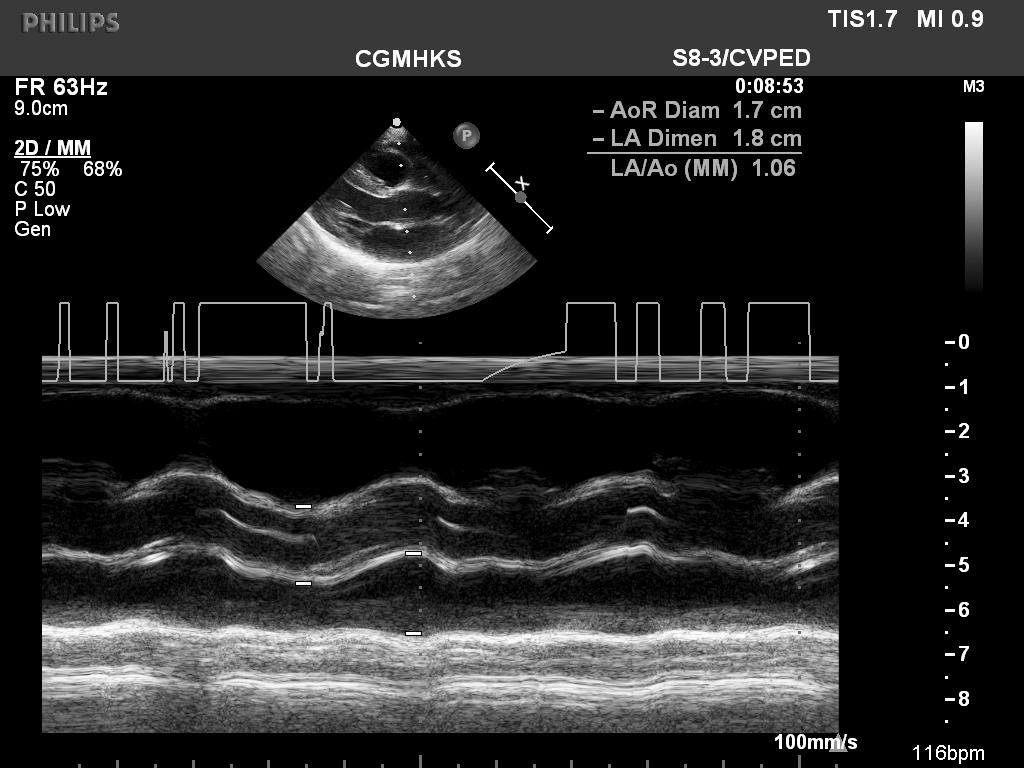} &
        \includegraphics[width=0.3\textwidth]{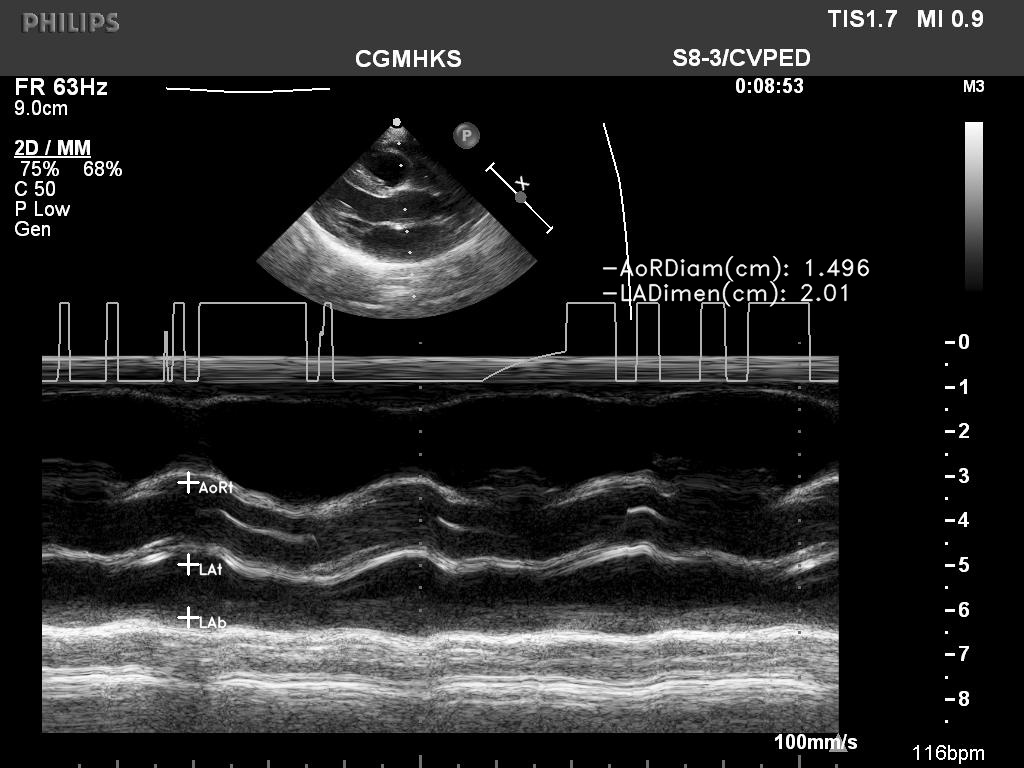} &
        \includegraphics[width=0.3\textwidth]{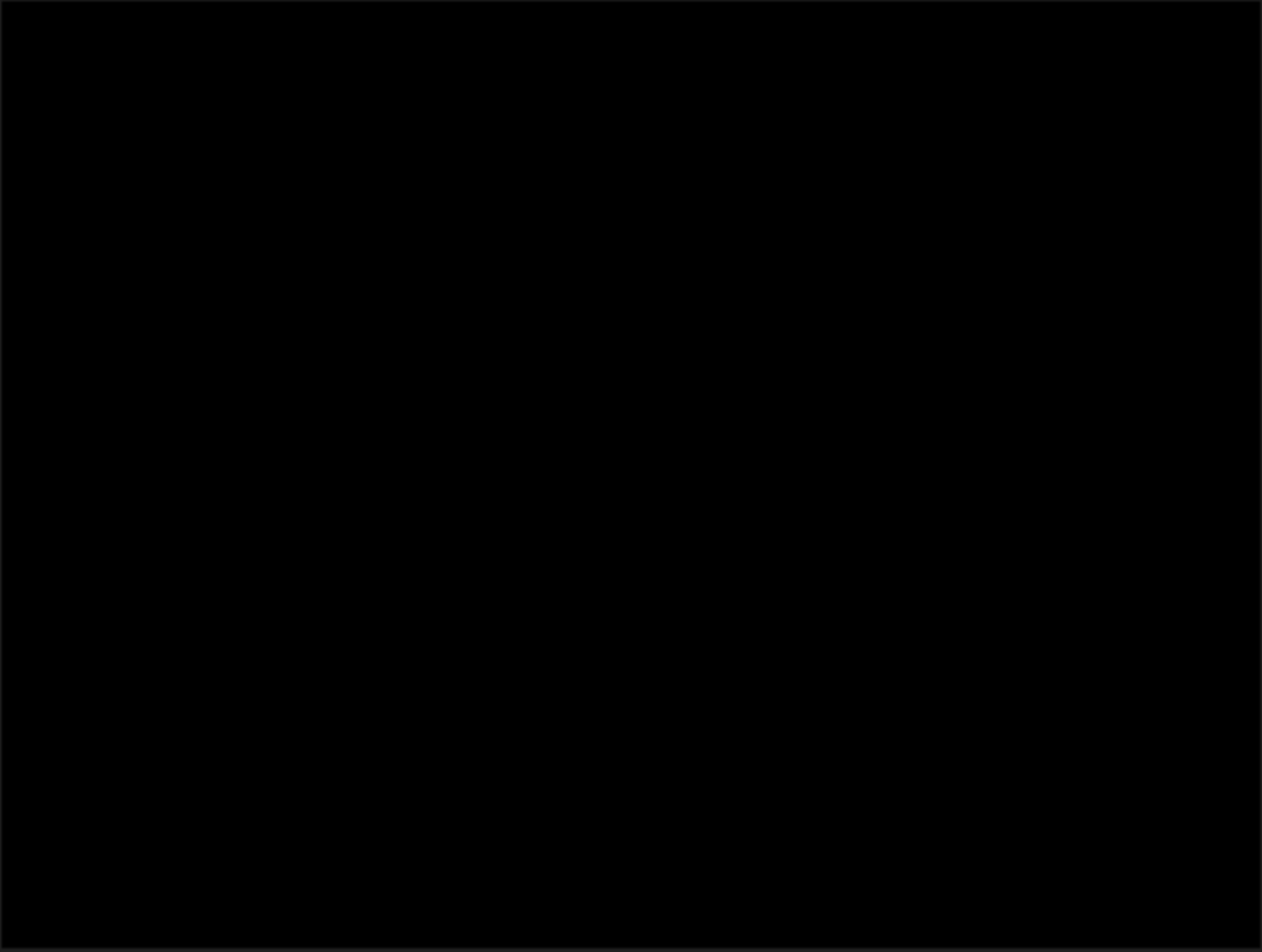} &
        \includegraphics[width=0.3\textwidth]{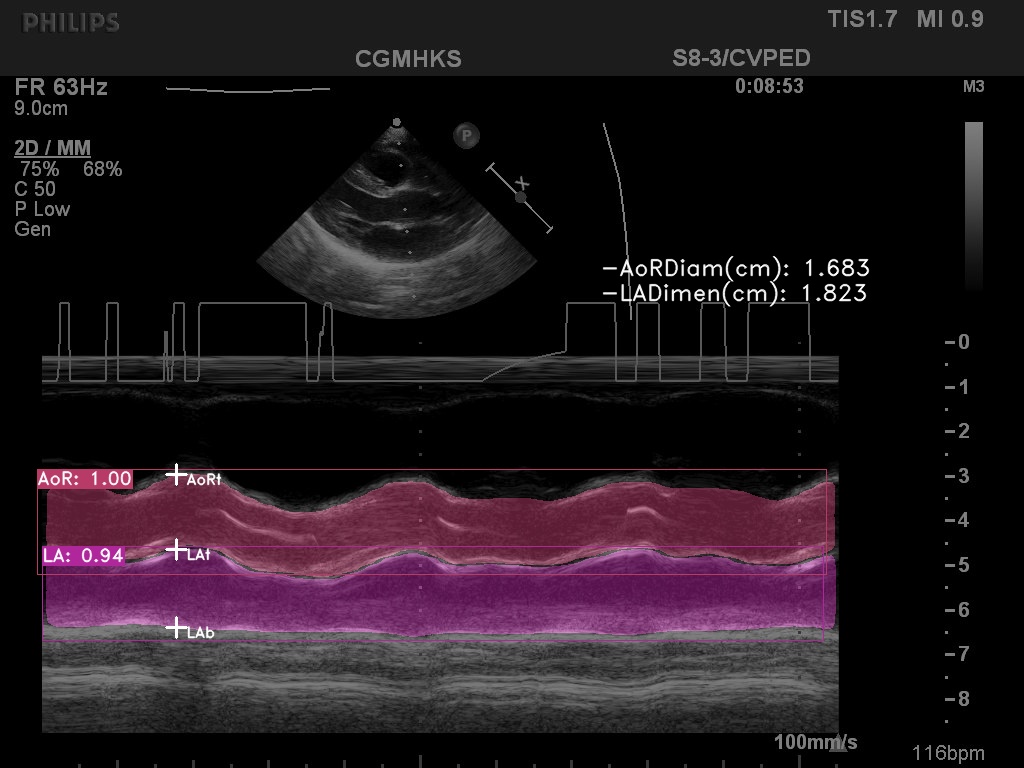} &
        \includegraphics[width=0.3\textwidth]{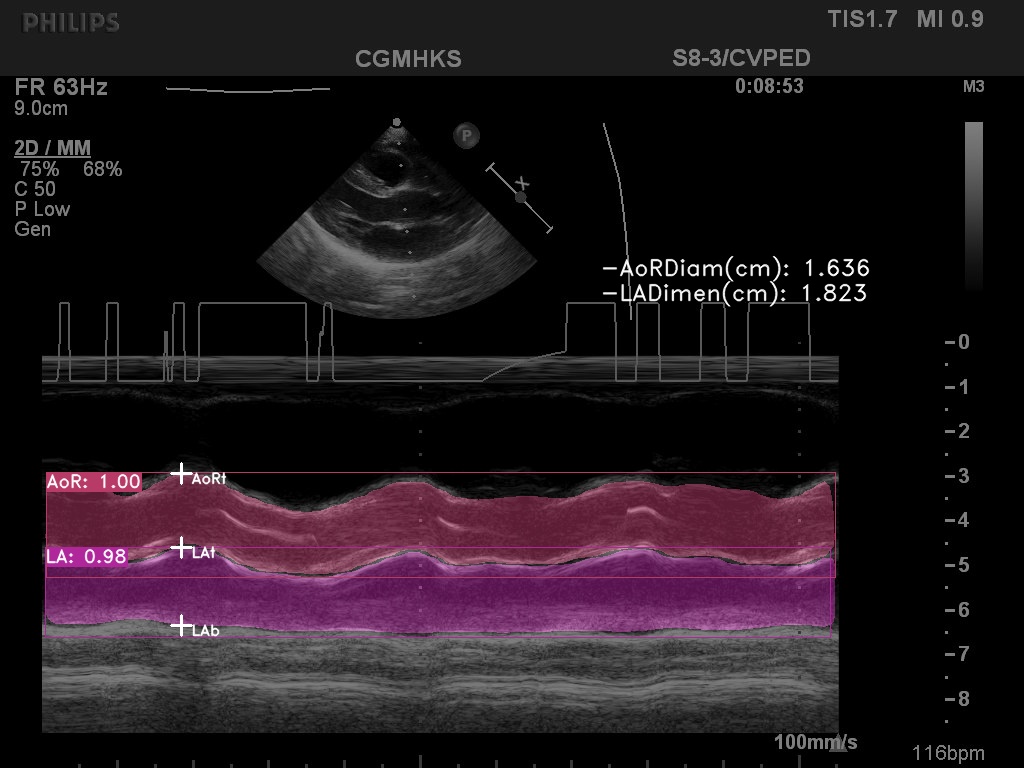} &
        \includegraphics[width=0.3\textwidth]{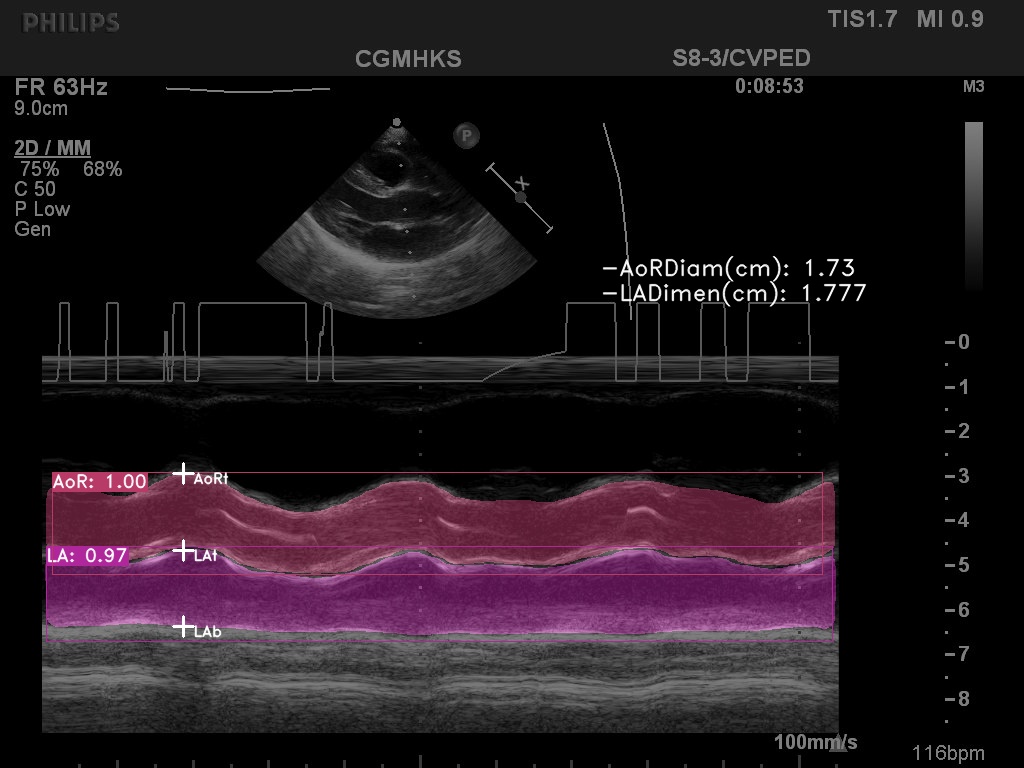}
        \\
        \includegraphics[width=0.3\textwidth]{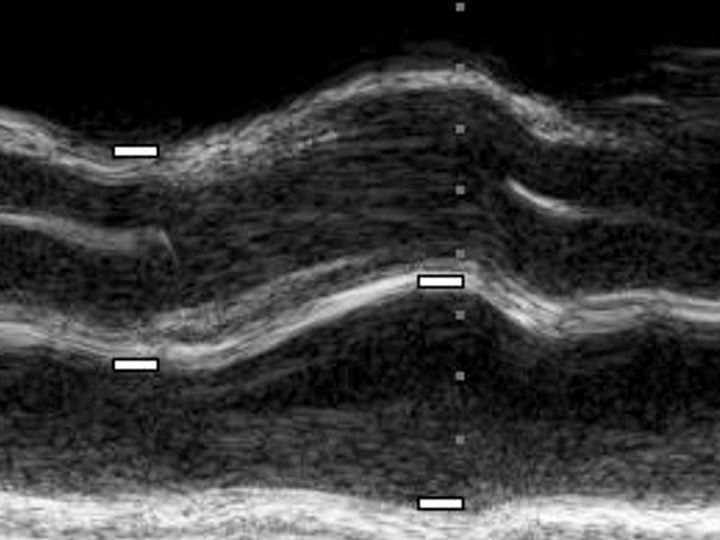} &
        \includegraphics[width=0.3\textwidth]{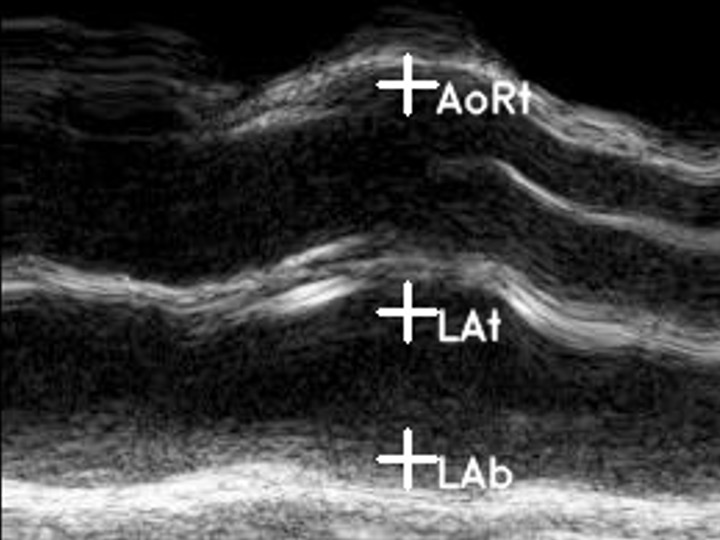} &
        \includegraphics[width=0.3\textwidth]{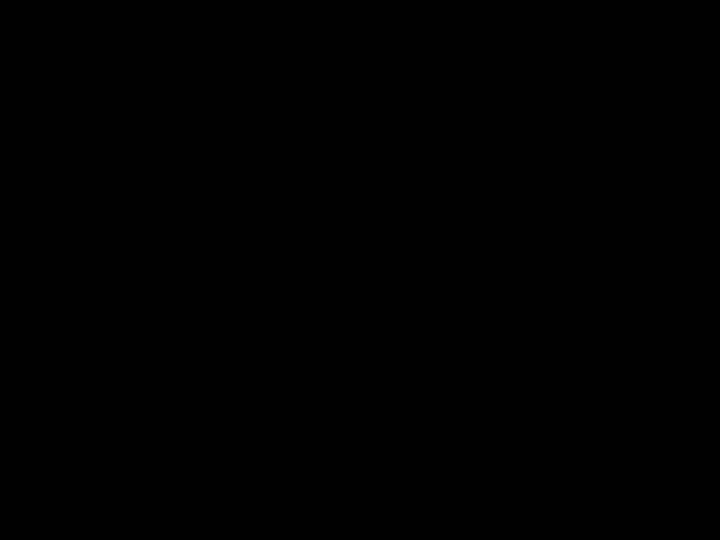} &
        \includegraphics[width=0.3\textwidth]{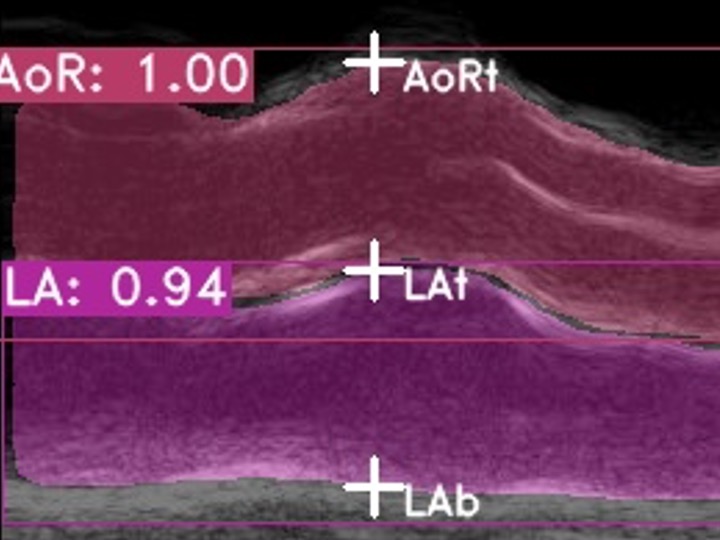} &
        \includegraphics[width=0.3\textwidth]{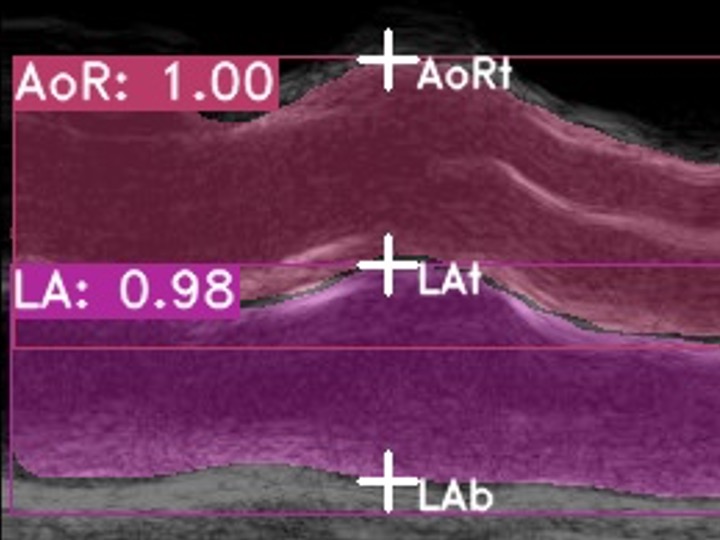} &
        \includegraphics[width=0.3\textwidth]{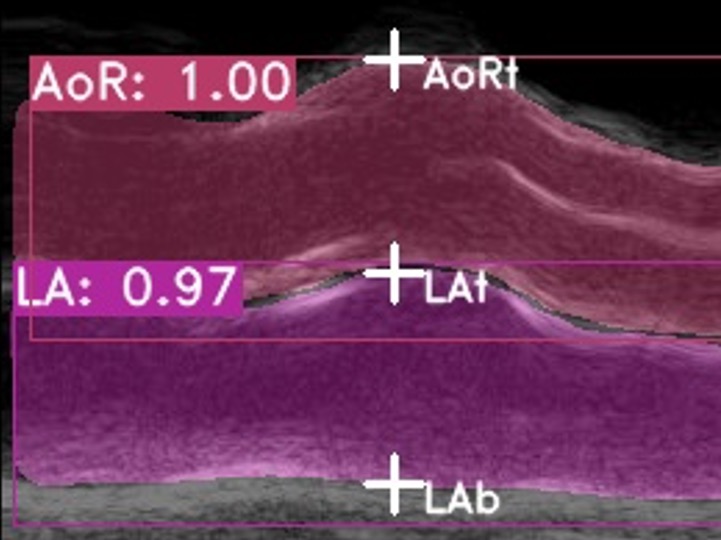}
        \\
    \end{tabular}%
    }
    \caption{AV sampled measurement results from an obvious view. Human labelling has some bias of inaccurate locating boundaries. As MENN does not target toward AV view, a blank sample remains. The t indicates the top of the boundary, and the b indicates the bottom.}
    \label{fig10}
\end{figure*}

\begin{figure*}[!t]
    \centering
    \resizebox{\textwidth}{!}{%
    \begin{tabular}{ccccc>{\columncolor[HTML]{F2F2F2}}c}
        Ground truth & Human Candidate1 & MENN UPANet80 V2 & AMEM maYOLACT &  AMEM RTMDet & AMEM UPANet80 V2 (RAMEM)\\
        \includegraphics[width=0.3\textwidth]{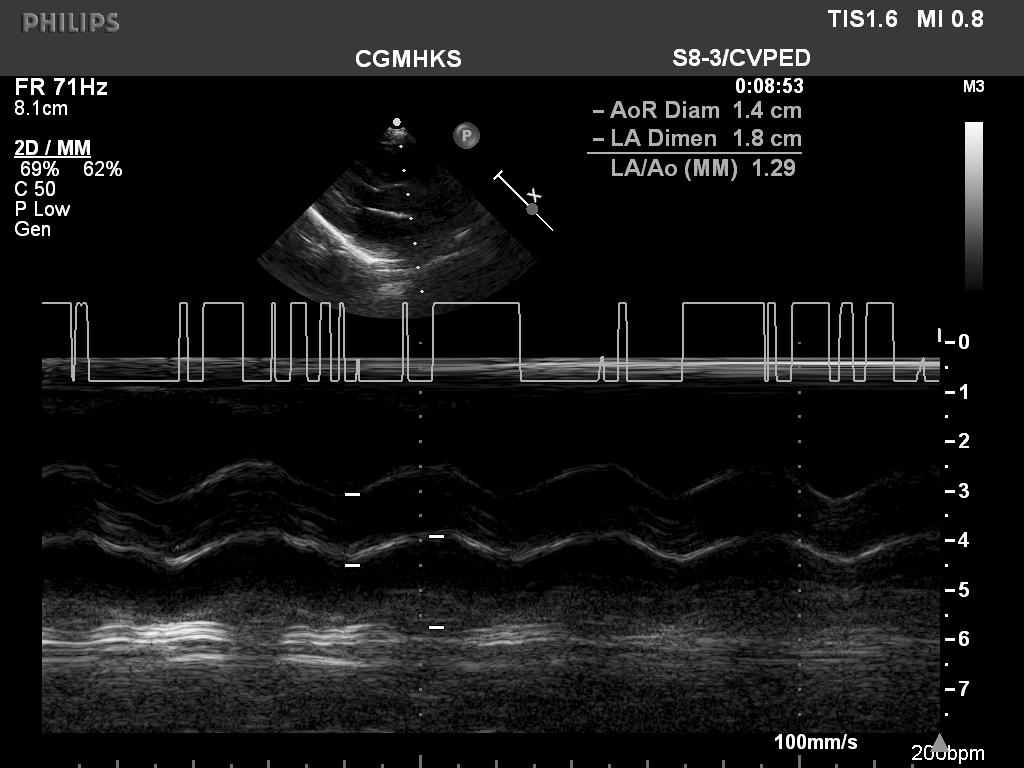} &
        \includegraphics[width=0.3\textwidth]{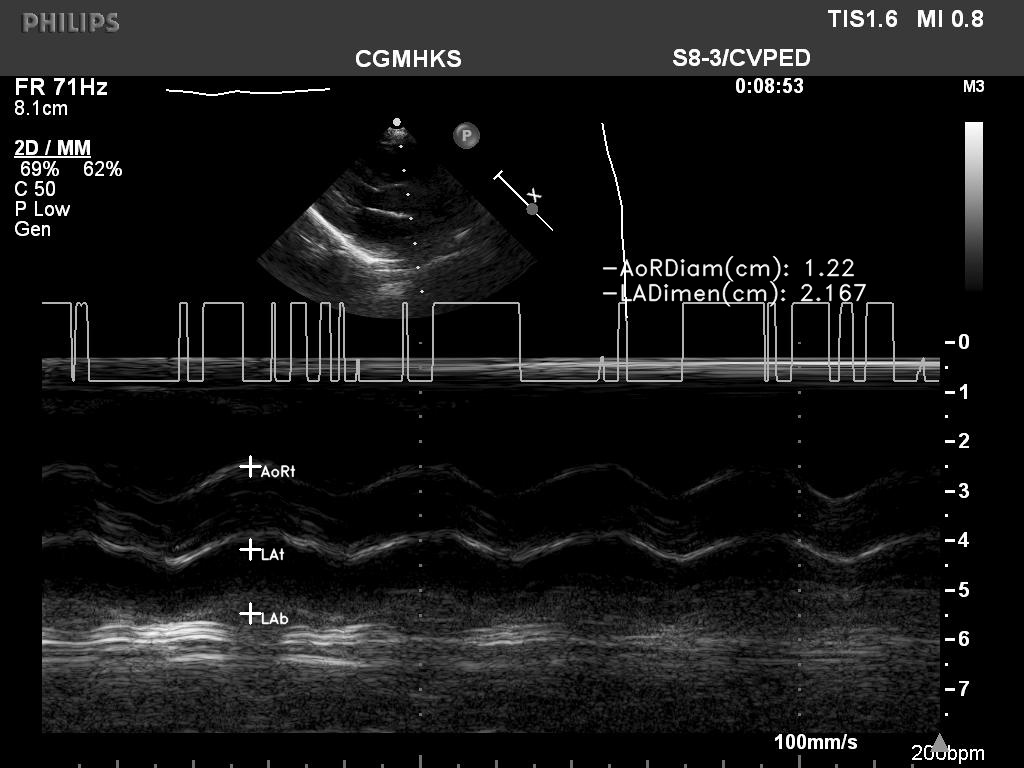} &
        \includegraphics[width=0.3\textwidth]{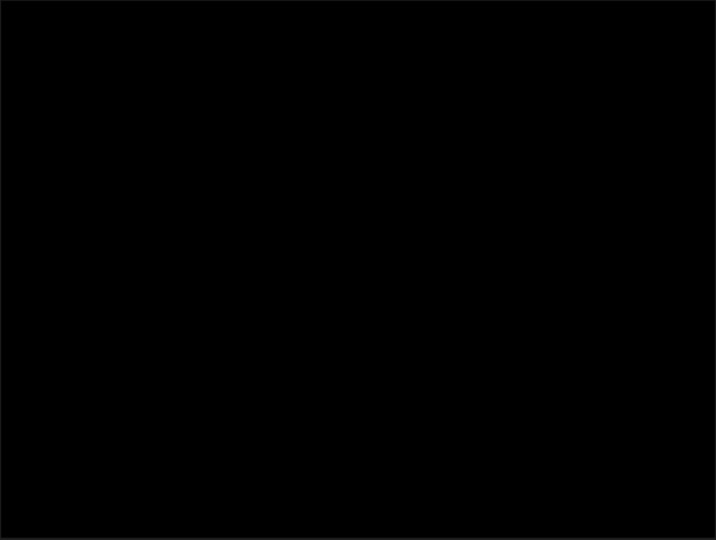} &
        \includegraphics[width=0.3\textwidth]{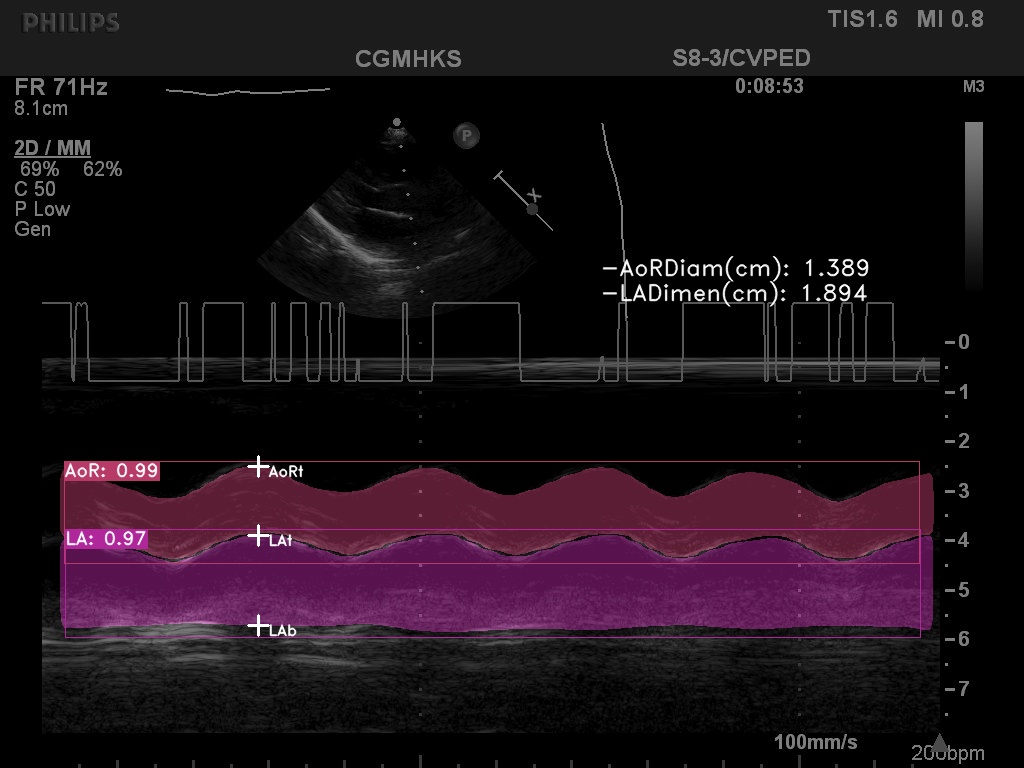} &
        \includegraphics[width=0.3\textwidth]{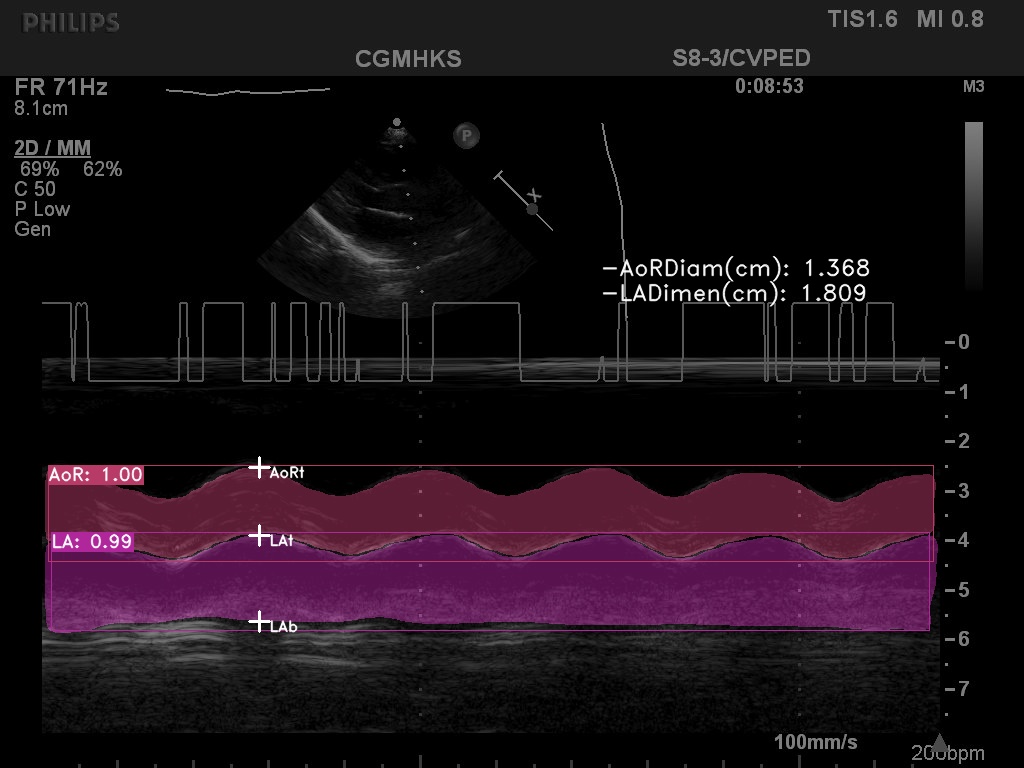} &
        \includegraphics[width=0.3\textwidth]{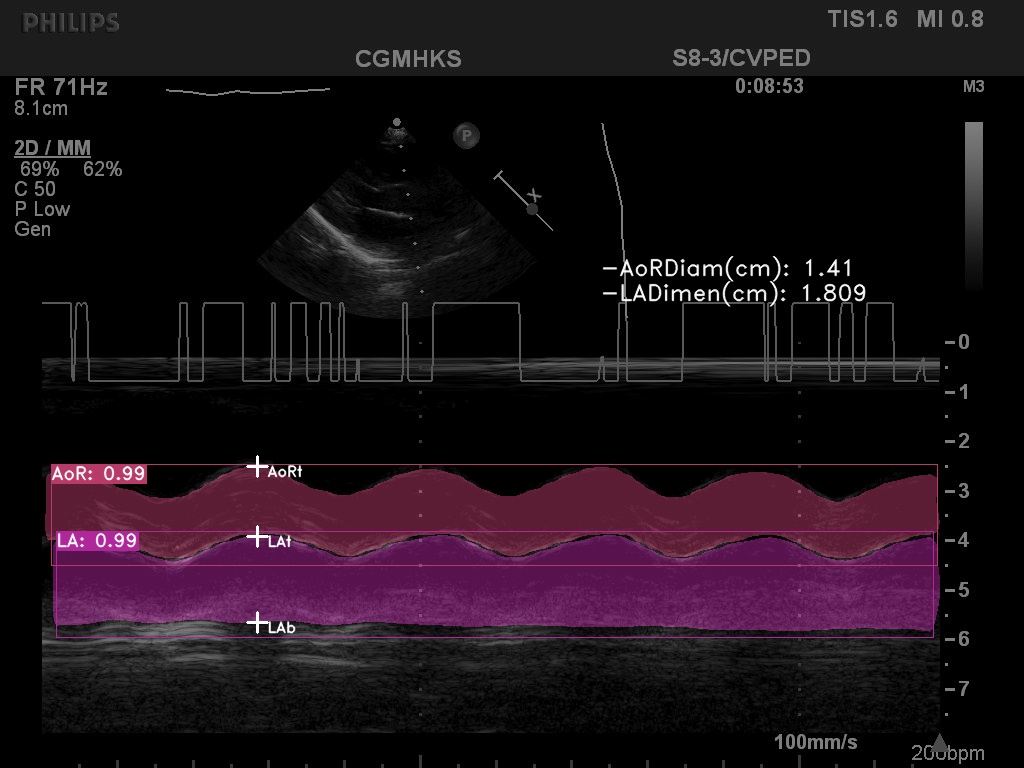}
        \\
        \includegraphics[width=0.3\textwidth]{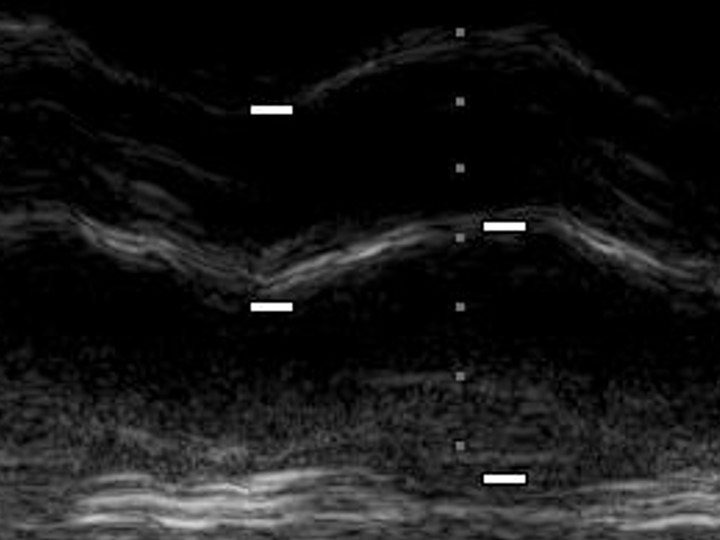} &
        \includegraphics[width=0.3\textwidth]{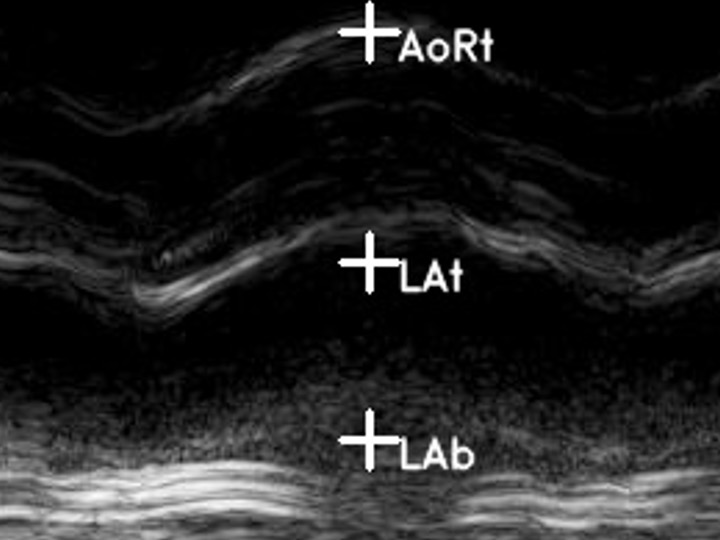} &
        \includegraphics[width=0.3\textwidth]{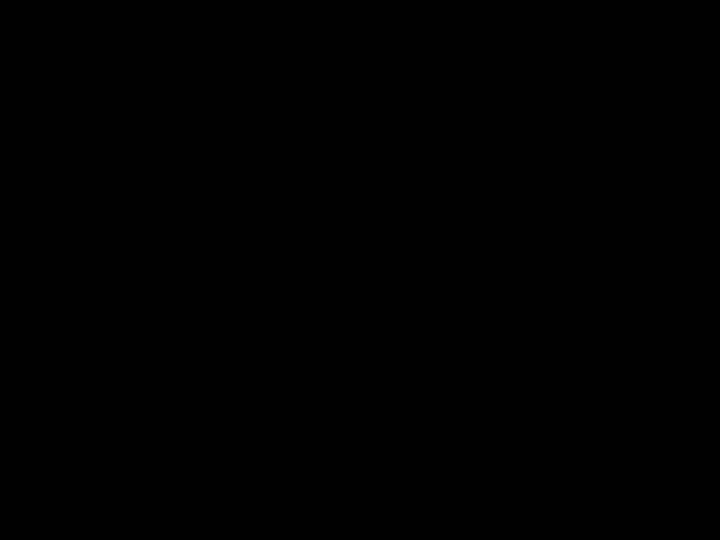} &
        \includegraphics[width=0.3\textwidth]{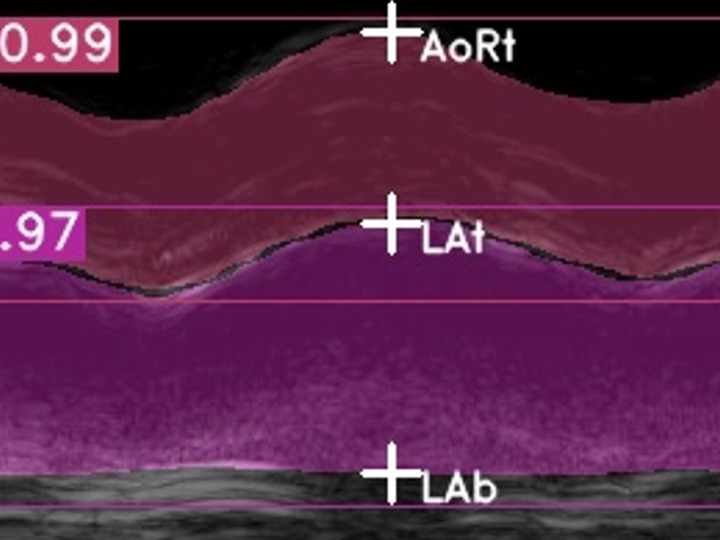} &
        \includegraphics[width=0.3\textwidth]{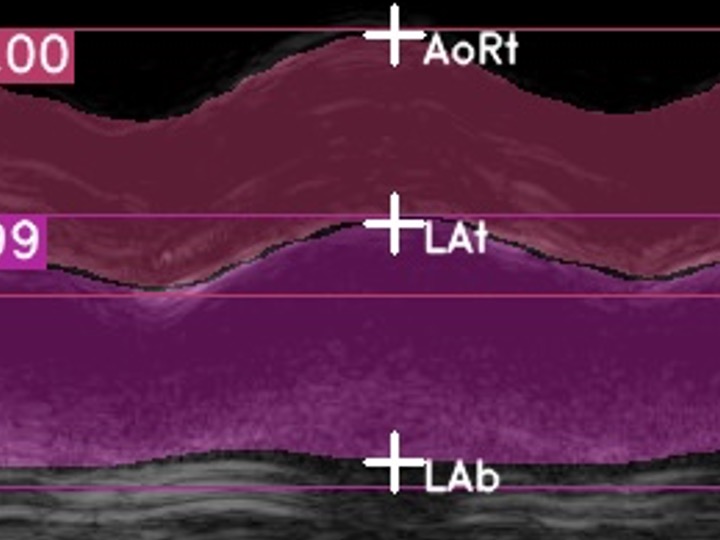} &
        \includegraphics[width=0.3\textwidth]{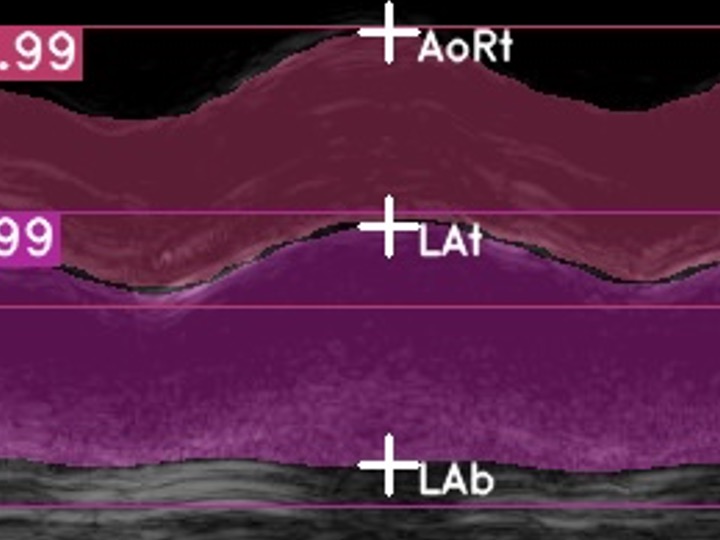}
        \\
    \end{tabular}%
    }
    \caption{AV sampled measurement results from an ambiguous view. The same inaccurate scene from the human can also be seen in this figure. In fact, a slightly unstable locating boundaries result also happened in the ground truth, which further indicates the aforementioned variance problem.}
    \label{fig11}
\end{figure*}

\begin{figure*}[!t]
    \centering
    \resizebox{\textwidth}{!}{%
    \begin{tabular}{ccccc>{\columncolor[HTML]{F2F2F2}}c}
         Ground truth & Human candidate1 & MENN UPANet80 V2 & AMEM maYOLACT & AMEM RTMDet & AMEM UPANet80 V2 (RAMEM)\\
        \includegraphics[width=0.3\textwidth]{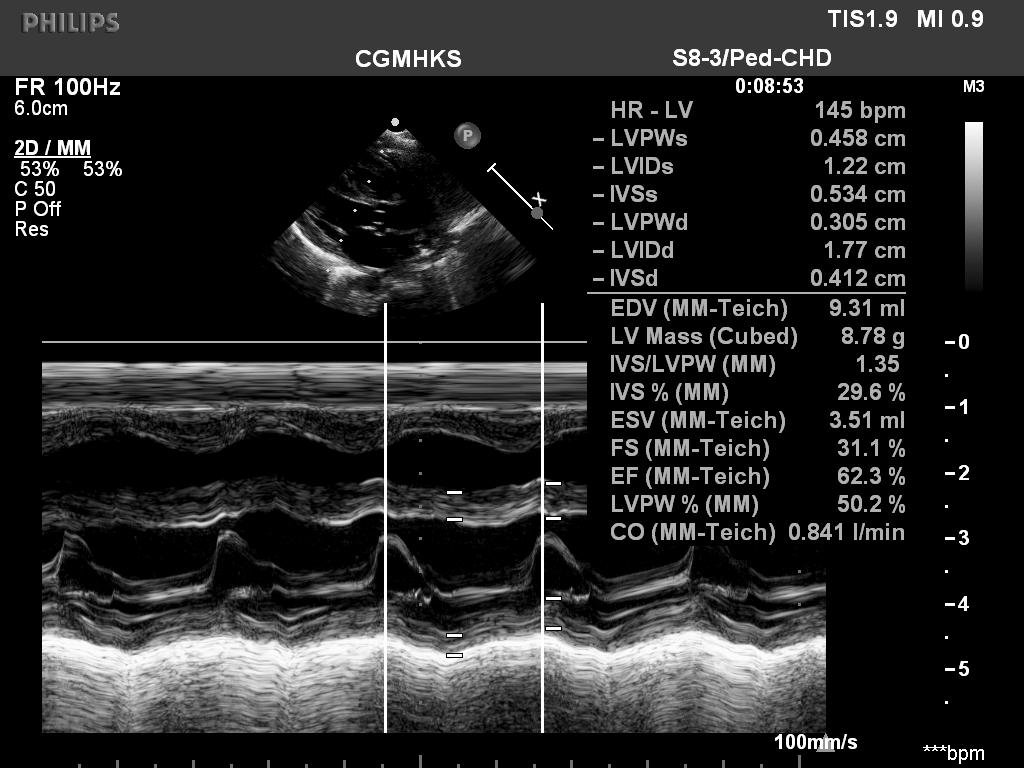} &
        \includegraphics[width=0.3\textwidth]{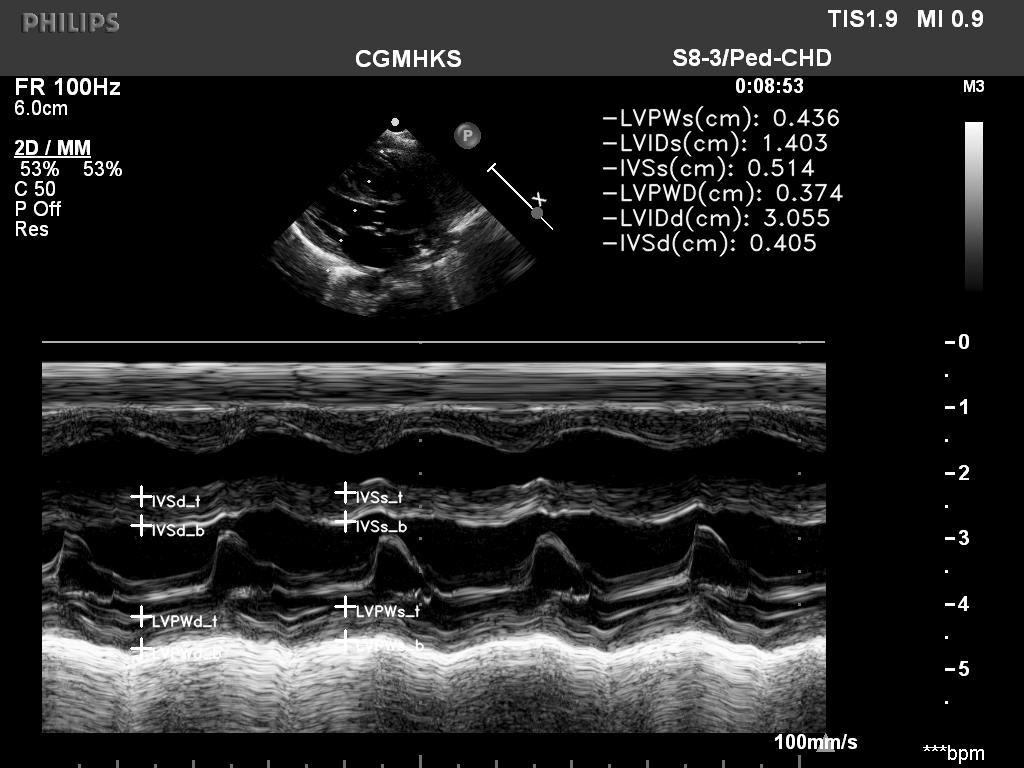} &
        \includegraphics[width=0.3\textwidth]{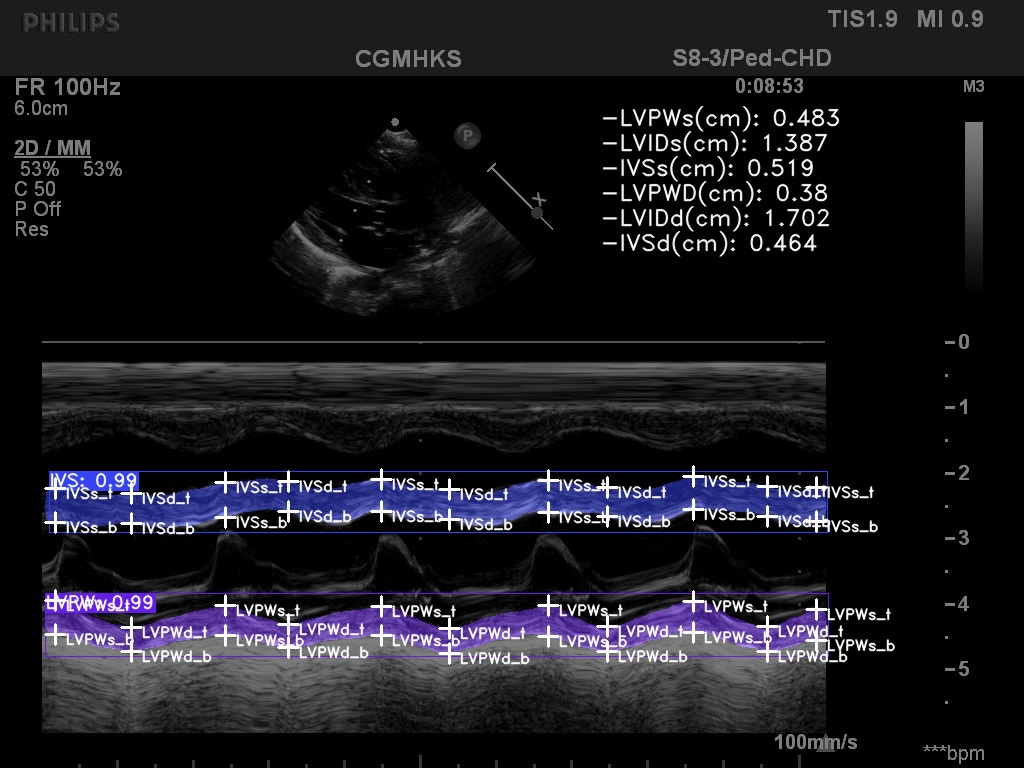} &
        \includegraphics[width=0.3\textwidth]{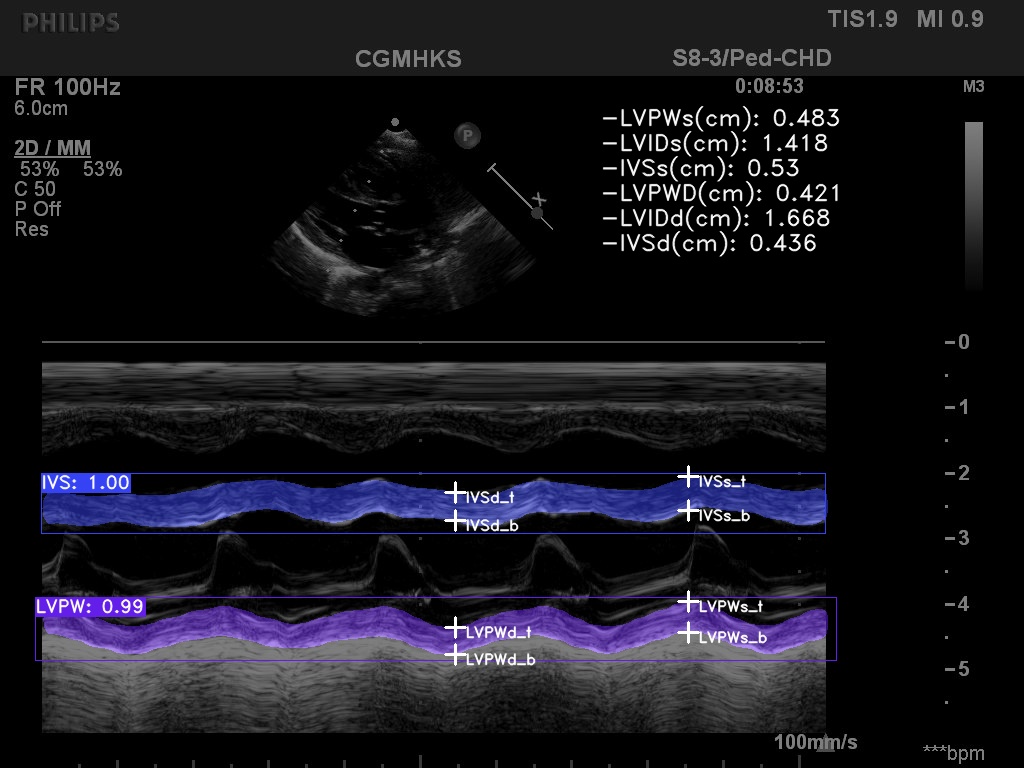} &
        \includegraphics[width=0.3\textwidth]{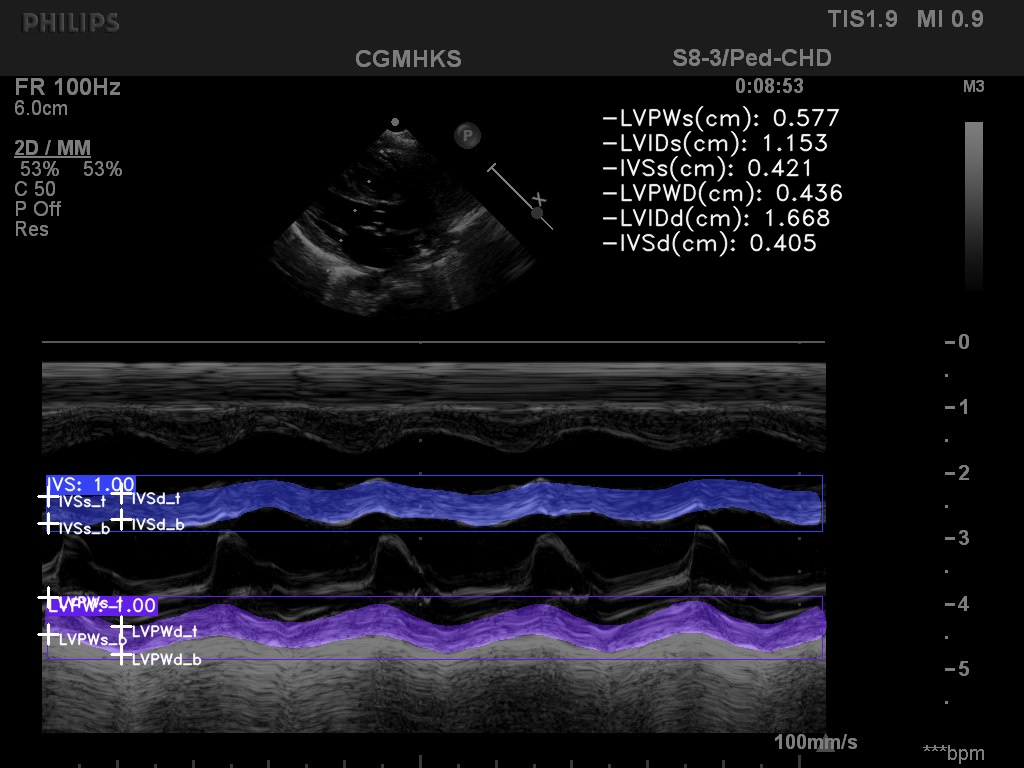} &
        \includegraphics[width=0.3\textwidth]{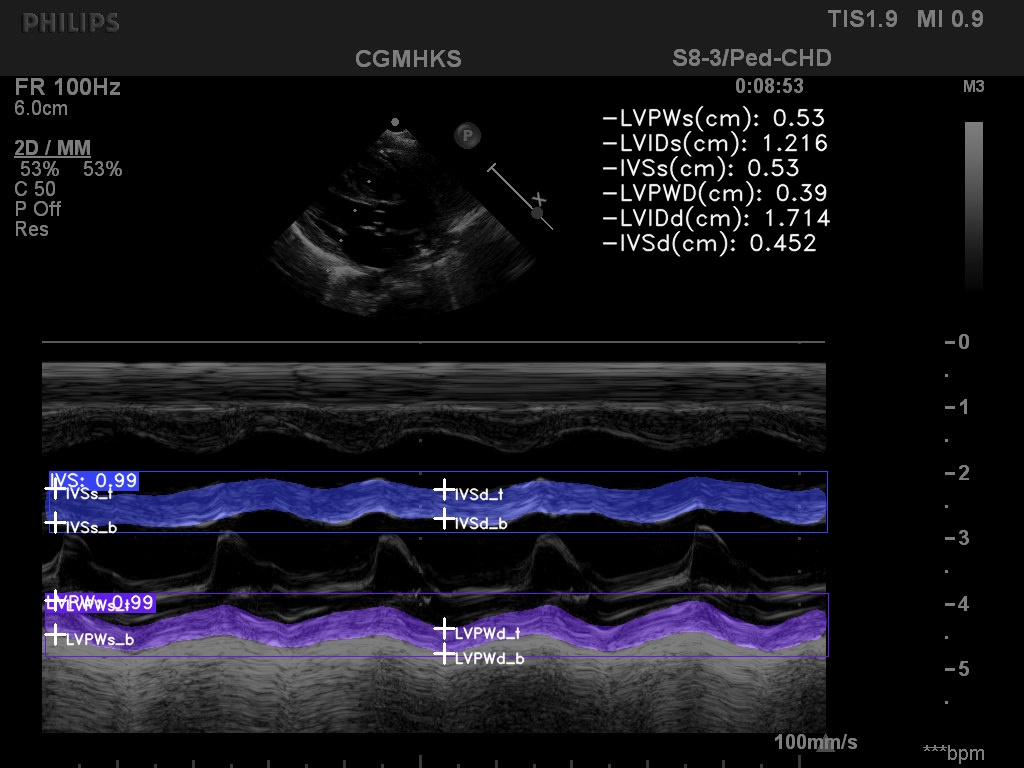}
        \\
        \includegraphics[width=0.3\textwidth]{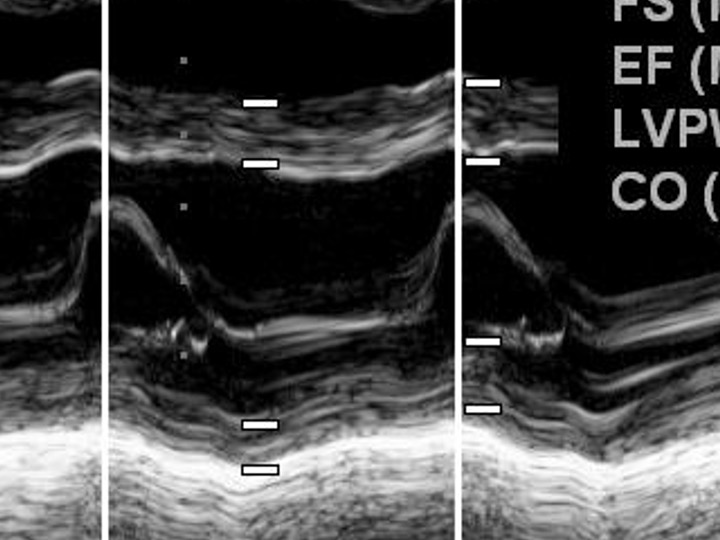} &
        \includegraphics[width=0.3\textwidth]{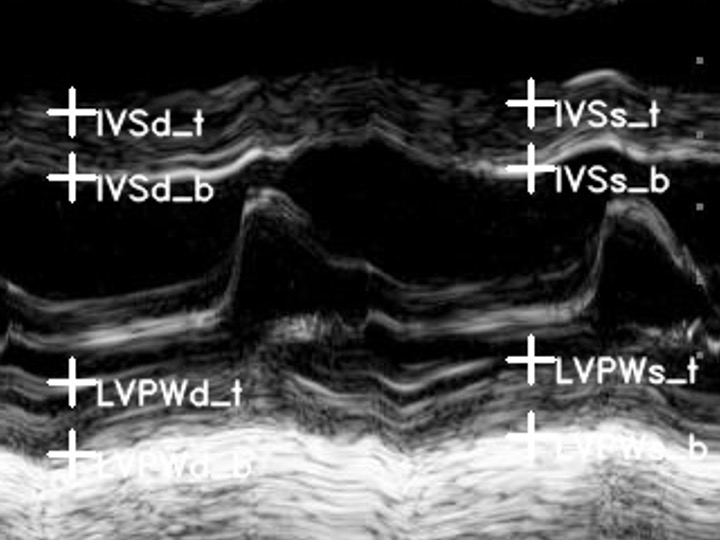} &
        \includegraphics[width=0.3\textwidth]{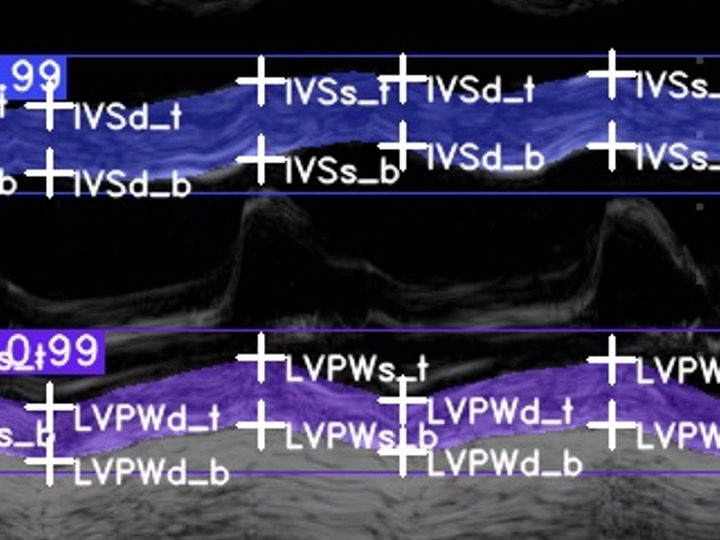} &
        \includegraphics[width=0.3\textwidth]{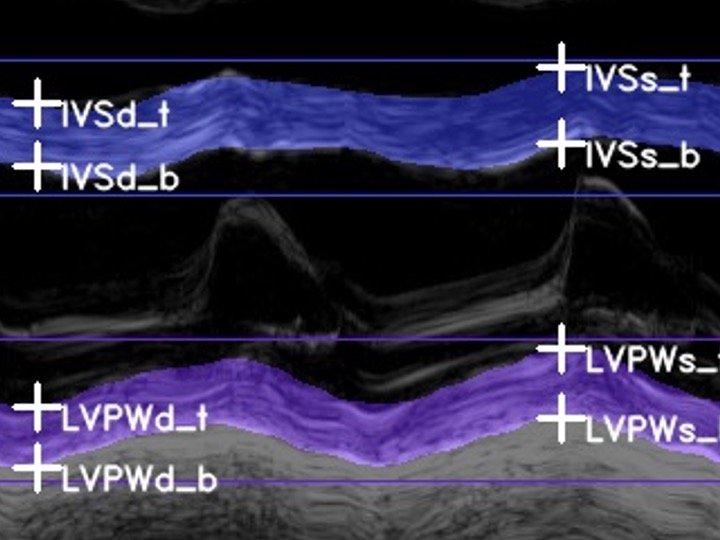} &
        \includegraphics[width=0.3\textwidth]{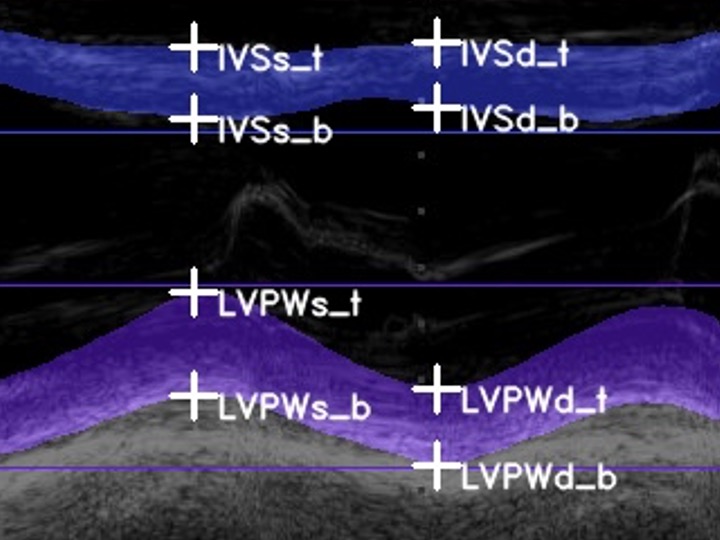} &
        \includegraphics[width=0.3\textwidth]{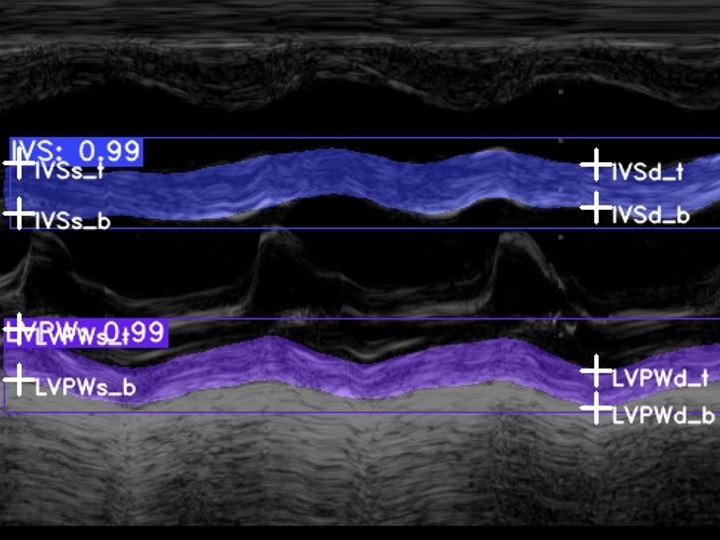}
        \\
    \end{tabular}%
    }
    \caption{LV sampled measurement results from an obvious view. As there are diastole and systole moments, each moment is abbreviated into s and d, along with the same abbreviations of the top and bottom boundary as the above figures. Considering AMEM takes the topmost point as the systole and the lowest point as the diastole, the locations may vary.}
    \label{fig12}
\end{figure*}

\begin{figure*}[!t]
    \centering
    \resizebox{\textwidth}{!}{%
    \begin{tabular}{ccccc>{\columncolor[HTML]{F2F2F2}}c}
        Ground truth & Human candidate1 & MENN UPANet80 V2 & AMEM maYOLACT & AMEM RTMDet & AMEM UPANet80 V2 (RAMEM)\\
        \includegraphics[width=0.3\textwidth]{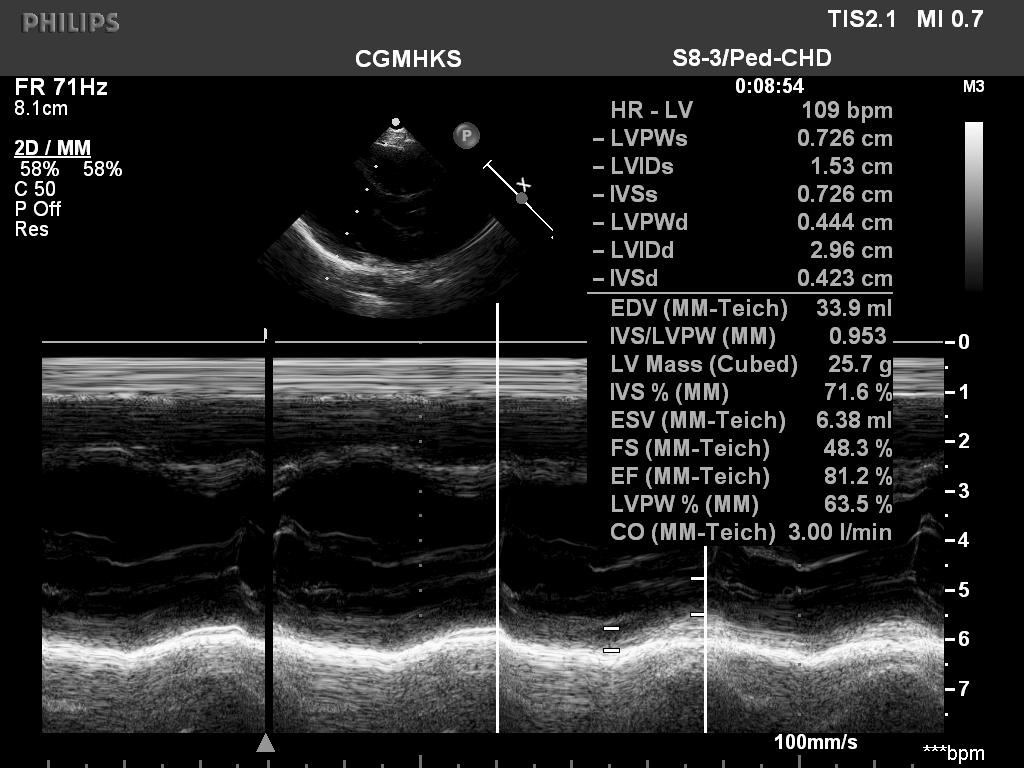} &
        \includegraphics[width=0.3\textwidth]{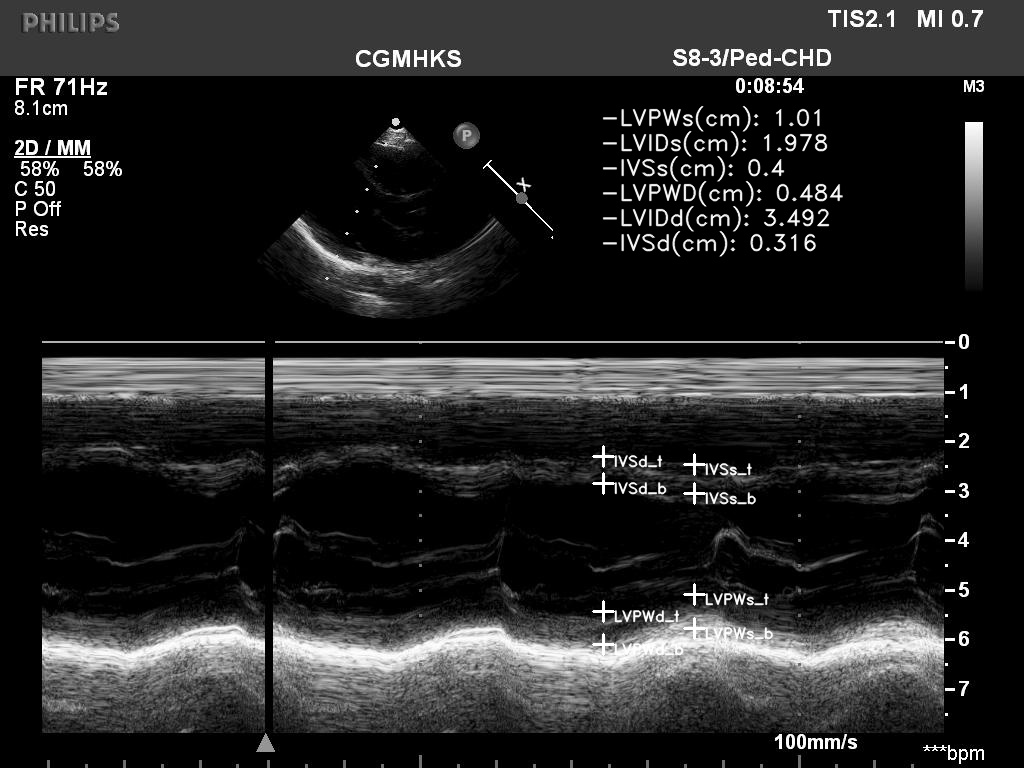} &
        \includegraphics[width=0.3\textwidth]{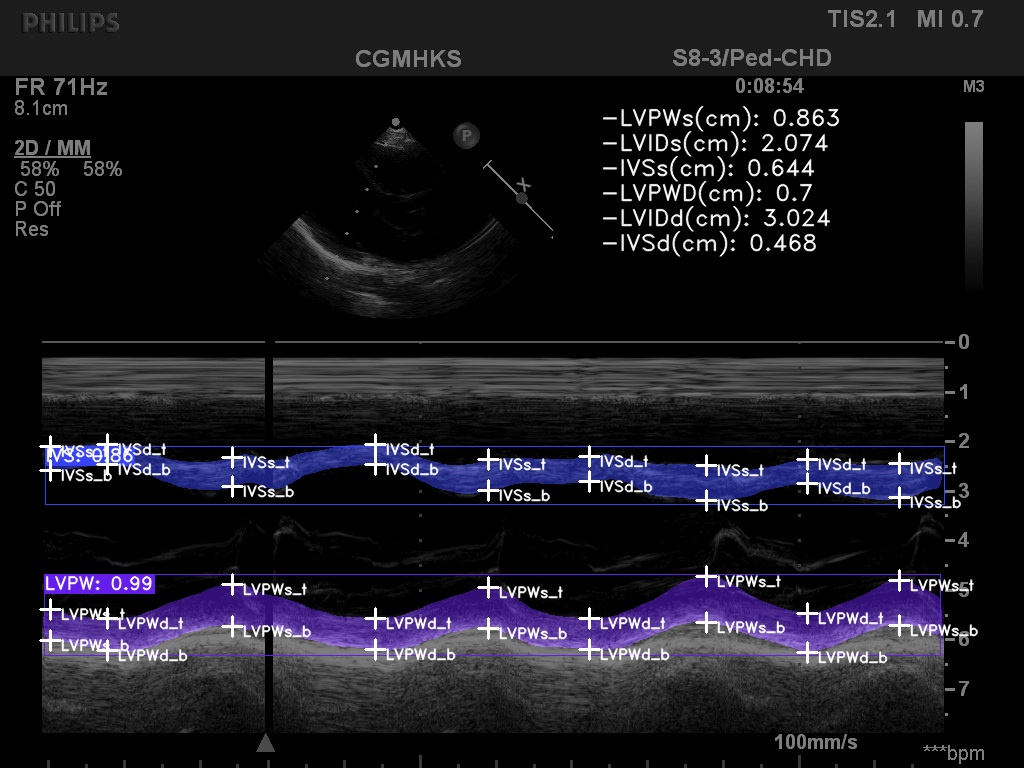} &
        \includegraphics[width=0.3\textwidth]{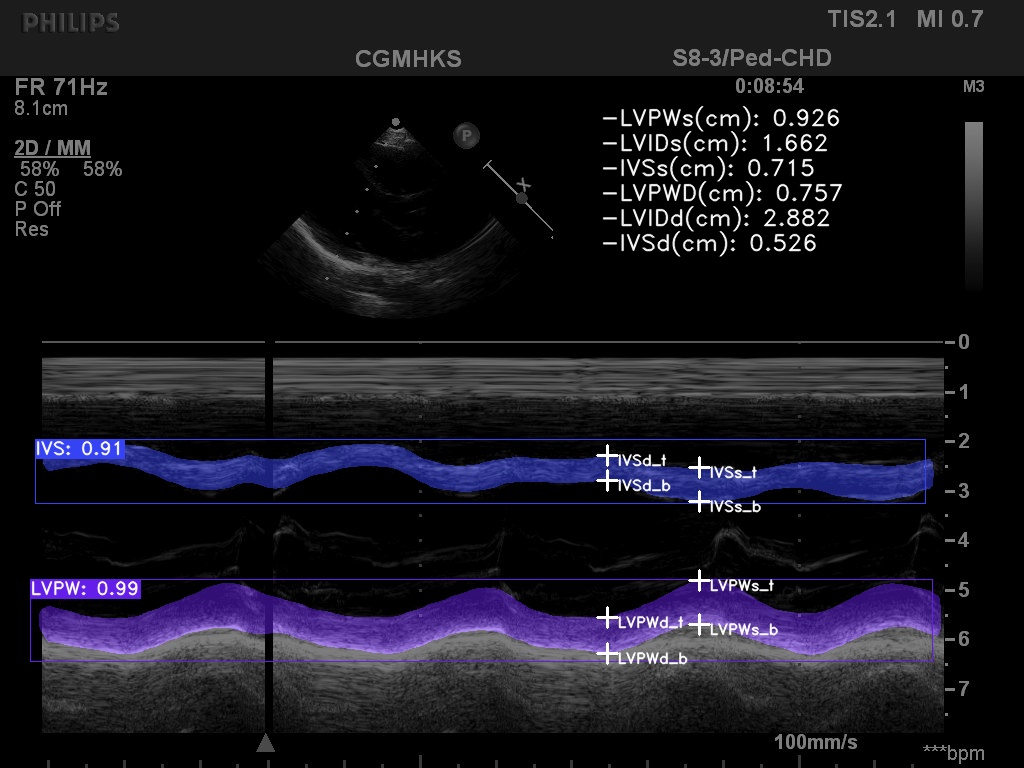} &
        \includegraphics[width=0.3\textwidth]{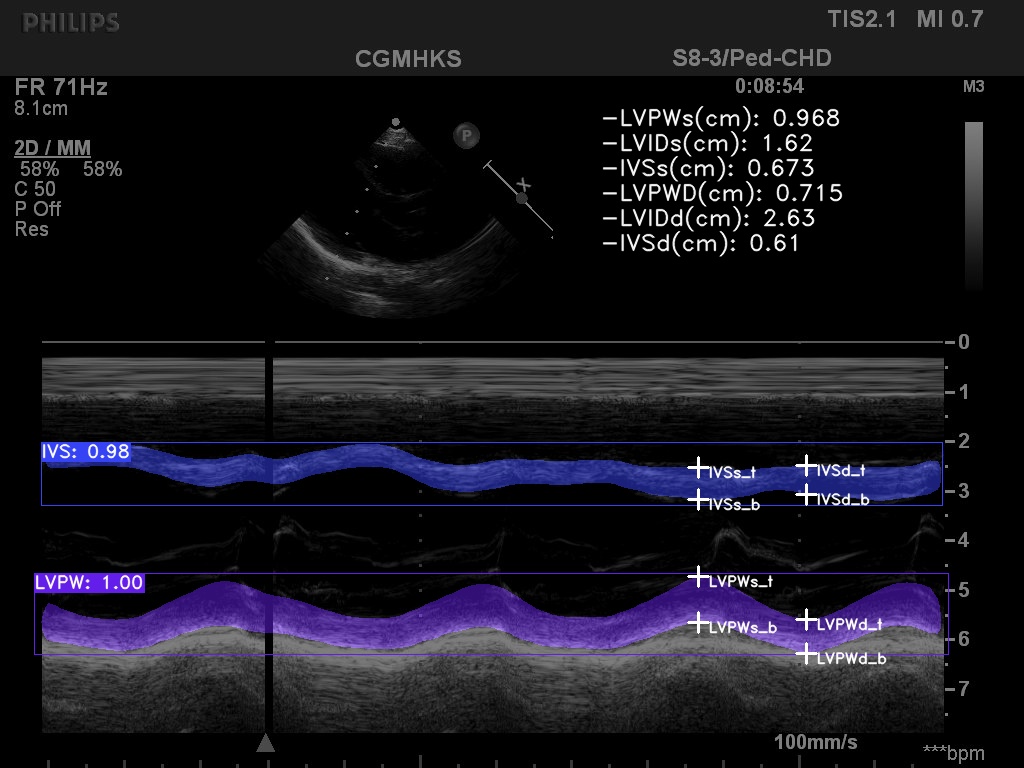} &
        \includegraphics[width=0.3\textwidth]{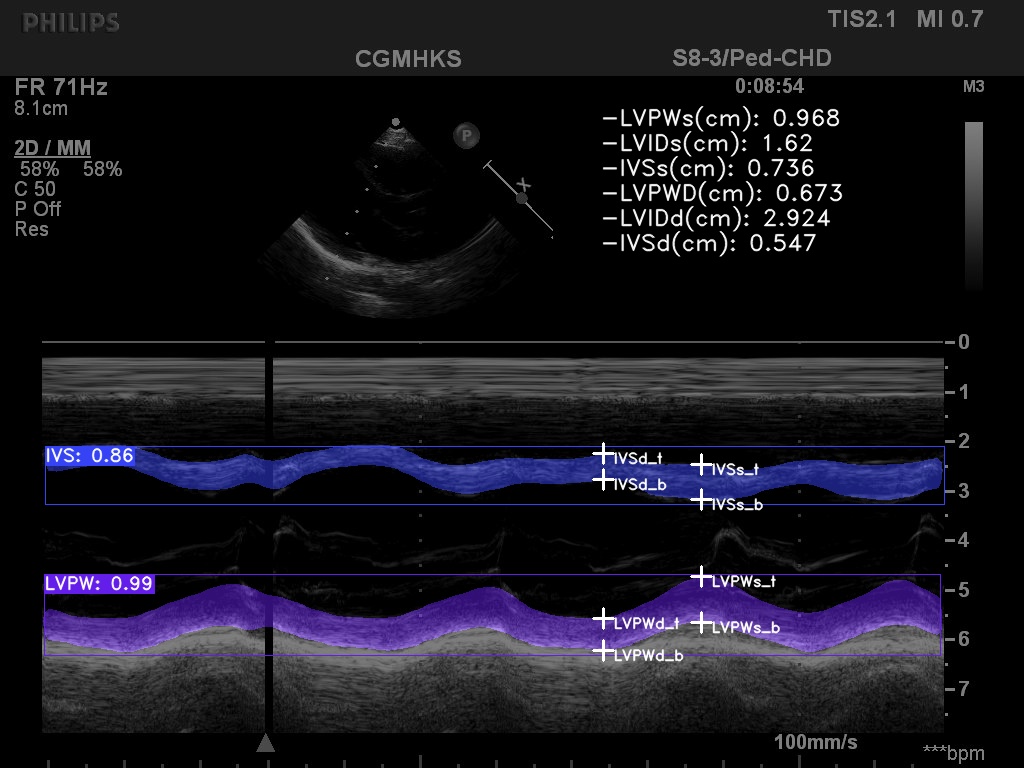}
        \\
        \includegraphics[width=0.3\textwidth]{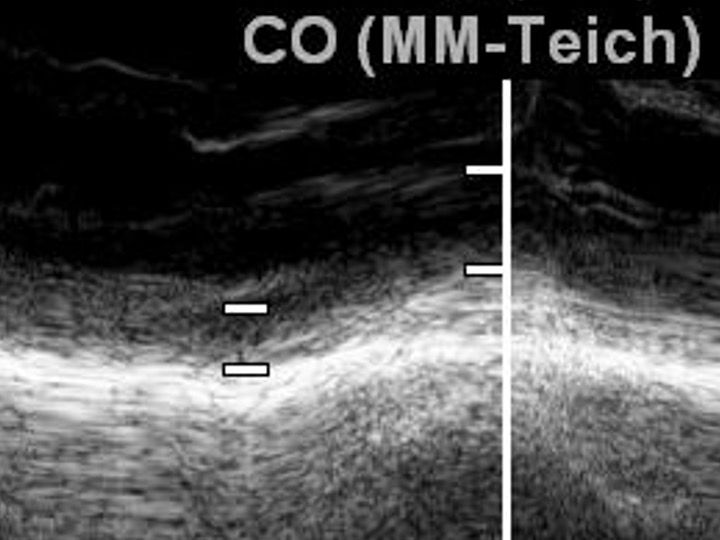} &
        \includegraphics[width=0.3\textwidth]{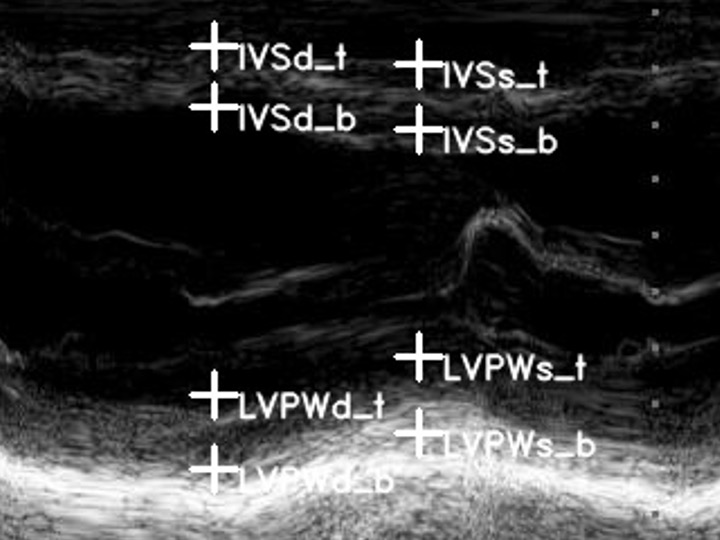} &
        \includegraphics[width=0.3\textwidth]{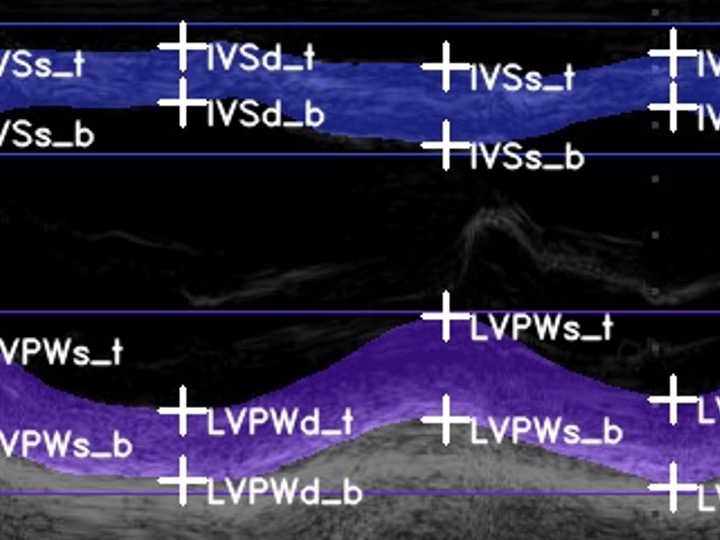} &
        \includegraphics[width=0.3\textwidth]{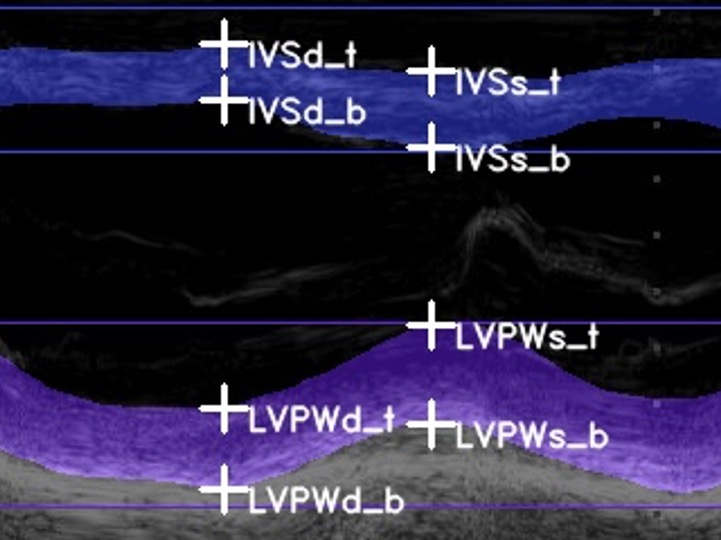} &
        \includegraphics[width=0.3\textwidth]{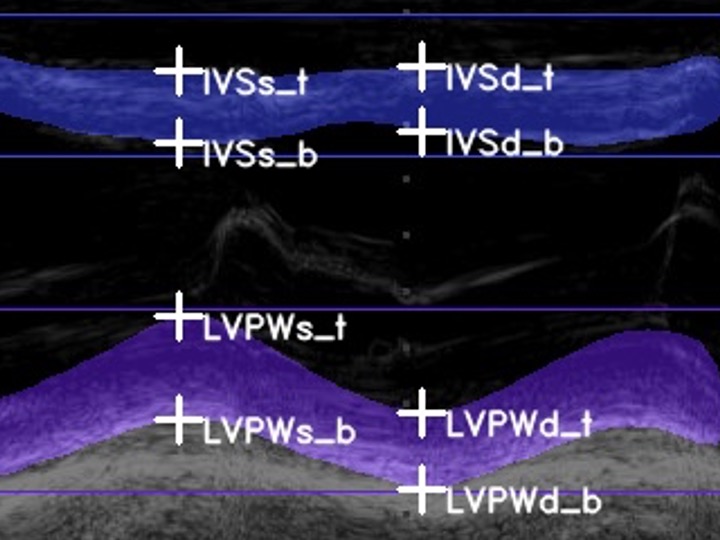} &
        \includegraphics[width=0.3\textwidth]{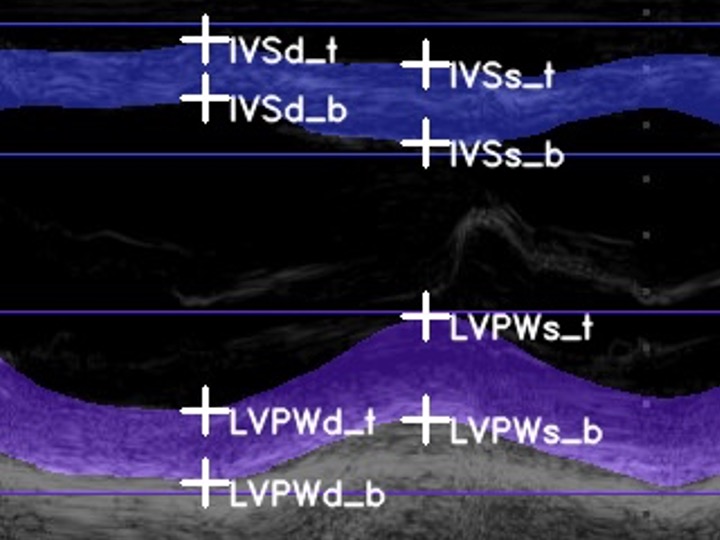}
        \\
    \end{tabular}%
    }
    \caption{LV sampled measurement results from an ambiguous view. The samples show the importance of taking the topmost and the lowest point is more aligned to the ground truth. Moreover, a great mask prediction is essential when comparing maYOLACT, RTMDet, and UPANet80 V2 upon the same measurement, AMEM.}
    \label{fig13}
\end{figure*}

\subsection{AMEM in MEIS}
\label{sec5.4}
We apply our model UPANet80 V2 in Table \ref{tab5} \#6 as the final scheme of embedded with the proposed AMEM and test upon MEIS. Therefore, it makes RAMEM and offers the labelling into an automatic end-to-end scheme. To evaluate the automatic scheme on a real-world daily based, the 38 patients’ examinations are collected from clinical visits in Table \ref{tab6}, which are contained in the testing data. Also, the sampled images from obvious and ambiguous views in AV are in Fig. \ref{fig10} and Fig. \ref{fig11}, separately.  LV follows the same in Fig. \ref{fig12} and Fig. \ref{fig13}. Among the sampled images, ground truth images show a possible location of diastole and systole. The key is to locate one of the period locations in order to get the indicators. For showing the capability of different viewing situations, sampled results toward clear and fuggy echocardiograms have been implemented along with the original results on the top row and zoom-in results on the bottom. The compared candidates include the practical manual examination results from Humans, another existing automatic measurement toward LV from MENN, and our proposed method AMEM. The error evaluation index in Table \ref{tab6} covers mean absolute error (MAE), mean square error (MSE), and costing time, including measurement (FPS). The mean and standard deviation toward each error index and indicator are presented to make a comprehensive comparison even more effortless. 

\textbf{AMEM upon UPANet80 V2 (RAMEM) surpasses human in real-time detection} (Table \ref{tab6}. Fig. \ref{fig10}, Fig. \ref{fig11}, Fig. \ref{fig12}, and Fig. \ref{fig13}) – The results show two milestones: 1) seamless diagnosis  by reducing the whole scheme costing time less than 0.041 seconds, and 2) minimizing mean and variance of error while surpassing human performance, MENN, and other backbones. 1) can be seen from the FPS from input to output wanted indicators with 24 FPS. The mean and std errors of 2) outshine existing automatic LV measurement (MENN) and humans in significant margins. Sampled figures further solidfy this by showing a more accurate anchor on the mask. When comparing with MENN, as MENN inclines to calculate the average length from each diastole and systole point, the cost time and detection are jeopardized by the number of anchors and in-accurate points. Although such measurement covers all possible points that can prevent outlier effects, a poor sample period could cause the result and sensitive to a specific scenario. In other words, MENN is relatively more inflexible than AMEM. When comparing with the results from human candidate 1, biased locates have occurred. The bias happened at systole and diastole or vessel boundary. A worse scene can be seen in Fig. \ref{fig13} when the echocardiogram has less clarity. Conversely, the sampled measurement results from AMEM have accurately located end-systole in AV, even in less obvious echocardiograms. The same result is also shown in LV with less bias. The mean and stand deviation of the error index from ours in Table \ref{tab6} shows less bias and hesitation measurement among all. With better-predicted masks from a better backbone as UPANets V2, the result is more accurate compared to different backbones.

\subsection{Discussion}
\label{sec5.5}
\begin{figure*}[!t]
    \centering
    \resizebox{\textwidth}{!}{%
    \begin{tabular}{ccccc>{\columncolor[HTML]{F2F2F2}}c}
        Ground truth & Human candidate1 & MENN UPANet80 V2 & AMEM maYOLACT & AMEM RTMDet & AMEM UPANet80 V2 (RAMEM)\\
        \includegraphics[width=0.3\textwidth]{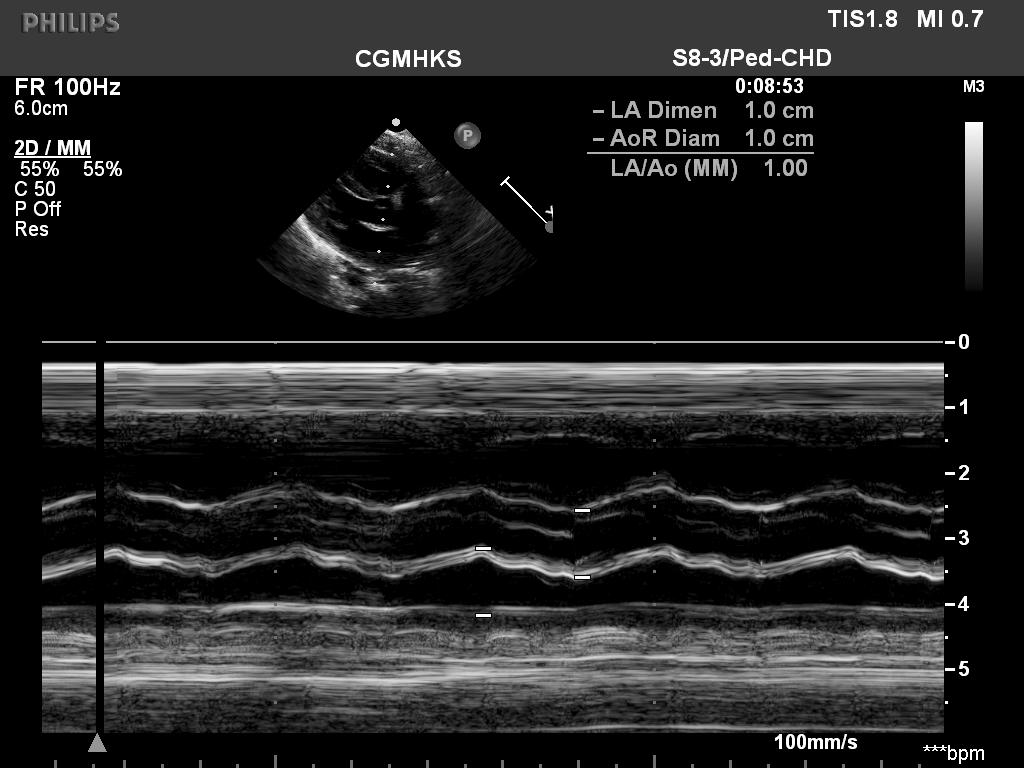} &
        \includegraphics[width=0.3\textwidth]{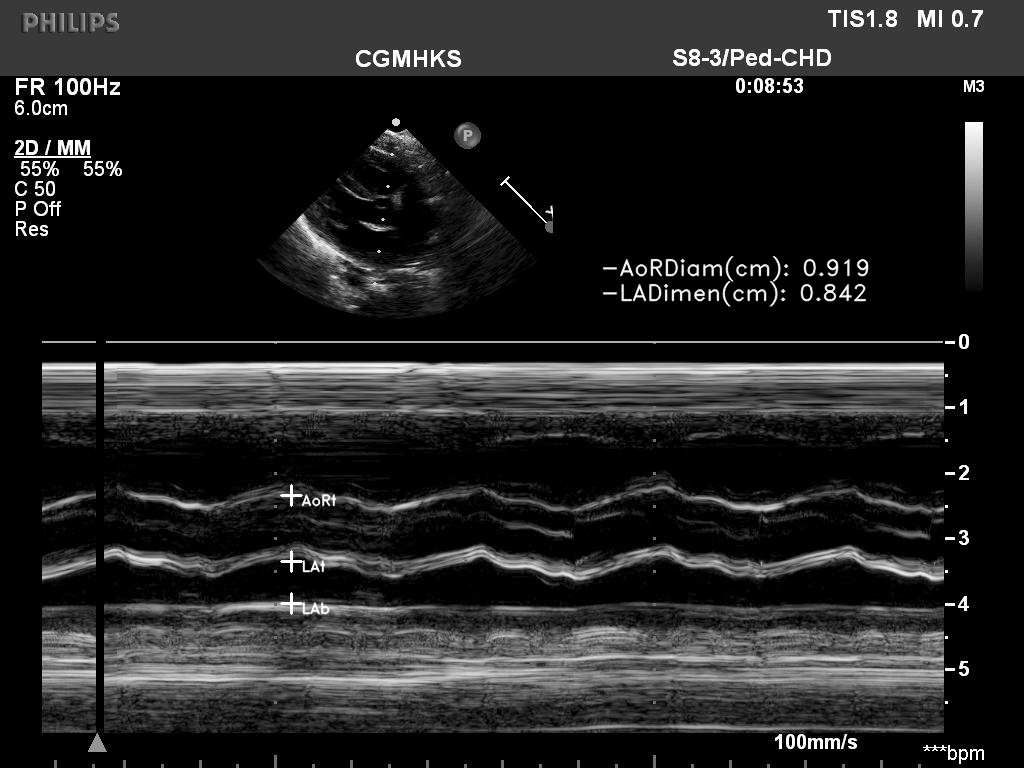} &
        \includegraphics[width=0.3\textwidth]{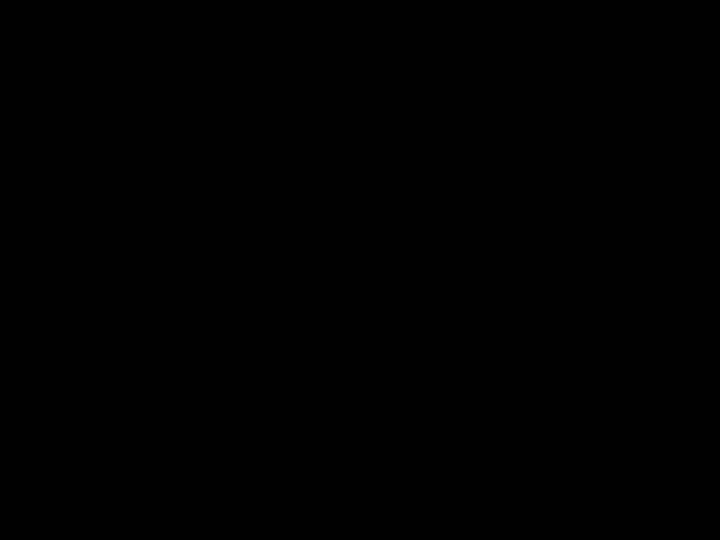} &
        \includegraphics[width=0.3\textwidth]{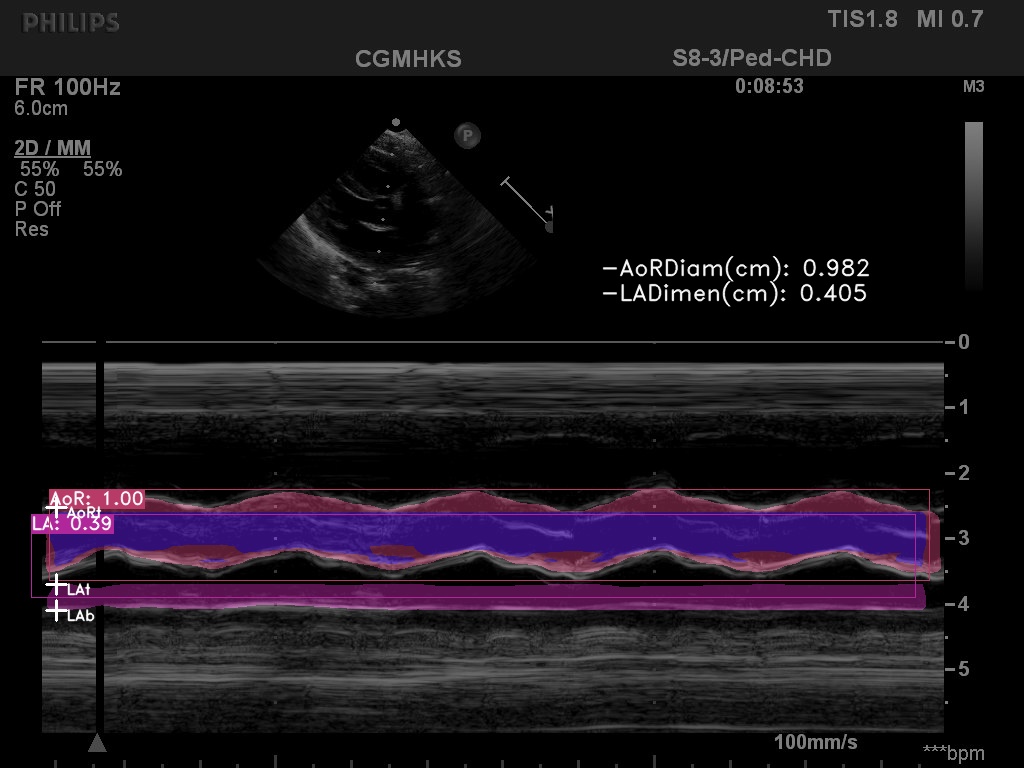} &
        \includegraphics[width=0.3\textwidth]{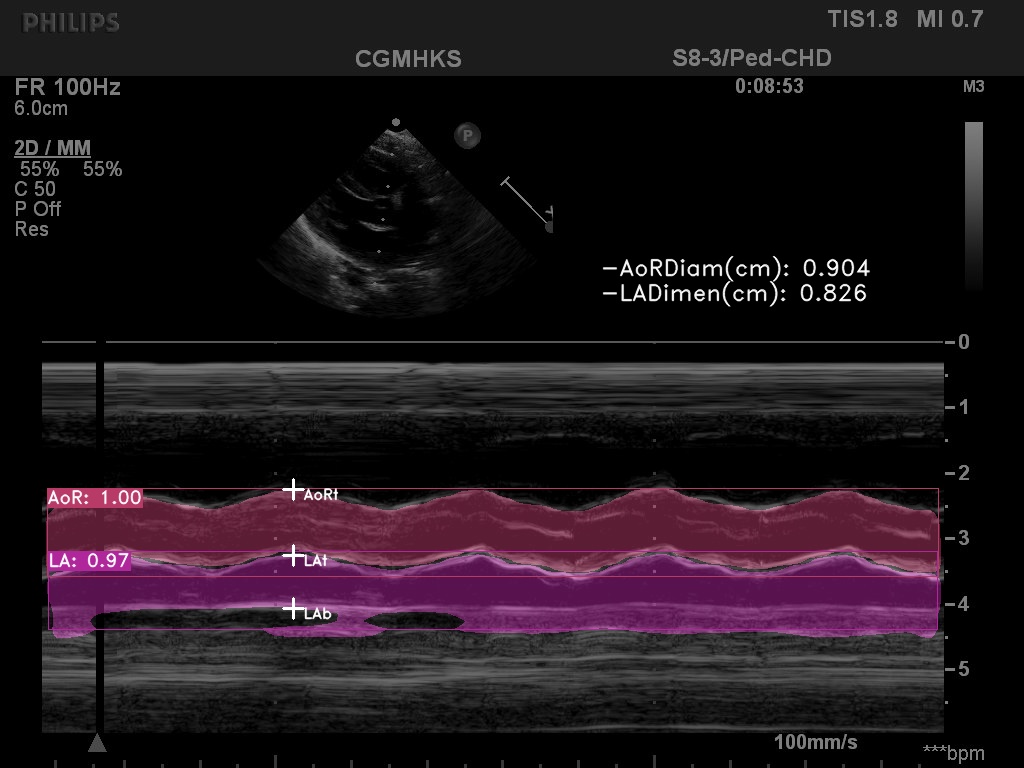} &
        \includegraphics[width=0.3\textwidth]{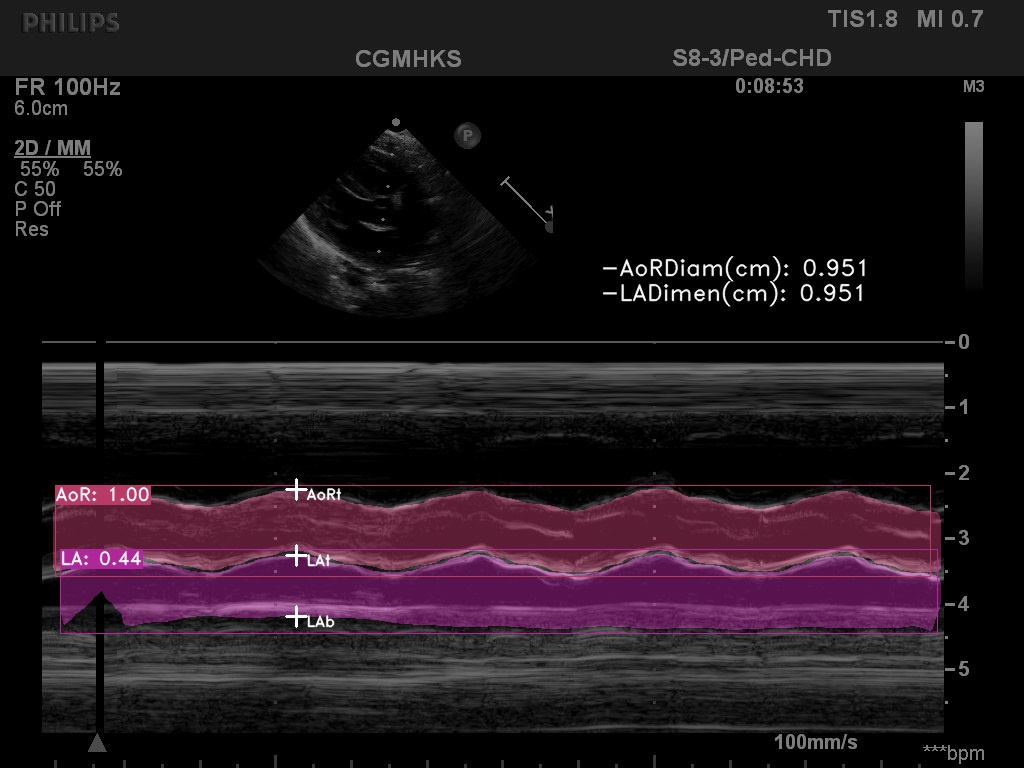}
        \\
    \end{tabular}%
    }
    \caption{AV sampled defective results. ResNet50 in maYOLACT produced an inferior mask, which messed up each other. Dilated masks occurred in RTMDet, using depth-wise convolutions. The same issue happened in ours, from the dilated effect of pixel-unshuffling operation in panel attention.}
    \label{fig14}
\end{figure*}

\begin{figure*}[!t]
    \centering
    \resizebox{\textwidth}{!}{%
    \begin{tabular}{ccccc>{\columncolor[HTML]{F2F2F2}}c}
        Ground truth & Human candidate1 & MENN UPANet80 V2 & AMEM maYOLACT & AMEM RTMDet & AMEM UPANet80 V2 (RAMEM)\\
        \includegraphics[width=0.3\textwidth]{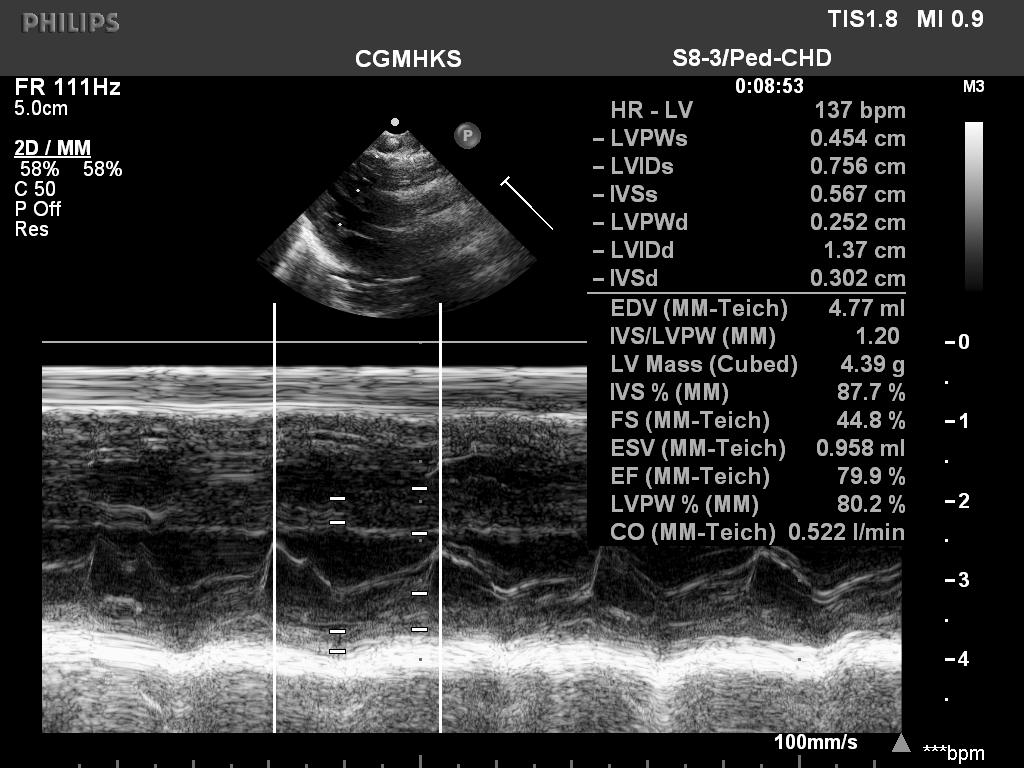} &
        \includegraphics[width=0.3\textwidth]{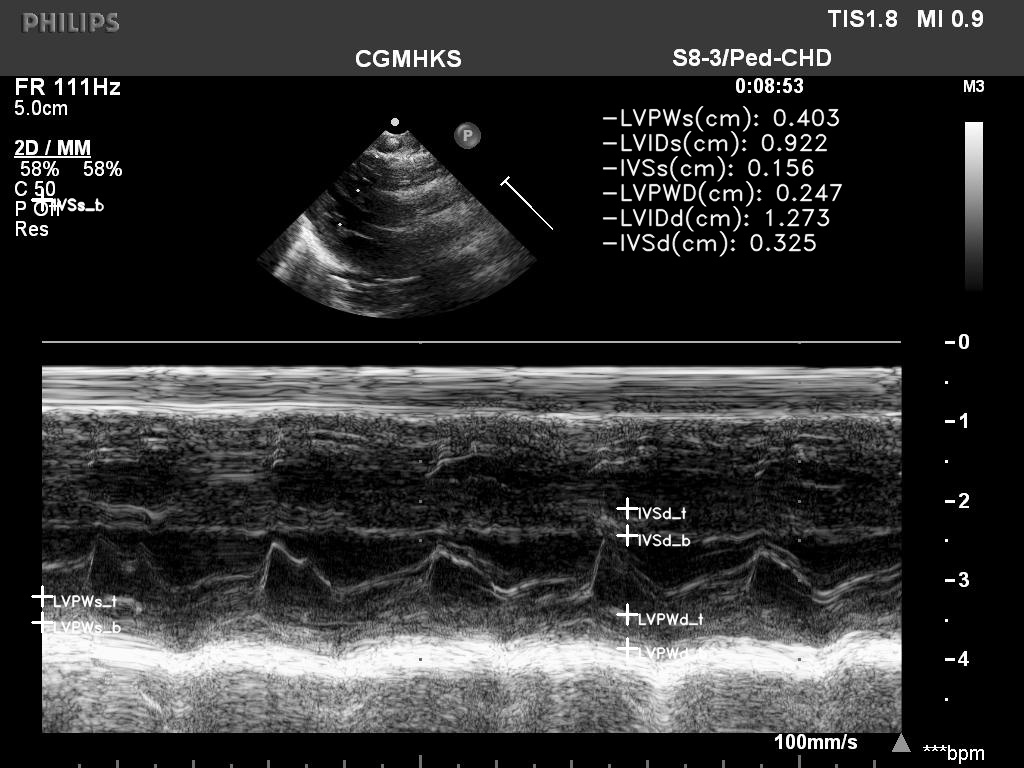} &
        \includegraphics[width=0.3\textwidth]{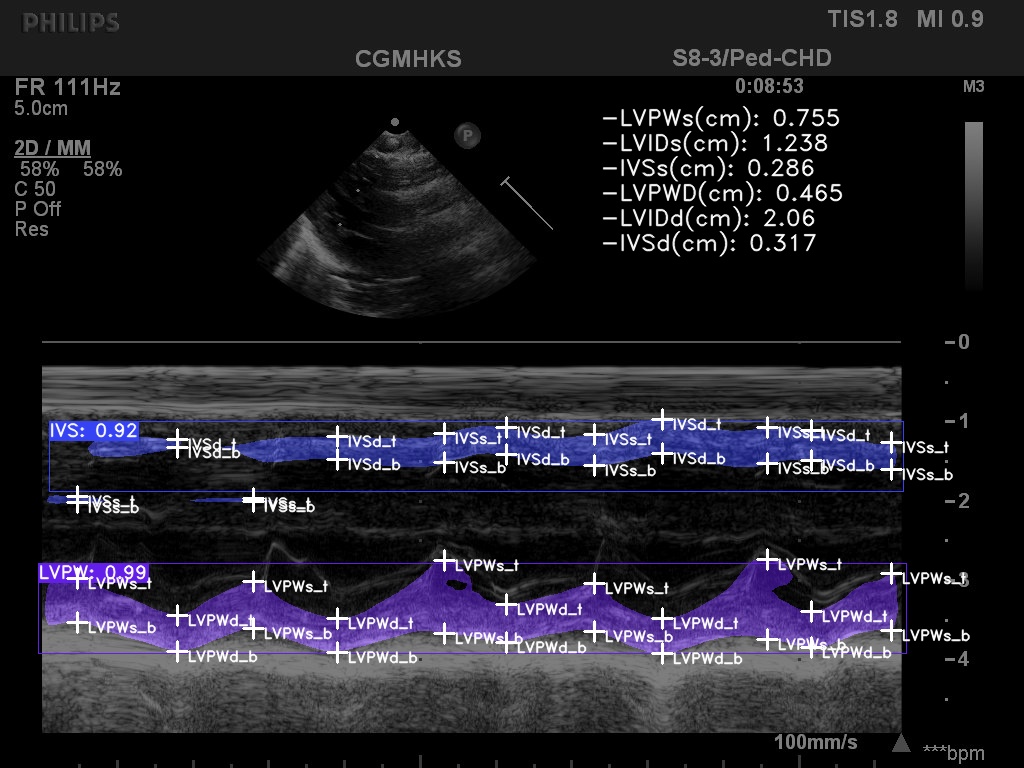} &
        \includegraphics[width=0.3\textwidth]{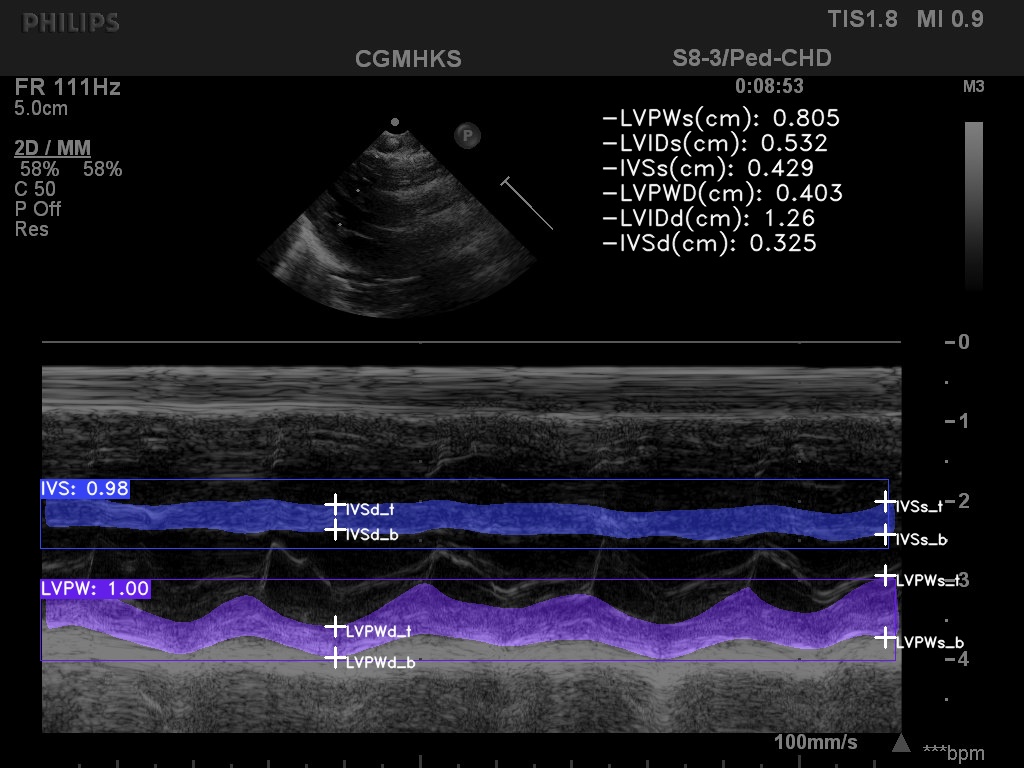} &
        \includegraphics[width=0.3\textwidth]{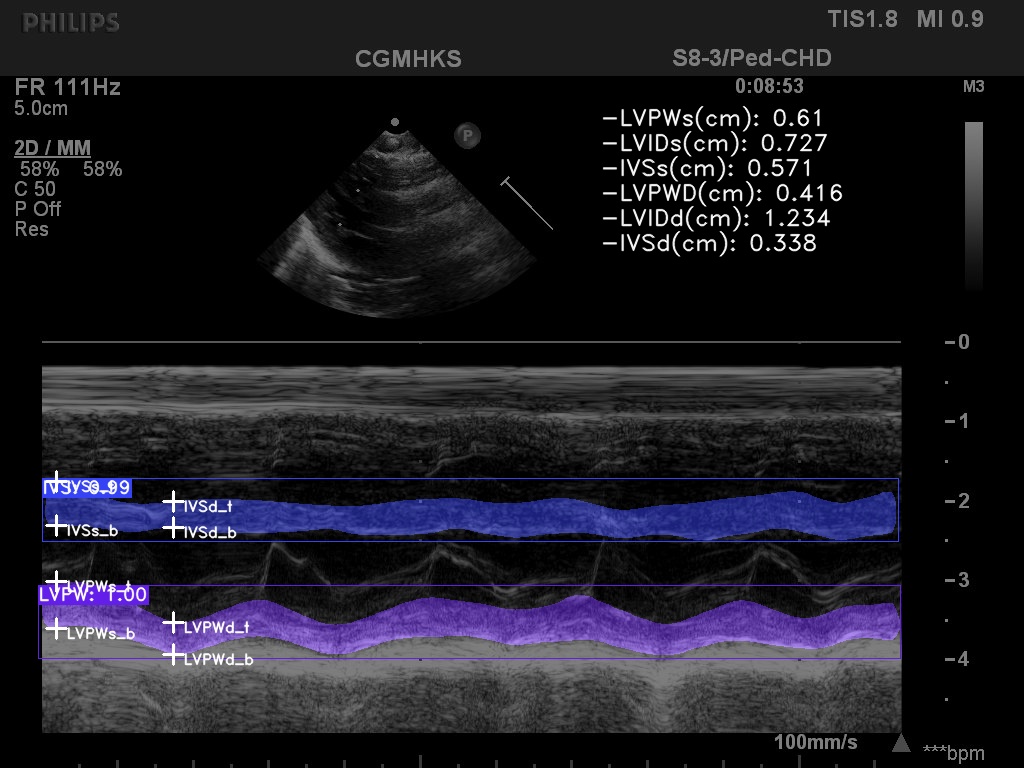} &
        \includegraphics[width=0.3\textwidth]{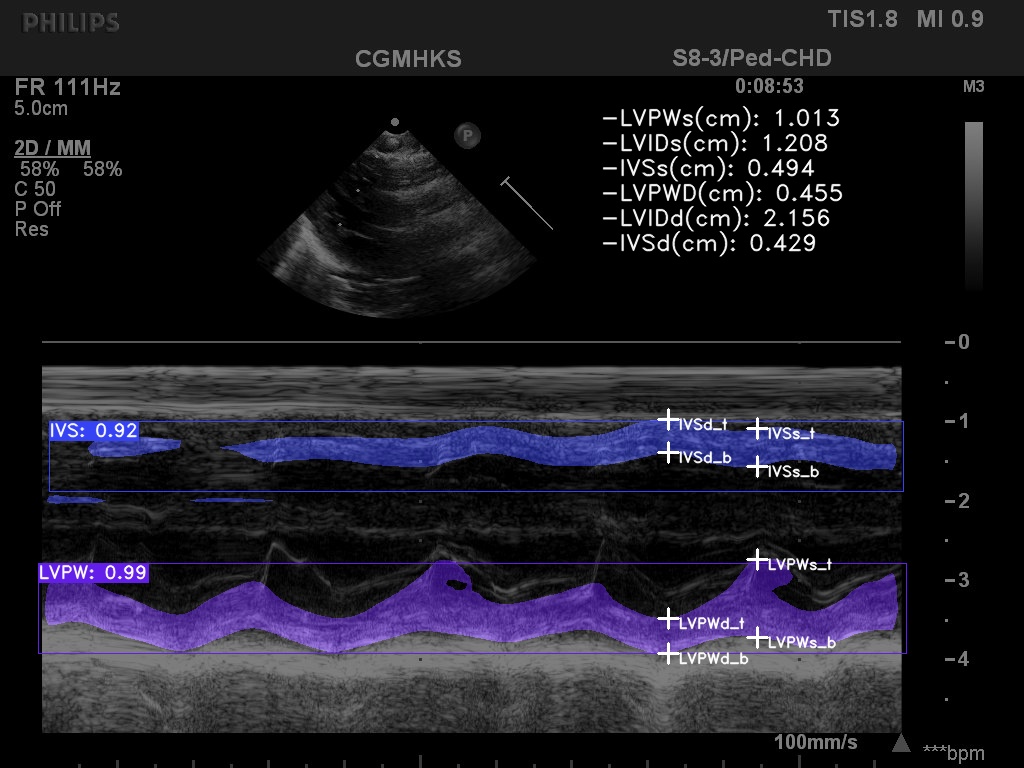}
        \\
    \end{tabular}%
    }
    \caption{LV sampled defective results. The same dilated issue can be seen in ours. However, a bigger, more accurate mask becomes possible because of a bigger receptive field.}
    \label{fig15}
\end{figure*}

This work, whereas, proposes the MEIS dataset along with panel attention in the updated backbone and AMEM toward M-mode echocardiography, each contribution has imperfect. As the outcomes of the proposed methods have been widely discussed, the limitations and the possible solutions are worth mentioning in the following.

MEIS dataset – The dataset consists of 2.6k images with labels across AV and LV, but the number is relatively small compared with PASCAL 2012 SBD. There are 3.9k images in total. In other words, the remaining 1.3k images are unlabeled. On the other hand, considering the storage efficiency, the data format is an image instead of a clip or video. Although the scheme trained upon images can still be applied to detect clips, it contradicts the typical notion of directly training on clips. Therefore, a bigger dataset is needed by completing the rest of the images' labelling. A more intuitive format of training on videos should also be considered.

Panel attention and UPANets V2 – This proposed attention contributes to better detection results by ushering global attention during down-sampling. This makes the receptive field global and diminishes the side effect of high complexity as conventional non-local blocks. Despite reducing a great margin complexity, it is still larger than pure CNNs and non-parametric-involved methods, such as HAMs. In the scenario of using non-parametric-involved attention, the hyper-parameters setting toward the best converge iterations in HAMs or EMANet could become another issue. Also, although UPANets V2 make big object detection easier, the dilated effect from pixel-unshuffling could cause the outputs mask defect, seeing masks in Fig. \ref{fig14}, despite more flaws occurring in ResNets and RTMDet in AV. Regarding the issues, non-parametric-involved attention without an expert hyper-parameter setting is possible, which could also bypass the dilated effect from current pixel-unshuffling.  

AMEM - The proposed measuring algorithm is built upon practical medical experience; although it makes results close to expertises' manual labelling, there are two downsides: high time cost and high reliance on masks. For high time cost, when observing in Table \ref{tab5}, it is noticeable that FPS is higher than in Table \ref{tab3}. However, the total FPS, including measurement, slides down by around 28 in Table \ref{tab6}, which indicates the algorithm, alone, contributes high computational overhead, even possessing a relatively more efficient than MENN and manual labelling. For using the backbone as ResNet50, a limited receptive field will drag down the FPS, which leaves insufficient spare time for AMEM. The reasons behind this dilemma lie in both FindCountours and non-mask pixels upwards (or downwards) searching operating in CPU. Viewing that optimizing the FindCountours is far beyond the scope of this work, Boosting pixel searching is a possible route. That route could be adding heads for making regression predictions similar to the bounding box in objection detection. By this, a time costing issue could be diminished by consorting GPU accelerating in directly outputting the coordinate of the topmost point; In terms of the high reliance on masks, the defective masks predictions in Fig. \ref{fig14} and Fig. \ref{fig15} can explain such the situation, which broken masks might make anchors lie in the wrong location. While the errors in Table \ref{tab6} are pleasing compared with humans, IVSd and LVPWs are still worse than MENN. The underperformed IVS in diastole is because of the nature of the less-obvious (or less-fluctuated) mask in IVS, seeing the blue mask in Fig. \ref{fig7}. As the IVSd depends on the diastole moment on the LVPW mask, the extraction of IVSd in AMEM could further be affected by defective masks. Therefore, the locating systole of a bad mask could also affect the LVPWs value. Therefore, a measurement based on multiple points selection as combing our AMEM and MENN should balance accuracy and FPS.

\section{Conclusion}
\label{sec6}
This work presents RAMEM aiming at fixing 1) the variance of manual labelling, 2) compromising information \& efficiency in NL attention, and 3) being unable seamlessly diagnosis in M-mode diagnosis. The proposed dataset of MEIS toward 1) is to bridge the RIS and M-mode echocardiography, making an automatic detection scheme with consistent results possible. To make an even more accurate result, viewing the big object in M-mode echocardiograms, panel attention targets to fix 2) efficiently and precisely by finding an outstanding balance. The automatic measurement algorithm of AMEM 3) is designed to be a great candidate for daily clinical practices by showing impressive performance with low error and real-time scene. With these proposals, we hope the bridge can establish the connection and flourish in medical images and general computer vision, especially for M-mode echocardiography.

\section*{Acknowledgments}
Funding: This research was funded by Kaohsiung Chang Gung Memorial Hospital, Kaohsiung, Taiwan (CORPG8L0101) and Taiwan National Science and Technology Council grant number NSTC 112-2410-H-A49 -033

Institutional Review Board Statement: The study was approved by the Institutional Review Board of Chang Gung Medical Foundation in Taipei, Taiwan (IRB No: 202001552B0C1 and 202101520B0), and all procedures were performed according to the guidelines of the Declaration of Helsinki.

\bibliographystyle{unsrt}  
\bibliography{refs}  






\end{document}